\documentclass[a4paper,fleqn]{cas-dc}

\usepackage[super,sort&compress]{natbib}
\usepackage{xcolor}
\usepackage{soul}
\usepackage{subcaption}

\newif\ifshowchanges

\showchangesfalse

\ifshowchanges
    \newcommand{\added}[1]{\textcolor{blue}{#1}}
    \newcommand{\deleted}[1]{\textcolor{red}{\st{#1}}}
    
\else
    \newcommand{\added}[1]{#1}
    \newcommand{\deleted}[1]{}
    
\fi

\def\tsc#1{\csdef{#1}{\textsc{\lowercase{#1}}\xspace}}
\tsc{WGM}
\tsc{QE}
\tsc{EP}
\tsc{PMS}
\tsc{BEC}
\tsc{DE}

\begin{document}

\newcommand{\method}{\textit{DataCook}}

\let\WriteBookmarks\relax
\def\floatpagepagefraction{1}
\def\textpagefraction{.001}

\shorttitle{\method}
\shortauthors{Sihan Shang et~al.}


\title[mode=title]{\method: Exploring Anti-Adversarial Fingerprints for Healthcare Data Usage Control}

\author[1,2,3]{Sihan Shang}[style=chinese]
\ead{shangsihan@stu.hit.edu.cn}
\credit{Conceptualization, Data curation, Formal analysis,
Investigation, Methodology, Software, Validation, Visualization,
Writing -- original draft, Writing -- review \& editing}

\author[4]{Jun Yan}[style=chinese]
\ead{yanjun@ieee.org}
\credit{Formal analysis, Validation,
Writing -- original draft, Writing -- review \& editing}

\author[5,6]{Danli Shi}[style=chinese]
\credit{Resources}

\author[7]{Haizhong Zheng}[style=chinese]
\ead{haizhonz@andrew.cmu.edu}
\credit{Writing -- review \& editing}

\author[8,9]{Shuangjia Zheng}[style=chinese]
\ead{shuangjia.zheng@sjtu.edu.cn}
\credit{Resources}

\author[10]{Wei Zhao}[style=chinese]
\ead{wei.zhao@csu.edu.cn}
\credit{Funding acquisition, Resources,
Writing -- review \& editing}

\author[1,2]{Jiancheng Yang}[style=chinese]
\ead{jiancheng.yang@aalto.fi}
\cormark[1]
\credit{Conceptualization, Funding acquisition, Project administration,
Resources, Supervision, Writing -- original draft,
Writing -- review \& editing}

\affiliation[1]{
    organization={ELLIS Institute Finland},
    city={Espoo},
    country={Finland}
}

\affiliation[2]{
    organization={School of Electrical Engineering, Aalto University},
    city={Espoo},
    country={Finland}
}

\affiliation[3]{
    organization={Institute of Cyberspace Security, School of Computer Science and Technology, Harbin Institute of Technology},
    city={Shenzhen},
    country={China}
}

\affiliation[4]{
    organization={IT College, Shanghai Ocean University},
    city={Shanghai},
    country={China}
}

\affiliation[5]{
    organization={School of Optometry, The Hong Kong Polytechnic University},
    city={Hong Kong},
    country={China}
}

\affiliation[6]{
    organization={Research Centre for SHARP Vision (RCSV), The Hong Kong Polytechnic University},
    city={Hong Kong},
    country={China}
}

\affiliation[7]{
    organization={Department of Electrical and Computer Engineering, Carnegie Mellon University},
    city={Pittsburgh},
    country={United States}
}

\affiliation[8]{
    organization={Global Institute of Future Technology, Shanghai Jiao Tong University},
    city={Shanghai},
    country={China}
}

\affiliation[9]{
    organization={School of Artificial Intelligence, Shanghai Jiao Tong University},
    city={Shanghai},
    country={China}
}

\affiliation[10]{
    organization={Department of Radiology, The Second Xiangya Hospital, Central South University},
    postcode={410011},
    city={Changsha},
    state={Hunan},
    country={China}
}

\cortext[cor1]{Corresponding author}


\begin{abstract}
\noindent \textit{Purpose:}
Healthcare AI increasingly relies on high-quality datasets, yet medical data are scarce, costly to curate, and tightly regulated in terms of data access. More critically, once medical data are leaked or accessed without authorization, data owners have limited means to prevent such data from being used to train and deploy downstream models. To address this gap, we propose \(\method\), a deployment-stage usage-control framework based on anti-adversarial fingerprints.

\noindent 
\textit{Materials and Methods:}
Specifically, \(\method\) transforms raw samples through a private cooking procedure that embeds anti-adversarial fingerprints. Models trained on the cooked data retain strong utility when evaluated on similarly cooked inputs, while their performance on raw inputs is reduced, making downstream model utility dependent on access to the private cooking function. We evaluate \(\method\) on 2D, 3D, and high-resolution medical images, as well as structured healthcare tabular data, across multiple model architectures and adaptive attack settings.

\noindent \textit{Results:}
Across the evaluated settings, \(\method\) generally preserves
authorized model utility while reducing performance on unauthorized raw
inputs. The authorization-dependent behavior is observed across multiple
medical imaging settings, model architectures, and adaptive attack
conditions, supporting the feasibility of the proposed deployment-stage
usage-control mechanism.

\noindent \textit{Conclusion:}
These results demonstrate the feasibility of authorization-dependent model use at deployment, while the effectiveness of the current framework remains dependent on the specific task and evaluation metric.

\end{abstract}


\maketitle


\section{Introduction}
\label{sec1}

In healthcare, the rapid development and widespread application of artificial intelligence have increased both the value of medical data and the risks of healthcare data misuse and copyright protection~\cite{canadian2019canadian,kim2021privacy,pan2024opportunities}. Given the sensitivity and confidentiality of medical data, it is crucial to protect patient privacy as well as the intellectual property rights of data providers and creators~\cite{price2019privacy,ienca2023dontpause,smith2023stoptalking}. 
Recent biomedical studies further integrate medical imaging with proteomic, radiomic, genomic, and molecular information to characterize complex disease processes and support clinical risk assessment
\cite{lian2026multimodal,yang2026radiomics, ran2026splicing,kuang2023neurodevelopmental, FENG2024100068}. As healthcare datasets become increasingly heterogeneous and valuable, their unauthorized acquisition may enable downstream models to be trained and deployed with limited control from the original data owners.

\begin{figure*}[!t]
    \centering
    \scriptsize
    \includegraphics[width=0.75\textwidth]{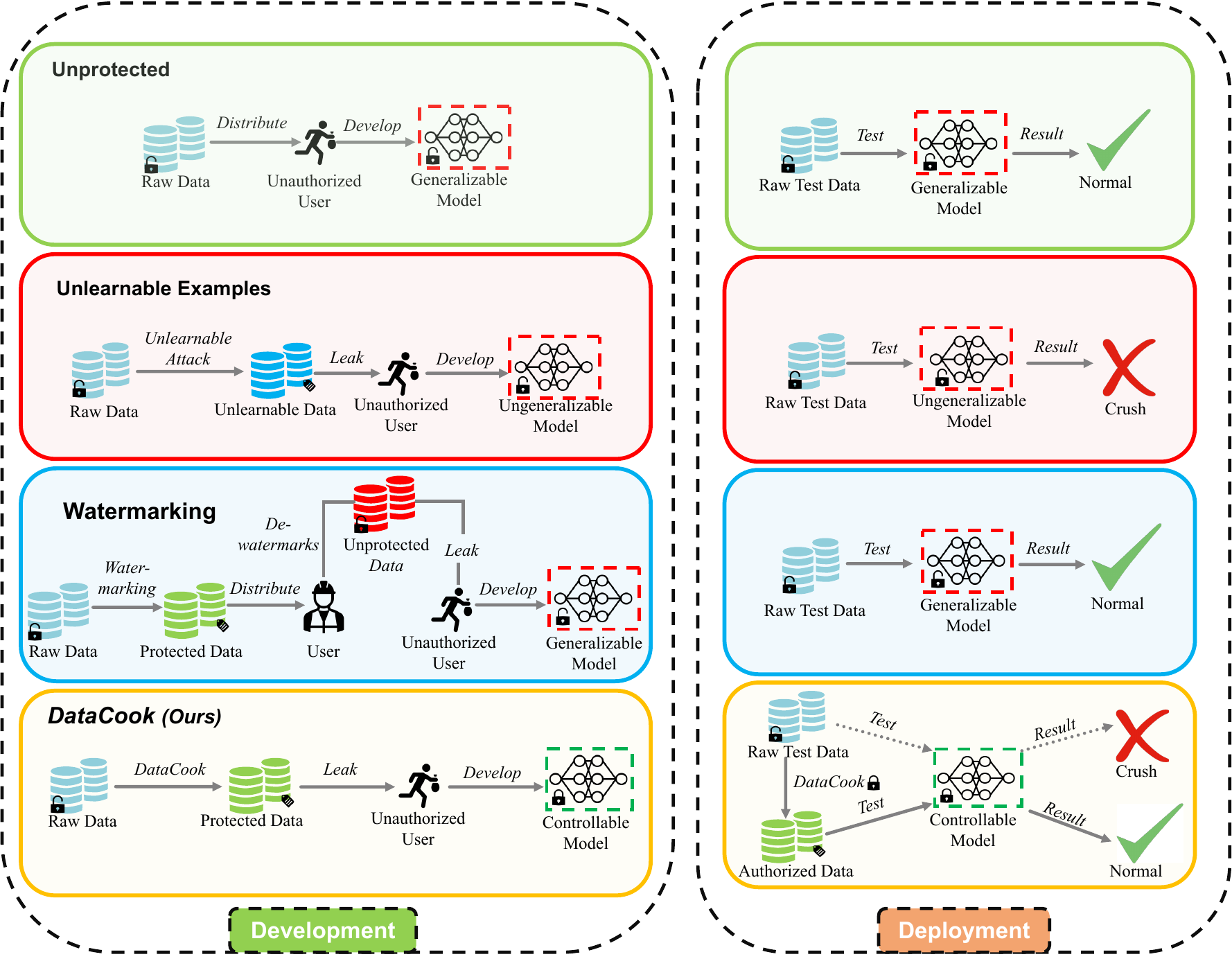}
    \caption{ \textbf{Comparison of Data Protection Paradigms.} 
    \textbf{Unprotected Data:} Data can be used to train and deploy downstream models without restriction. 
    \textbf{Unlearnable Examples:} Perturbations prevent models from learning useful representations, but also reduce data utility for legitimate use. 
    \textbf{Watermarking:} Watermarks support ownership verification or discourage copying, but generally do not restrict model utility once usable data are obtained. 
    \textbf{\(\method\):} Cooked data retain their utility for authorized users, while models trained on them exhibit reduced performance on raw inputs, extending usage control to model deployment.}
    \label{fig:Overview}
\end{figure*}

Existing data-protection techniques address only part of this problem,
as summarized in Fig.~\ref{fig:Overview}. Watermarking embeds visible or hidden patterns into protected data and can discourage direct copying or help identify whether a model has been trained on a protected dataset~\cite{wei2023preventing}. However, watermarking generally does
not restrict the downstream model utility of a trained model once the data have been acquired.
Unlearnable examples (UEs)~\cite{EM,LSP,UC,TAP}, by contrast, introduce imperceptible perturbations that prevent models from learning useful representations from the released data. Although effective in reducing unauthorized training utility, UEs also make the protected data difficult to use
for authorized users. This limitation is particularly important in medical imaging, where data and expert annotations are scarce and costly~\cite{litjens2017survey}.  These observations reveal a key gap: how can data owners restrict data from misuse in unauthorized training while preserving its utility for authorized users?


To \deleted{address}\added{investigate} this gap, we introduce \(\method\),  a framework that \added{explores the }use\deleted{s} \added{of} anti-adversarial fingerprints \deleted{to control} \added{for }healthcare data us\deleted{e}\added{age control} during model deployment. \deleted{It}\(\method\) applies a private “cooking” procedure to raw data before distribution. \deleted{When a downstream model is trained on the resulting cooked samples, its utility during deployment dependent on whether incoming inputs are processed by the same private cooking function.} \added{Downstream models trained on the resulting cooked samples may exhibit different performance when evaluated on raw versus similarly cooked inputs, making their utility potentially dependent on access to the private cooking function. } \deleted{Authorized users can therefore retain normal model performance by cooking their inputs, whereas evaluation on raw inputs results in reduced performance. In this way, }\added{Our experiments investigate this authorization-dependent behavior under a specified threat model and examine its effectiveness across different tasks, architectures, and evaluation metrics. We position }\(\method\) \deleted{makes downstream model utility dependent on access to the private cooking function, extending data usage control from the distribution stage to model deployment.} \added{ as a proof-of-concept approach rather than a universally effective mechanism for preventing unauthorized data use. } Our main contributions are summarized as follows:

\begin{itemize}
    \item We formulate healthcare data protection as a deployment stage usage-control problem, aiming to preserve utility for authorized users while restricting the downstream utility available to unauthorized users.

    \item  We propose \(\method\), an anti-adversarial fingerprinting framework that uses pseudo labels generated by a surrogate model to guide perturbations that enhance model confidence, preserving utility on cooked inputs while reducing performance on raw inputs.

    \item We conduct extensive evaluations on 2D, 3D, and high resolution medical images, as well as structured healthcare tabular data. Across different model architectures, perturbation budgets, and adaptive attack \deleted{, the results demonstrate the effectiveness of Datacook
    in preserving authorized utility while reducing unauthorized performance across diverse healthcare tasks. } \added{settings. The results reveal authorization-dependent performance differences in several medical imaging tasks while highlighting limitations in multi-label classification and structured tabular data. These findings characterize the potential and task-dependent boundaries of the proposed framework.}
    
\end{itemize}


\section{Related Work}
\label{sec:related_work}

Medical data protection covers different objectives and stages of the data lifecycle, including ownership verification, prevention of unauthorized learning, privacy-preserving training and computation, and deployment-stage usage control. We review these directions and clarify the positioning of \(\method\).

\paragraph{Dataset Watermarking.}
Dataset watermarking embeds secret patterns or signals into protected data to enable subsequent ownership verification. Backdoor-based methods introduce trigger-dependent behaviors into models trained on protected datasets, allowing data owners to detect unauthorized use through abnormal outputs~\cite{adi2018turning,li2023black}. Other methods apply removable transformations to released data and provide authorized users with a private restoration procedure~\cite{wei2023preventing}. However, watermarking mainly provides post-hoc ownership evidence or controls access to the original data. It does not necessarily prevent an unauthorized model from retaining normal predictive performance, and reversible methods remain vulnerable if restored data or restoration procedures are leaked. In contrast, \(\method\) focuses on making downstream model utility dependent on access to a private cooking procedure.

\paragraph{Unlearnable Examples.}
Unlearnable examples originate from availability poisoning attacks, which modify training samples to reduce the generalization performance of models trained on them~\cite{biggio2012poisoning}. Representative methods include Error-Minimum perturbations~\cite{EM}, Adversarial Poisoning~\cite{TAP}, Synthetic Perturbations based on linearly separable noise~\cite{LSP}, and the label-agnostic Unlearnable Clusters method, which constructs cluster-wise perturbations in the feature space~\cite{UC}. These methods are the most directly related baselines because, like \(\method\), they protect released data through input-level perturbations and can be evaluated under a consistent downstream training protocol. Nevertheless, conventional unlearnable examples primarily aim to suppress model generalization and generally do not consider how authorized users can preserve or recover model utility. \(\method\) instead uses pseudo-label-guided anti-adversarial fingerprints to create an authorization-dependent performance gap between cooked and raw inputs.

\paragraph{Privacy-Preserving Learning and Computation.}
Federated learning enables multiple institutions to collaboratively train models while keeping raw data locally distributed~\cite{mcmahan2017communication,sheller2020federated}. Differential privacy limits the influence of individual records through mechanisms such as gradient clipping and calibrated noise~\cite{dwork2006calibrating,abadi2016deep}, while homomorphic encryption supports computation directly over encrypted inputs~\cite{gentry2009fully,giladbachrach2016cryptonets}. These techniques have also been combined in privacy-preserving medical-imaging systems~\cite{kaissis2021endtoend}. They primarily protect data confidentiality during collaborative training, model release, transmission, or computation, whereas \(\method\) targets downstream usage control after protected data have been released. Because their objectives, threat models, and evaluation criteria differ substantially from CP and PP, they are complementary technologies rather than directly comparable numerical baselines.


\section{Materials and Methods}\label{sec2}

\subsection{Threat Model and Deployment Workflow}

We consider three entities in the \(\method\) deployment scenario: the data owner, authorized users, and unauthorized users. The data owner releases a cooked training dataset and its labels while retaining the private cooking function, surrogate model, and perturbation-generation parameters. Authorized users can access the cooking function through a controlled interface or use pre-cooked test inputs, whereas unauthorized users may train arbitrary downstream models on the released data and attempt input purification, data augmentation, or fine-tuning to recover raw-input performance. However, they cannot cook new test inputs without the private function and associated parameters. Under this threat model, \(\method\) aims to create an authorization-dependent performance gap, with models performing well on cooked inputs but poorly on raw inputs. Because leakage of the cooking function would enable unauthorized users to recover authorized inference capability, \(\method\) should be understood as a deployment-stage technical usage-control mechanism rather than a complete solution for legal copyright enforcement, privacy protection, or secure data governance.

\subsection{Problem Formulation}

Given a raw healthcare dataset to be released as $\mathcal{D}^r = \{(x_i, y_i)\}_{i=1}^{N}$, where the input space $\mathcal{X} = \{x_i\}_{i=1}^{N} \subset \mathbb{R}^d$  represents an input sample and $\mathcal{Y} =\{y_i\}_{i=1}^{N} = \{1, \ldots, c\}$ denotes its associated target. Here, $N$ denotes the number of samples, and \(\mathcal{Y}\) may represent binary, multiclass, or multilabel targets.

Normally, to protect data copyright, unlearnable examples are created using availability poisoning attacks. Aiming to solve the following bi-level objective in terms of perturbations $\delta_i$ which are applied to the elements $x_i$ in dataset $\mathcal{D}^r$:
\begin{align}
     &\underset{\delta \in S}{\text{max }}
     \mathbb{E}_{(x,y) \sim D^r}[\mathcal{L}(f(x,\theta(\delta)), y)] \label{eq:max}, \\
     \text{s.t. } \theta \in &\underset{\theta}{\text{arg min}} \sum_{(x_i,y_i )\in D^r} \mathcal{L}(f(x_i+\delta_i;\theta), y_i) \label{eq:min},
\end{align}
where $S$ represents the set of constraints on the perturbations. Based on the setting of previous unlearnable examples~\cite{EM}\cite{LSP}\cite{TAP}. The perturbations are constrained by an $L_\infty$ budget $\epsilon$, whose value depends on the experimental setting. The modality-specific configurations are provided in Appendix~\ref{app:image_config}.

The outer objective reduces the generalization performance of a model trained on the perturbed data, while the inner objective describes downstream model training. Directly solving this bi-level problem is computationally expensive because the inner optimization process (Eq.~(\ref{eq:min})) requires multiple iterations to converge, and each gradient update affects the subsequent iteration, making the backpropagation process highly complex. 

Unlike unlearnable examples, \(\method\) aims to reduce model utility under unauthorized raw-input evaluation while preserving utility for authorized users. For each raw sample, we construct a cooked
sample \(x_i^{p}=x_i+\delta_i\), producing the cooked dataset
\(\mathcal{D}^{p}=\{(x_i^{p},y_i)\}_{i=1}^{N}\).
Let \(f^{r}\) denote a reference model trained on
\(\mathcal{D}^{r}\), and let \(f^{p}\) denote a protected model trained
on \(\mathcal{D}^{p}\). Given an evaluation metric
\(\mathcal{E}(f,\mathcal{D})\), \(\method\) is designed to satisfy three
objectives:
\begin{align}
    \max \quad &
    \left[
    \mathcal{E}(f^{r},\mathcal{D}^{r})
    -
    \mathcal{E}(f^{p},\mathcal{D}^{r})
    \right],
    \label{eq:minimize}
    \\
    \min \quad &
    \left|
    \mathcal{E}(f^{p},\mathcal{D}^{p})
    -
    \mathcal{E}(f^{r},\mathcal{D}^{r})
    \right|,
    \label{eq:minimize1}
    \\
    \text{s.t.}\quad &
    \|\delta_i\|_{\infty}\leq\epsilon,
    \quad \forall i.
    \label{eq:minimize2}
\end{align}
where \(\epsilon\) denotes the maximum perturbation allowed in pixel space. This max-min formulation captures the core objective of \(\method\): maximizing degradation of unauthorized utility on raw
inputs while minimizing the performance gap between authorized cooked inputs and the clean reference, subject to a bounded \(\ell_{\infty}\) perturbation. These objectives are quantified by  \textit{CP} and \textit{PP} introduced in Section~\ref{sec:metrics}.

\subsection{\method}

\begin{figure*}[t]
    \centering
	\includegraphics[width=0.9\textwidth]{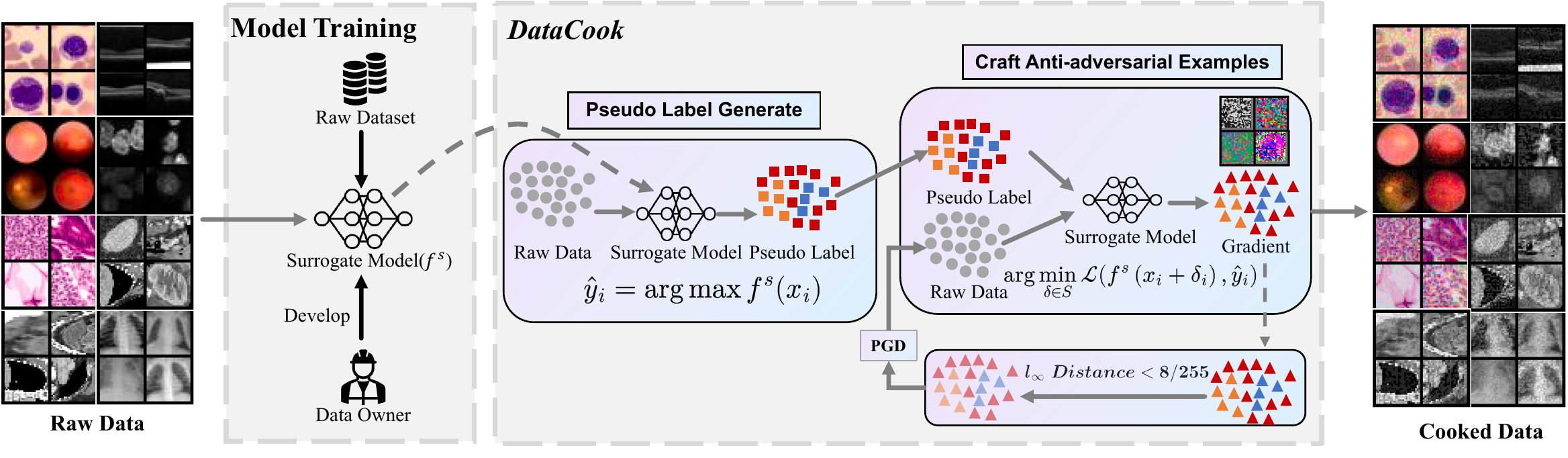}
	\caption{\textbf{Generating Anti-Adversarial Examples with \method{}.} The surrogate model's highest-probability prediction is used as the pseudo label \(\hat{y}\) to guide iterative anti-adversarial perturbations within a bounded budget.}
	\label{fig:datacook}
\end{figure*}

Based on the objectives above,we propose \(\method\), which generates controllable anti-adversarial fingerprints using pseudo labels rather than ground-truth labels. The overall workflow is illustrated in Fig. \ref{fig:datacook}. Given a raw sample \(x_i\), a surrogate model \(f^{s}\) produces a pseudo label from its highest-probability prediction. The pseudo label then serves as the target of the anti-adversarial optimization, which generates a sample-specific perturbation \(\delta_i\). The resulting perturbed sample \(x_i^{p}=x_i+\delta_i\) is treated as a cooked sample, and all cooked samples together form the dataset used for downstream training. A downstream model trained on the cooked dataset therefore learns from the fingerprinted input distribution. During deployment, authorized users apply the same private cooking procedure to incoming inputs, whereas users without access to the procedure evaluate the model on raw inputs. This creates the authorization-dependent utility gap defined in the previous section. The following paragraphs describe the anti-adversarial objective, pseudo label construction, and optimization procedure in detail.


\paragraph{Crafting Anti-Adversarial Examples}

To construct cooked samples, we use adversarial examples to address the optimization problem defined above. This approach is inspired by prior work~\cite{TAP} which uses
adversarial examples to approximate bi-level optimization in poison generation. Let
\(S_{\epsilon}=\{\delta:\|\delta_i\|_{\infty}\leq\epsilon,\forall i\}\)
denote the bounded perturbation set.  Adversarial examples introduce perturbations into raw inputs to maximize the prediction loss:
\begin{equation}
    \label{equation:equation2datacook}
    \arg\max_{\delta \in S} \left[ \sum_{(x_i, y_i) \in D^r} \mathcal{L} \left( f^s \left( x_i + \delta_i \right), y_i \right) \right].
\end{equation}
Here, $f^s$ denotes a surrogate model trained on clean data, whose parameters remain fixed during perturbation generation. Because the perturbations are informed by data gradients, they encode targeted features rather than arbitrary noise~\cite{Adversarialfeatures}. Moreover, studies on adversarial training show that models can learn from perturbed inputs and improve their robustness to such perturbations~\cite{goodfellow2014explaining,madry2017towards,zhang2019tradeoff,yang2020learning}. This motivates the consistent use of perturbed samples during training and authorized inference, in line with the performance preservation objective in Eq.~(\ref{eq:minimize1}). However, adversarial examples maximize prediction loss and move samples away from their predictive regions, which may reduce model
performance. We therefore introduce anti-adversarial examples~\cite{alfarra2022combating,wang2022watermarking}. In contrast,  anti-adversarial examples move samples toward regions of higher confidence under the surrogate model,  which aligns with the performance preservation objective in Eq.~(\ref{eq:minimize1}). Accordingly, we reverse the optimization direction:
\begin{equation}
    \label{equation:equation1datacook}
    \arg\min_{\delta \in S} \left[ \sum_{(x_i, y_i) \in D^r} \mathcal{L} \left( f^s \left( x_i + \delta_i \right), y_i \right) \right].
\end{equation}

\paragraph{Pseudo Label Guided Fingerprints.}

Although the anti-adversarial objective supports performance preservation, directly using the ground-truth labels injects class information into the perturbations. Because perturbations are generated using a supervised loss, they may overfit to features associated with the class labels~\cite{UC}. This issue may limit generalization when the same cooking procedure is applied to new inputs. To mitigate this, we use the highest probability prediction of the surrogate model as a pseudo label and define the final cooking objective as:
\begin{equation}
\label{equation:equationdatacook}
\begin{aligned}
    \arg\min_{\delta \in S} \quad 
    & \left[\sum_{(x_i, y_i) \in D^r} 
    \mathcal{L}\left(f^s(x_i + \delta_i), \hat{y}_i\right) \right], \\
    \text{where} \quad 
    & \hat{y}_i = \arg\max_c p^s(c|x_i) \\
    \text{and} \quad 
    & p^s(c|x_i) = \text{softmax}(f^s(x_i))_c.
\end{aligned}
\end{equation}
Here, \(\hat{y}_i\) denotes the pseudo label generated by the surrogate model \(f^s\). It provides a sample specific optimization target without using the reference label. The objective addresses the optimization goals in Eqs.~(\ref{eq:minimize})--(\ref{eq:minimize2}) by reducing model utility on raw inputs while preserving utility on cooked inputs. Pseudo label agreement with the reference labels and surrogate confidence are reported in Appendix~\ref{app:pseudo_quality}.

\paragraph{PGD Optimization.} Projected Gradient Descent (PGD) is commonly used to generate adversarial perturbations for deep networks~\cite{ madry2017towards}. Accordingly, we generate cooked data by applying PGD to the anti-adversarial objective in
Eq.~(\ref{equation:equationdatacook}). At each iteration, PGD minimizes the loss with respect to the pseudo label and projects the perturbation onto the permitted \(\ell_{\infty}\) region. The surrogate model remains fixed throughout this process. Unlike methods that require iterative updates to both the model and the perturbations~\cite{EM}, \(\method\) optimizes only the input
perturbations. In addition, the use of surrogate predictions removes the need for ground-truth labels when cooking new samples, allowing the same procedure to be applied to training data and incoming inputs during authorized deployment.

\subsection{Evaluation Metrics}
\label{sec:metrics}
We evaluate anti-adversarial fingerprints using two key metrics: Copyright Protection (\textit{CP}) and Performance Preservation (\textit{PP}). \textit{CP} corresponds to Eq.~(\ref{eq:minimize}).  \textit{CP} is defined as \(\mathcal{E}(f^p, \mathcal{D}^r) - \mathcal{E}(f^r, \mathcal{D}^r)\). More negative \textit{CP} indicates a larger reduction in unauthorized utility and therefore stronger protection. \textit{PP} corresponds to Eq.~(\ref{eq:minimize1}), we define \textit{PP} = \(\mathcal{E}(f^p, \mathcal{D}^p) - \mathcal{E}(f^r, \mathcal{D}^r)\). A \textit{PP} value closer to zero indicates better preservation of authorized utility. Both \textit{CP} and \textit{PP} are reported in percentage points. Following prior work on unlearnable examples~\cite{EM,LSP,TAP}, we use
accuracy where appropriate. Beyond accuracy (ACC), we report task-appropriate metrics to provide a more complete evaluation of authorized utility and residual unauthorized utility. For binary tasks, we report AUC, balanced accuracy, F1-score, sensitivity, specificity, and expected calibration error (ECE). For multiclass tasks, we report macro-AUC, balanced accuracy, macro-F1, macro-sensitivity, macro-specificity, and ECE. For multilabel tasks, we report macro/micro-AUC, macro/micro-F1, sensitivity, specificity, and Hamming loss. We further report the residual unauthorized utility relative to a task-specific random-guessing baseline. Complete task-aware results are provided in Appendix~\ref{app:task_aware_metrics}. Seed-level stability, confidence
intervals, paired effect sizes, and multiplicity-adjusted exploratory
tests are reported in Appendix~\ref{app:statistics}.

\section{Results}\label{sec3}
We describe the experimental setup and evaluate \(\method\) on MedMNIST 2D, 3D, and high-resolution image datasets, together with structured healthcare tabular data. We assess its protection effectiveness, performance preservation, and generalization across datasets and model architectures.


\begin{figure*}[t]
    \centering
    \includegraphics[width=0.75\textwidth]{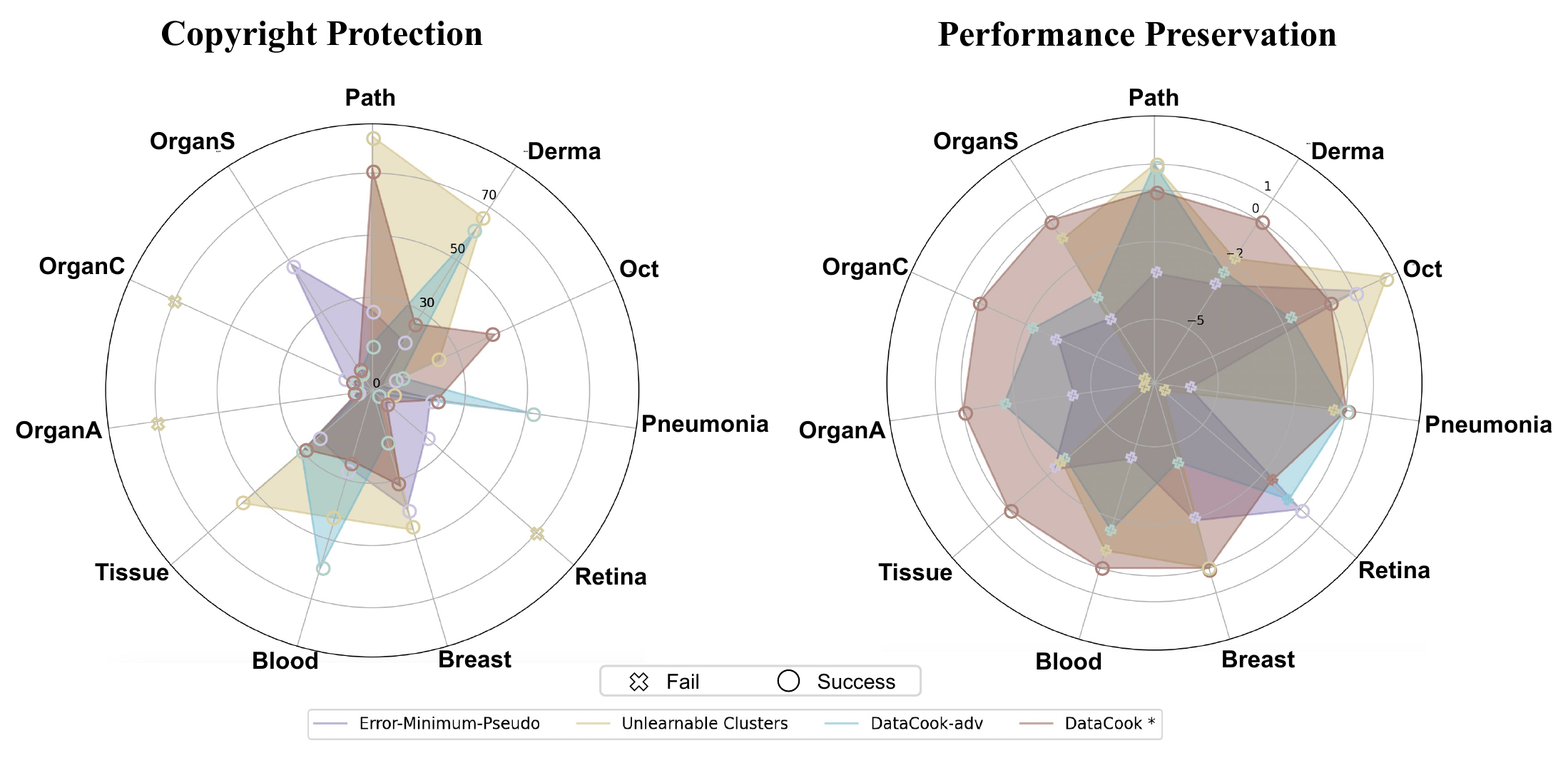}
    \caption{
    \textbf{Aggregate comparison of \(\method\) with perturbation-based
    data-protection baselines on 2D medical datasets.}
    The radar chart compares Error-Minimum (Pseudo),
    Unlearnable Clusters~\cite{UC}, \(\method\) (Adv), and \(\method\)
    across the evaluated dimensions. Lower \textit{CP} indicates
    stronger protection, while \textit{PP} values closer to zero
    indicate better authorized utility preservation.
    }
    \label{fig:main_2d_comparison}
\end{figure*}

\begin{table*}[[!t]]
\small

\caption{\textbf{Protection Results on MedMNIST 2D Datasets.} \added{CP and PP are reported in accuracy percentage points as mean $\pm$ standard deviation across datasets. The standard deviations reflect cross-dataset heterogeneity rather than seed-level uncertainty. Label indicates ground-truth label usage, and Type indicates perturbation granularity.}}

\label{tab:MedMNIST2D}
\centering

\begin{tabular}{lccccc}


\toprule[1pt]\midrule[0.3pt]

\multirow{2}{*}{Method} & 
\multirow{2}{*}{Label} &
\multirow{2}{*}{Type} & 
\multirow{2}{*}{SSIM} & 
\multirow{2}{*}{$\textit{CP}$$\downarrow$} & 
\multirow{2}{*}{$\textit{PP} \rightarrow 0$} \\

& & & & & \\

\midrule[1pt]

Error-Minimum~\cite{EM} & W/ & Sample & \( 0.94\pm 0.02 \) & \(-26.36\pm13.54\) & - \\

Adversarial Poison~\cite{TAP} & W/ & Sample & \( 0.94\pm 0.01 \) & \(-27.61\pm17.05\) & - \\

Synthetic Perturbations~\cite{LSP} & W/ & Class & \( \colorbox{red!20}{0.87}\pm0.08 \) & \(-38.88\pm16.24\) & - \\

\midrule

Unlearnable Clusters~\cite{UC} & W/O & Class & \( 0.96\pm 0.02 \) & \(-28.61\pm\colorbox{red!20}{29.76}\) & \(\colorbox{red!20}{-1.09}\pm \colorbox{red!20}{1.29}\) \\

Error-Minimum (Pseudo) & W/O & Sample & \( 0.93\pm 0.02 \) & \(\colorbox{red!20}{-21.82}\pm13.11 \) & \(\colorbox{red!20}{-2.92}\pm\colorbox{red!20}{1.97}\) \\


\midrule

\(\method\) (Adv) & W/O & Sample & \( \colorbox{red!20}{0.94}\pm 0.01 \) & \(-24.01\pm\colorbox{red!20}{21.98} \) & \(\colorbox{red!20}{-1.79}\pm\colorbox{red!20}{1.45}\) \\

\(\method\) & W/O & Sample & \( 0.95\pm 0.01 \) & \(-24.65\pm19.01 \) & \(-\textbf{0.14}\pm\textbf{0.45}\) \\

\midrule[0.3pt]\bottomrule[1pt]

\end{tabular}
\end{table*}
\subsection{Experimental Setup} 

\paragraph{Datasets and Models.}
We evaluate \(\method\) on the publicly available MedMNIST benchmark~\cite{medmnistv2,medmnistv1}, covering 2D, 3D, and high-resolution medical images from modalities including X-ray, magnetic resonance imaging, computed tomography, ultrasound, and microscopy. We further include structured healthcare tabular datasets~\cite{smith1988using,hcv_data_571,sudlow2015uk} to assess performance across heterogeneous data types. All experiments are implemented in PyTorch 2.0~\cite{paszke2019pytorch} using ResNet18, ResNet50~\cite{he2016deep}, VGG16~\cite{simonyan2014very}, and ConvNeXt-Tiny~\cite{convnext}. Except for VGG16, the models are initialized with ImageNet-1k pre-trained weights~\cite{deng2009imagenet}. Training follows the hyperparameter settings of prior work~\cite{EM}. \added{Detailed image experiment settings are provided in Appendix~\ref{app:image_config}.}
\paragraph{Baselines and Evaluation.}
We compare \(\method\) with four unlearnability methods: Error-Minimum~\cite{EM},  Adversarial Poison~\cite{TAP}, Synthetic Perturbations~\cite{LSP}, and Unlearnable Clusters~\cite{UC}. These methods respectively construct loss-minimizing sample-wise perturbations, adversarial perturbations, synthesized linearly separable perturbations, and cluster-wise perturbations. Synthetic
Perturbations and Unlearnable Clusters construct perturbations shared within classes, we collectively refer to them as class-wise methods. All methods are evaluated using Copyright Protection (\textit{CP}) and Performance Preservation (\textit{PP}).

\subsection{Protection on 2D Medical Images}
\label{sec:MedMNIST2D_standards}

\paragraph{Settings.}
We evaluate \(\method\) on the 2D subsets of MedMNIST~\cite{medmnistv2,medmnistv1}, covering Path~\cite{pathmnist}, Derma~\cite{dermamnist1,dermamnist2}, OCT~\cite{octmnist}, Pneumonia, Retina, Breast~\cite{breastmnist}, Blood~\cite{bloodmnist}, Tissue~\cite{tissuemnist}, OrganA, OrganC, OrganS. We additionally evaluate \(\method\) on the multi-label ChestMNIST dataset~\cite{chestmnist} using independent sigmoid outputs and a multi-label classification objective. ChestMNIST is evaluated separately because its formulation is not directly compatible with the original implementations of the selected unlearnability baselines, and the complete results are provided in Appendix~\ref{app:chestmnist_analysis}. For a consistent evaluation of authorized utility\deleted{, Error-Minimum and Adversarial Poison are implemented using the pseudo labels generated by}\added{ under the label-free inference setting, we additionally consider pseudo-label-based adaptations of the label-dependent baselines. The original Error-Minimum and Adversarial Poison methods are retained as references for evaluating unauthorized performance degradation. However, their label-dependent formulations cannot be directly applied to newly encountered test inputs without access to ground-truth labels. We therefore report adapted variants to examine their behavior under a label-free inference protocol.} 

\added{Specifically, Error-Minimum (Pseudo) represents a pseudo-label-based adaptation of the loss-minimizing baseline, while \(\method\)(Adv), serves as an adversarial-direction ablation of \(\method\). \(\method\) (Adv) shares the same surrogate model, pseudo-labels, perturbation budget, and optimization configuration as \(\method\), but maximizes rather than minimizes the surrogate loss. Accordingly, \(\method\) (Adv) should not be interpreted as an independent reproduction of Adversarial Poison. These adaptations are intended to support controlled comparisons of perturbation objectives rather than establish superiority over the original methods. Synthetic Perturbations and Unlearnable Clusters retain their original formulations.}

\paragraph{Results.}
As shown in Table~\ref{tab:MedMNIST2D},\deleted{the observed mean results for} \(\method\) \added{exhibits an average accuracy-based CP of $-24.65\pm19.01$ percentage points and an average PP of $-0.14\pm0.45$ percentage points across the evaluated 2D datasets. The relatively small mean PP indicates that authorized accuracy is largely preserved on average, although the protection effect varies across datasets. Several conventional unlearnability methods produce more negative CP values, but their authorized-utility preservation cannot always be evaluated under the same label-free inference protocol. These results should therefore be interpreted as descriptive observations of the CP–PP trade-off rather than evidence of statistically established superiority. }\deleted{show a favorable trade-off between Copyright Protection (\textit{CP}) and Performance Preservation (\textit{PP}). On ResNet18, its average PP is only -0.14 ± 0.45, indicating that authorized performance is largely preserved across the evaluated 2D datasets. In contrast, several conventional unlearnability methods achieve stronger degradation of raw-input performance at the cost of substantially reduced authorized utility.}

Some class-wise methods show more negative \textit{CP} than \(\method\) on individual datasets, but their results are less consistent across tasks. Synthetic Perturbations, for example, obtain an average SSIM of 0.87, indicating more visible perturbations, while Unlearnable Clusters exhibit larger cross-dataset variability in both CP and PP. The sample-wise perturbations used by \(\method\) show relatively small authorized-utility changes across several datasets, although the resulting protection remains task-dependent. In particular,as shown in Fig.~\ref{fig:main_2d_comparison}, the relatively small CP values on OrganA, OrganC, and OrganS indicate that the learned models can still generalize from cooked training data to raw inputs when the underlying anatomical features remain highly transferable.

The comparison between \(\method\) and \(\method\) (Adv) further illustrates the CP-PP trade-off. Both variants use the same surrogate model, pseudo-labels, perturbation budget, and optimization configuration, but differ in their loss direction. In the observed results, the adversarial variant shows more negative \textit{CP} on some datasets but also larger \textit{PP} deviations, whereas \(\method\) generally shows smaller authorized-utility changes. Similarly, although Error-Minimum (Pseudo) also uses loss-minimizing perturbations, it relies on an untrained surrogate model, whereas \(\method\) uses a pretrained surrogate to generate more task-aligned fingerprints.

ChestMNIST reveals a metric-dependent limitation. Although \(\method\) produces a large authorization gap under subset accuracy, authorized macro-AUC is 59.54\%, compared with 67.18\% for unauthorized inputs, yielding a negative authorization gap. Macro- and micro-F1 show a similar trend. This indicates that strong accuracy-based protection does not necessarily imply loss of ranking-based predictive information in the multi-label setting. Complete results are provided in Appendix~\ref{app:chestmnist_analysis}.

Overall, these results show that \(\method\) can preserve authorized utility while creating an authorization-dependent performance gap on several 2D medical imaging tasks. However, the protection strength varies across datasets and should be interpreted as a task-dependent trade-off rather than uniform effectiveness.

\begin{table}[!t]
\centering
\scriptsize
\caption{
\textbf{Task-aware AUC-based protection results of DataCook.}
For binary tasks, \textit{CP} and \textit{PP} are computed using AUC;
for multiclass and multilabel tasks, macro-AUC is used.
More negative \textit{CP} indicates a larger reduction in unauthorized
performance, while \textit{PP} closer to zero indicates better
preservation of authorized utility.
Values are mean $\pm$ standard deviation across seeds.
\colorbox{red!20}{Red} indicates settings with a negative
AUC-based authorization gap (\textit{PP}$-\textit{CP}<0$).
}
\label{tab:main_auc_cp_pp}
\setlength{\tabcolsep}{2.8pt}
\renewcommand{\arraystretch}{1.05}
\resizebox{\columnwidth}{!}{%
\begin{tabular}{llcc}
\toprule
Dataset & Task & AUC-\textit{CP} $\downarrow$ & AUC-\textit{PP}$\rightarrow 0$ \\
\midrule
\multicolumn{4}{l}{\textit{MedMNIST 2D}} \\
BloodMNIST
& MultiClass & $-2.50\pm0.55$ & $-1.70\pm0.44$ \\
BreastMNIST
& Binary & $-16.25\pm5.34$ & $-5.17\pm3.02$ \\
ChestMNIST
& MultiLabel &
\colorbox{red!20}{$+1.27\pm0.39$} &
\colorbox{red!20}{$-6.37\pm0.50$} \\
DermaMNIST
& MultiClass &
\colorbox{red!20}{$-5.67\pm2.20$} &
\colorbox{red!20}{$-14.19\pm1.43$} \\
OrganCMNIST
& MultiClass &
\colorbox{red!20}{$-0.27\pm0.27$} &
\colorbox{red!20}{$-1.18\pm0.35$} \\
PneumoniaMNIST
& Binary &
\colorbox{red!20}{$-3.09\pm2.06$} &
\colorbox{red!20}{$-6.97\pm2.07$} \\
RetinaMNIST
& MultiClass &
\colorbox{red!20}{$-4.72\pm4.28$} &
\colorbox{red!20}{$-4.85\pm3.22$} \\
\midrule
\multicolumn{4}{l}{\textit{\textit{MedMNIST 224$\times$224}}} \\
BreastMNIST
& Binary & $-37.36\pm4.40$ & $-9.36\pm3.24$ \\
PneumoniaMNIST
& Binary & $-40.72\pm10.24$ & $-3.30\pm0.48$ \\
RetinaMNIST
& MultiClass & $-18.91\pm2.38$ & $-10.53\pm2.68$ \\
\midrule
\multicolumn{4}{l}{\textit{\textit{MedMNIST3D}}} \\
FractureMNIST3D
& MultiClass
& $-7.83\pm1.60$
& $-2.38\pm0.73$ \\

NoduleMNIST3D
& Binary
& \colorbox{red!20}{$-9.36\pm2.41$}
& \colorbox{red!20}{$-10.09\pm4.29$} \\

OrganMNIST3D
& MultiClass
& \colorbox{red!20}{$-2.65\pm0.63$}
& \colorbox{red!20}{$-2.77\pm0.33$} \\
\midrule
\multicolumn{4}{l}{\textit{\textit{Tabular Medical Data}}} \\
Diabetes
& Binary &
$-10.30\pm4.55$ & $-0.30\pm0.42$ \\
HepatitisC
& MultiClass & $-7.96\pm3.60$ & $-5.18\pm1.65$ \\
SystemicDiseases
& MultiLabel &
$-13.55\pm1.94$ &$-1.01\pm0.07$ \\
\bottomrule
\end{tabular}%
}
\end{table}
\added{
\paragraph{Task-aware AUC Analysis.}
To complement the ACC-based analysis, Table}~\ref{tab:main_auc_cp_pp}
\added{summarizes \textit{CP} and \textit{PP} computed using the task-aware AUC metric. The results reveal substantially more heterogeneous behavior than the ACC-based evaluation. \(\method\) reduces unauthorized AUC in several settings, with particularly large decreases on some high-resolution,  However, the magnitude and direction of the authorization gap vary across datasets and modalities. }

\added{ChestMNIST represents an important failure case. Its AUC-based
\textit{CP} is positive while \textit{PP} is negative, corresponding to
a negative authorization gap on macro-AUC. This indicates that the
current \(\method\) formulation does not consistently suppress
ranking-based predictive performance in the multilabel setting.
Overall, these results suggest that the protection effect is
metric- and task-dependent rather than uniform across evaluation
criteria. Complete task-aware results are provided in
Appendix}~\ref{app:task_aware_metrics}.

\subsection{Robustness Across Model Architectures}

\begin{table}[!t]
\caption{
\textbf{\textit{CP} and \textit{PP} Results Across Network Architectures.}
More negative \textit{CP} indicates stronger protection, while
\textit{PP} closer to zero indicates better performance preservation.
\textbf{Bold} indicates the best result for each dataset, and
\colorbox{red!15}{red} highlights adverse protection outcomes.
}
\label{tab:MedMNIST2D_Full}
\centering
\fontsize{5.2}{5.8}\selectfont
\setlength{\tabcolsep}{0.9pt}
\renewcommand{\arraystretch}{1.02}

\resizebox{\columnwidth}{!}{%
\begin{tabular}{lcccccccc}
\toprule[1pt]
\multirow{2}{*}{Dataset}
& \multicolumn{2}{c}{ResNet18}
& \multicolumn{2}{c}{ResNet50}
& \multicolumn{2}{c}{VGG16}
& \multicolumn{2}{c}{ConvNeXt-Tiny} \\
\cmidrule(lr){2-3}
\cmidrule(lr){4-5}
\cmidrule(lr){6-7}
\cmidrule(lr){8-9}
& \textit{CP}$\downarrow$ & $\textit{PP} \rightarrow 0$
& \textit{CP}$\downarrow$ & $\textit{PP} \rightarrow 0$
& \textit{CP}$\downarrow$ & $\textit{PP} \rightarrow 0$
& \textit{CP}$\downarrow$ & $\textit{PP} \rightarrow 0$ \\
\midrule

Path
& -71.36 & \textbf{0.00}
& -47.19 & 2.51
& -65.23 & 1.03
& \textbf{-76.92} & -0.65 \\

Derma
& \textbf{-25.44} & \textbf{0.00}
& -10.12 & 1.59
& -7.83 & 2.90
& -11.68 & 3.23 \\

OCT
& -43.40 & \textbf{0.01}
& \textbf{-49.60} & -3.99
& -30.40 & -2.79
& -10.30 & -3.29 \\

Pneumonia
& -21.15 & \textbf{0.03}
& \textbf{-26.44} & 0.35
& -16.18 & 0.67
& -8.33 & 1.32 \\

Retina
& -6.00 & -1.57
& -3.75 & \textbf{-0.10}
& \textbf{-45.51} & 3.18
& -2.75 & -4.58 \\

Breast
& -32.05 & \textbf{0.00}
& \textbf{-48.72} & 0.64
& \cellcolor{red!15}{6.00} & 4.48
& -30.13 & \textbf{0.00} \\

Blood
& -23.62 & \textbf{0.00}
& -27.56 & 0.53
& \textbf{-28.70} & 0.93
& -21.25 & 2.30 \\

Tissue
& -28.95 & \textbf{0.00}
& -27.28 & 1.69
& -32.54 & 1.47
& \textbf{-35.11} & 4.59 \\

OrganA
& \textbf{-4.83} & \textbf{0.00}
& -4.55 & 1.00
& -4.06 & \textbf{0.00}
& -4.27 & 4.01 \\

OrganC
& -6.26 & \textbf{0.00}
& -5.89 & 0.87
& -5.98 & 2.23
& \textbf{-10.05} & 3.03 \\

OrganS
& -8.02 & \textbf{0.01}
& -7.73 & 2.39
& -7.28 & 3.40
& \textbf{-8.49} & 5.39 \\

\midrule[1pt]
\rowcolor{gray!12}
Average
& \textbf{-24.65} & \textbf{-0.14}
& -23.53 & 0.68
& -21.61 & 1.59
& -19.94 & 1.39 \\

\bottomrule[1pt]
\end{tabular}%
}
\end{table}

We evaluate whether fingerprints generated using ResNet18 as the surrogate model transfer to downstream architectures with different depths and inductive biases, including ResNet18, ResNet50, VGG16, and ConvNeXt-Tiny.

As shown in Table.~\ref{tab:MedMNIST2D_Full}, \(\method\) generally preserves the CP-PP trade-off across the four evaluated architectures. In several settings, ResNet50 and ConvNeXt-Tiny achieve comparable or improved \textit{PP} relative to ResNet18, suggesting that the learned fingerprints are not restricted to the surrogate architecture.

However, the transferability remains dataset- and architecture dependent. For example, the Breast dataset exhibits weaker \textit{CP} with VGG16, while Retina and OCT show greater variation in \textit{PP} across architectures. These results indicate that \(\method\) can transfer across the evaluated convolutional backbones, but its effectiveness is influenced by the underlying dataset and downstream model.

\subsection{Protection on 3D Medical Images}
\label{sec:MedMNIST3DandMedMNIST+n}
\begin{table*}[!t]
\caption
{
\textbf{Protection Results on 3D and High Resolution 2D Medical Images.}
We report copyright protection (\textit{CP}) and performance preservation
(\textit{PP}) as percentage points. More negative \textit{CP} indicates
stronger protection, while \textit{PP} closer to zero indicates better
utility. \textbf{Bold} highlights best results, and
\colorbox{red!20}{red} indicates adverse effects relative to \(\method\).
Dashes denote unavailable results.
}

\label{tab:highres_3d_combined}

\centering
\scriptsize
\setlength{\tabcolsep}{2.2pt}
\renewcommand{\arraystretch}{1.08}

\resizebox{\textwidth}{!}{%
\begin{tabular}{lcccccccc|cccccc}
\toprule

\multirow{3}{*}{Method}
& \multicolumn{8}{c|}{3D}
& \multicolumn{6}{c}{High Resolution 2D} \\

\cmidrule(lr){2-9}
\cmidrule(lr){10-15}

& \multicolumn{2}{c}{Fracture}
& \multicolumn{2}{c}{Nodule}
& \multicolumn{2}{c}{Organ}
& \multicolumn{2}{c|}{Synapse}
& \multicolumn{2}{c}{Breast}
& \multicolumn{2}{c}{Pneumonia}
& \multicolumn{2}{c}{Retina} \\

& \textit{CP}$\downarrow$ & \textit{PP}$\rightarrow0$
& \textit{CP}$\downarrow$ & \textit{PP}$\rightarrow0$
& \textit{CP}$\downarrow$ & \textit{PP}$\rightarrow0$
& \textit{CP}$\downarrow$ & \textit{PP}$\rightarrow0$
& \textit{CP}$\downarrow$ & \textit{PP}$\rightarrow0$
& \textit{CP}$\downarrow$ & \textit{PP}$\rightarrow0$
& \textit{CP}$\downarrow$ & \textit{PP}$\rightarrow0$ \\

\midrule

Error-Minimum~\cite{EM}
& -5.83 & --
& -6.77 & --
& -3.28 & --
& -4.83 & --
& -13.46 & --
& -29.81 & --
& -43.75 & -- \\

Adversarial Poison~\cite{TAP}
& -10.42 & --
& -38.39 & --
& -5.74 & --
& 1.98 & --
& -13.46 & --
& -29.81 & --
& -16.50 & -- \\

Synthetic Perturbations~\cite{LSP}
& -18.75 & --
& -14.84 & --
& -14.59 & --
& -31.53 & --
& -3.86 & --
& -2.25 & --
& -14.75 & -- \\

\midrule

Unlearnable Clusters~\cite{UC}
& \cellcolor{red!20}{-1.25}
& \cellcolor{red!20}{-1.67}
& \cellcolor{red!20}{-6.77}
& \cellcolor{red!20}{-1.61}
& \cellcolor{red!20}{-6.89}
& \textbf{3.44}
& \cellcolor{red!20}{2.83}
& \textbf{3.13}
& -13.46 & 0
& -29.81 & \cellcolor{red!20}{-2.25}
& -23.00 & \textbf{0.25} \\

Error-Minimum (Pseudo)
& \cellcolor{red!20}{-5.00}
& 10.42
& -66.45
& \cellcolor{red!20}{-0.66}
& -8.20
& \cellcolor{red!20}{-9.18}
& \textbf{-18.75}
& 1.42
& \cellcolor{red!20}{0}
& \cellcolor{red!20}{-0.65}
& \cellcolor{red!20}{-9.94}
& \cellcolor{red!20}{-3.05}
& \cellcolor{red!20}{-12.75}
& \cellcolor{red!20}{-4.50} \\

\midrule

\(\method\) (Adv)
& \cellcolor{red!20}{-16.67} & 0
& -37.42 & 0
& \cellcolor{red!20}{-5.41} & 0.98
& \cellcolor{red!20}{1.14} & 1.14
& -13.46 & 0
& -29.81 & 0
& -54.25 & \cellcolor{red!20}{-0.50} \\

\(\method\)
& \textbf{-26.67} & \textbf{0}
& -23.23 & \textbf{0}
& \textbf{-10.82} & 0.16
& -9.94 & 1.70
& -13.46 & \textbf{0}
& -29.81 & \textbf{0}
& -26.00 & 0 \\

\bottomrule
\end{tabular}%
}
\end{table*}
\paragraph{Settings.}
We extend the evaluation to four volumetric MedMNIST subsets: Fracture~\cite{jin2020deep}, Organ~\cite{organmnist1}, Nodule~\cite{samuel2011lung}, and Synapse~\cite{medmnistv2}. These experiments examine whether the CP–PP trade-off observed on 2D images can be maintained for higher-dimensional inputs.
\paragraph{Results.}
The 3D section of Table~\ref{tab:highres_3d_combined} presents the results on 3D datasets, which align with the trends observed in the 2D experiments, albeit with some notable differences.

First, the observed mean results for Anti-Adversarial (Anti-Adv) fingerprints show a \deleted{relatively stable} trade-off between \textit{CP} and \textit{PP} across 3D datasets. Second, \added{the relative performance of \(\method\) and \(\method\) (Adv) varies across datasets. \(\method\) achieves more negative CP on FractureMNIST3D and OrganMNIST3D, whereas \(\method\) (Adv) achieves
more negative CP on NoduleMNIST3D. These results indicate dataset-dependent differences rather than consistent superiority of either perturbation objective.} \deleted{although adversarial (Adv) examples achieve comparable and, in some cases, slightly better performance on the Fracture and Nodule datasets, they exhibit a marked decline in \textit{CP} on the Organ and Synapse datasets. This highlights their inconsistency in 3D domains and suggests that classical adversarial perturbations may be less effective for volumetric data. Consequently, Anti-Adv examples offer better protection for 3D applications.} Third,
\added{Unlearnable Clusters exhibits limited protection on several 3D datasets. Some label-dependent baselines achieve stronger accuracy-based degradation on individual datasets, but their relative performance is not consistent across the evaluated tasks.} \added{Finally, the varying performance of Error-Minimum (Pseudo) highlights the sensitivity of perturbation-based protection to the selected configuration. Overall, the evaluated results do not establish statistically superior or uniformly more stable protection on 3D data.}
\deleted{the Unlearnable Clusters method performs poorly across all 3D datasets, indicating a lack of transferability from 2D to 3D. This may stem from the increased complexity of defining effective class-wise perturbations in three-dimensional feature space. In contrast, label-dependent methods such as Error-Minimum and Adversarial Poison demonstrate relatively stronger performance, suggesting that label-driven perturbation strategies are more adaptable to 3D settings—likely due to their reliance on sample-wise gradients rather than global class structure. 
Finally, the inconsistent performance of Error-Minimum (Pseudo) further highlights the challenge of generating stable unlearnable perturbations without pretrained knowledge. In comparison, perturbations generated using pretrained models—as in datacook—exhibit significantly improved stability and protection performance on 3D data.}

\added{The AUC-based results in Table~\ref{tab:main_auc_cp_pp} further reveal task-dependent behavior. FractureMNIST3D exhibits a positive authorization gap of 5.45 percentage points, whereas NoduleMNIST3D and OrganMNIST3D show slightly negative gaps of -0.73 and -0.12 points, respectively. Thus, accuracy-based protection does not consistently translate into AUC-based authorization differences in volumetric tasks.}

In summary, the 3D experiments \added{ illustrate } that \(\method\) can extend to volumetric medical data while preserving authorized utility and
creating an authorization-dependent performance gap in several
datasets. However, both the magnitude and consistency of this effect
remain task-dependent.

\subsection{Protection on High-Resolution Medical Images}
\label{sec:MedMNISTHighResolution}

\paragraph{Settings.}
We evaluate \(\method\) on the $224 \times 224$ versions of Pneumonia, Retina, and Breast~\cite{breastmnist}, each containing approximately 1,500–2,000 images. These experiments examine how increasing input dimensionality affects the CP–PP trade-off of perturbation-based protection methods.

\paragraph{Results.}
As shown in high resolution 2D section of Table~\ref{tab:highres_3d_combined}, several methods produce similar accuracy-based \textit{CP} values under the original perturbation budget. This convergence is related to the use of a fixed perturbation constraint in a higher-dimensional input space, together with the discrete nature of accuracy-based evaluation.

\added{The AUC-based results in Table~\ref{tab:main_auc_cp_pp} provide further differentiation. BreastMNIST, PneumoniaMNIST, and RetinaMNIST exhibit unauthorized AUC reductions of 37.36, 40.72, and 18.91\%, respectively, exceeding their corresponding authorized reductions. These results reveal authorization-dependent ranking differences not fully captured by accuracy alone.}

Overall, the observed results show relatively small
authorized-utility changes for \(\method\), together with substantial raw-input performance degradation on several high-resolution datasets. These results indicate that the proposed cooking mechanism can be applied to the evaluated high-resolution medical images, although its performance remains dataset-dependent.

\subsection{ Protection on Structured Tabular Data}
\label{sec:Non-image}
















\begin{table}[!t]
\caption
{
\textbf{Protection Results on Structured Healthcare Datasets.}
We report \textit{CP} and \textit{PP} in percentage points. More negative \textit{CP} indicates
stronger protection, while \textit{PP} closer to zero indicates better
utility. \textbf{Bold} marks best label-free results, and
\colorbox{red!20}{red}  highlights weaker results.
}
\label{tab:MedMNISTNonImage}

\centering
\scriptsize
\setlength{\tabcolsep}{3.2pt}
\renewcommand{\arraystretch}{1.08}

\begin{tabular}{@{}llccc@{}}
\toprule
Dataset & Method & Label
& \textit{CP}$\downarrow$
& \textit{PP}$\rightarrow 0$ \\
\midrule

\multirow{4}{*}{Pima}
& Adversarial
& W/  & -31.82 & 22.08 \\
& Anti-Adversarial
& W/  & -21.43 & 22.08 \\
& \(\method\) (Adv)
& W/O & \cellcolor{red!20}{-9.74} & \textbf{0} \\
& \(\method\)
& W/O & \textbf{-14.37} & \cellcolor{red!20}{-0.08} \\

\midrule

\multirow{4}{*}{Hepatitis C}
& Adversarial
& W/  & -72.36 & -1.63 \\
& Anti-Adversarial
& W/  & -30.89 & 8.13 \\
& \(\method\) (Adv)
& W/O & \cellcolor{red!20}{-7.48}
& \cellcolor{red!20}{-2.44} \\
& \(\method\)
& W/O & \textbf{-36.28} & \textbf{0.00} \\

\midrule

\multirow{4}{*}{\shortstack{Systemic\\Diseases}}
& Adversarial
& W/  & -85.64 & -1.21 \\
& Anti-Adversarial
& W/  & -39.03 & 0.72 \\
& \(\method\) (Adv)
& W/O & \cellcolor{red!20}{-23.53}
& \cellcolor{red!20}{-1.79} \\
& \(\method\)
& W/O & \textbf{-32.49}& \textbf{-0.48} \\

\bottomrule
\end{tabular}
\end{table}

\paragraph{Settings.}
We extend the evaluation of \(\method\) beyond medical imaging to three structured healthcare datasets representing different classification settings. The Pima Indians Diabetes dataset~\cite{smith1988using} contains 768 samples for binary diabetes prediction, while the Hepatitis C Prediction dataset~\cite{hcv_data_571} contains 615 samples for five-class liver-condition classification. The Systemic Diseases dataset~\cite{sudlow2015uk} includes 61,605 samples and supports multi-label prediction of chronic diseases from baseline clinical and demographic variables.

All experiments use the same MLP architecture and consistent training configurations, with complete dataset and implementation details provided in Appendix~\ref{appendix:tabular-details}. We compare two label-free variants, \(\method\) and \(\method\)~(Adv), along with adversarial and anti-adversarial oracle references that use ground-truth labels during perturbation generation. Image-specific baselines whose original formulations do not naturally transfer to structured tabular inputs are excluded from this comparison.

\paragraph{Results.}
Table~\ref{tab:MedMNISTNonImage} summarizes the results. The oracle variants provide reference results under direct access to ground-truth labels, whereas \(\method\) and \(\method\)~(Adv) generate perturbations using surrogate-model predictions without label access.

Across the three evaluated tabular datasets, \(\method\) achieves more negative accuracy-based CP than \(\method\) (Adv). It also exhibits smaller absolute PP deviations on HepatitisC and SystemicDiseases, while \(\method\) (Adv) preserves the clean accuracy more closely on Diabetes. These observations are specific to the evaluated configurations and do not establish general superiority across tasks.

\added{The AUC-based results in Table~\ref{tab:main_auc_cp_pp} further demonstrate authorization-dependent behavior. At $\epsilon=5$, Diabetes, HepatitisC, and SystemicDiseases exhibit unauthorized AUC reductions of 10.30, 7.96, and 13.55\%, respectively. However,
authorized AUC also decreases by 5.18 points on HepatitisC, indicating that performance preservation remains task-dependent.}

The effectiveness of both variants remains dependent on the task and evaluation metric. In particular, the large-scale multi-label Systemic Diseases dataset presents a more challenging setting for maintaining a clear authorization gap. Overall, these experiments demonstrate that the cooking formulation can be applied to structured healthcare data, while also indicating that modality-specific optimization may be necessary to strengthen protection on complex tabular tasks.

\deleted{
To complement the ACC-based analysis, Table}
\deleted{summarizes \textit{CP} and \textit{PP} computed using the task-aware
AUC metric. The results reveal substantially more heterogeneous behavior
than the ACC-based evaluation. datacook reduces unauthorized AUC in
several settings, with particularly large decreases on some
high-resolution and 3D datasets, whereas the separation is limited on
several low-resolution and structured tabular tasks. }
\deleted{
ChestMNIST represents an important failure case. Its AUC-based
\textit{CP} is positive while \textit{PP} is negative, corresponding to
a negative authorization gap on macro-AUC. This indicates that the
current datacook formulation does not consistently suppress
ranking-based predictive performance in the multilabel setting.
Overall, these results suggest that the protection effect is
metric- and task-dependent rather than uniform across evaluation
criteria. Complete task-aware results are provided in
Appendix}

\subsection{Robustness to Adaptive Attacks}
\paragraph{Settings.}
We consider two groups of adaptive attacks. Input-level attacks include JPEG compression, Gaussian smoothing, median filtering, and augmentation-based transformations, which attempt to weaken the fingerprints. Training-time attacks include ordinary fine-tuning, robust fine-tuning, and fine-tuning with a limited number of labeled raw samples. Given the computational cost of evaluating every attack-dataset combination, we select representative \(28\times28\) MedMNIST datasets covering binary, multiclass, and multilabel classification tasks, as well as different protection levels. Both input-level and training-time attacks are evaluated on this subset. For the \(224\times224\) datasets, we focus on input transformations and augmentation-based attacks.

For each dataset and perturbation budget,
Table~\ref{tab:adaptive_attack_strongest} reports the strongest observed
attack according to the highest post-attack task-aware primary metric
within the largest common seed cohort. Binary tasks use AUC, whereas
multiclass and multilabel tasks use macro-AUC. We define
\[
\text{Attack Gain}
=
\text{After Attack}
-
\text{Unauthorized Before Attack},
\]
where positive values indicate recovery of unauthorized performance.
Complete results are provided in Appendix~\ref{app:Adaptive}.

\begin{table*}[!t]
\scriptsize
\setlength{\tabcolsep}{3.2pt}
\caption{\textbf{Strongest Observed Adaptive Attack per Dataset and Budget.} For each dataset, the attack with the highest post-attack task-aware primary metric is selected within the largest available common seed cohort. Binary tasks use AUC; multi-class and multi-label tasks use macro-AUC. Before/After are unauthorized performance, Gain is After minus Before, and CP after is After minus Clean, all in percentage points. Recovery is the clipped seed-wise fraction of the clean-to-unauthorized gap recovered by the attacker. Selection is descriptive and is not a significance claim.}
\label{tab:adaptive_attack_strongest}
\centering
\setlength{\tabcolsep}{3.0pt}
\renewcommand{\arraystretch}{1.08}
\resizebox{\textwidth}{!}{%
\begin{tabular}{llrrlccccc}
\toprule[1pt]\midrule[0.3pt]
Dataset & Input & \(\epsilon\) & \(n\) & Strongest attack & Before & After & Gain & CP after & Recovery (\%) \\
\midrule[0.3pt]
BloodMNIST & 28\(\times\)28 & 8 & 8 & FT-10 & \(97.3\pm0.5\) & \(98.9\pm0.3\) & \(1.7\pm0.7\) & \(-0.8\pm0.3\) & \(64.3\pm18.8\) \\
BreastMNIST & 28\(\times\)28 & 8 & 8 & FT-10 & \(72.5\pm4.2\) & \(76.9\pm3.0\) & \(4.5\pm2.5\) & \(-11.8\pm4.2\) & \(26.7\pm15.2\) \\
ChestMNIST & 28\(\times\)28 & 8 & 3 & RFT-10 & \(67.2\pm0.5\) & \(70.1\pm1.7\) & \(2.9\pm1.3\) & \(4.2\pm1.6\) & \(0.0\pm0.0\) \\
DermaMNIST & 28\(\times\)28 & 8 & 8 & FT-10 & \(85.5\pm2.2\) & \(87.3\pm1.1\) & \(1.8\pm2.0\) & \(-3.9\pm1.2\) & \(30.2\pm20.5\) \\
OrganCMNIST & 28\(\times\)28 & 8 & 8 & JPEG (Q=95) & \(99.2\pm0.2\) & \(99.2\pm0.2\) & \(-0.0\pm0.0\) & \(-0.3\pm0.3\) & \(0.2\pm0.4\) \\
PneumoniaMNIST & 28\(\times\)28 & 8 & 8 & FT-10 & \(93.5\pm2.4\) & \(94.6\pm0.7\) & \(1.1\pm2.0\) & \(-2.0\pm0.7\) & \(34.9\pm29.8\) \\
RetinaMNIST & 28\(\times\)28 & 8 & 8 & FT-10 & \(63.5\pm3.9\) & \(69.6\pm2.3\) & \(6.2\pm4.7\) & \(1.4\pm3.1\) & \(76.7\pm40.0\) \\
RetinaMNIST & 28\(\times\)28 & 12 & 8 & FT-10 & \(57.5\pm3.7\) & \(66.7\pm2.4\) & \(9.2\pm3.9\) & \(-1.5\pm3.2\) & \(77.8\pm21.3\) \\
\midrule[0.3pt]
BreastMNIST & 224\(\times\)224 & 8 & 3 & Gaussian ($\sigma=2$) & \(53.3\pm4.5\) & \(71.4\pm4.9\) & \(18.1\pm9.2\) & \(-19.3\pm5.4\) & \(47.2\pm18.6\) \\
PneumoniaMNIST & 224\(\times\)224 & 8 & 3 & Augmentation (L=2) & \(57.9\pm10.2\) & \(57.5\pm4.2\) & \(-0.3\pm8.1\) & \(-41.0\pm4.2\) & \(6.4\pm5.6\) \\
RetinaMNIST & 224\(\times\)224 & 8 & 3 & JPEG (Q=95) & \(64.0\pm2.3\) & \(65.1\pm1.7\) & \(1.1\pm0.7\) & \(-17.8\pm1.8\) & \(5.6\pm2.6\) \\
\midrule[0.3pt]\bottomrule[1pt]
\end{tabular}%
}
\end{table*}

\paragraph{Results.}
As shown in Table~\ref{tab:adaptive_attack_strongest}, attack effectiveness varies across datasets. On most \(28\times28\) datasets, 10-epoch ordinary fine-tuning is the strongest evaluated attack, but, the gains  on BloodMNIST, DermaMNIST, BreastMNIST, and PneumoniaMNIST remain between \(1.1\) and \(4.5\) percentage points. RetinaMNIST is more susceptible to model adaptation. Fine-tuning increases unauthorized performance by \(6.2\) and \(9.2\) percentage points at \(\epsilon=8\) and \(\epsilon=12\), recovering approximately \(76.7\%\) and \(77.8\%\) of the original performance gap. In contrast, JPEG compression produces almost no recovery on OrganCMNIST. On ChestMNIST, robust fine-tuning increases macro-AUC by \(2.9\) percentage points.

For the high-resolution datasets, Gaussian smoothing increases unauthorized performance on BreastMNIST-224 by \(18.1\) percentage points, although the post-attack AUC remains \(19.3\) percentage points below the clean reference. The strongest evaluated attacks on PneumoniaMNIST-224 and RetinaMNIST-224 yield gains of only \(-0.3\) and \(1.1\) percentage points.

Overall, adaptive attacks partially recover unauthorized performance in selected settings, particularly RetinaMNIST under fine-tuning and BreastMNIST-224 under Gaussian smoothing, while recovery remains limited on several other datasets.
\section{Discussion}\label{sec4}

\paragraph{Principal findings.}
We present \method{} as a deployment-stage usage-control framework based on sample-wise anti-adversarial fingerprints. Unlike conventional unlearnable examples~\cite{EM,LSP,UC,TAP}, which primarily aim to prevent effective model training, \method{} seeks to make downstream model utility dependent on access to a private cooking procedure. It uses surrogate-model predictions as pseudo-labels and therefore does not require ground-truth labels during fingerprint generation. Pseudo-label agreement and confidence are generally high but do not fully predict downstream protection strength. Detailed diagnostics are provided in Appendix~\ref{app:pseudo_quality}.

Across the evaluated datasets, \method{} preserves near-baseline authorized utility in several settings while reducing performance on unauthorized raw inputs. However, the protection effect varies across datasets, modalities, and evaluation metrics. The current results therefore support task-dependent authorization control rather than uniform protection across all healthcare data.

\paragraph{Protection-utility trade-off.}
\deleted{Compared with the selected baselines, }\added{The evaluated results illustrate different empirical trade-offs between copyright protection (\textit{CP}) and performance preservation (\textit{PP}) across perturbation-based methods.} \method{}\deleted{shows a favorable empirical balance between copyright protection (\textit{CP}) and performance preservation (\textit{PP}) across several evaluated settings. However, the magnitude of this advantage varies across datasets, and the corresponding between-method differences are not statistically significant after Holm correction} \added{exhibits relatively small authorized-accuracy changes in several settings, while some baselines produce greater unauthorized-performance degradation. However, these patterns vary across datasets as reported in Appendix~\ref{app:statistics}. Consequently, the observed differences are interpreted descriptively and should not be taken as evidence of statistically established superiority.} \deleted{ Compared with class-wise methods such as Unlearnable Clusters and Synthetic Perturbations, its sample-wise construction avoids assigning one shared perturbation pattern to all samples of the same class.}

The comparison between \method{} and \method{} (Adv) further clarifies this trade-off. Both variants use the same surrogate model, pseudo-labels, perturbation constraint, PGD configuration, and downstream training protocol, while differing mainly in optimization direction. \method{} (Adv) maximizes the surrogate loss and more aggressively disrupts predictive features, often producing stronger \textit{CP} but weaker \textit{PP}. In contrast, \method{} minimizes the loss toward the surrogate prediction, which better preserves authorized utility but may yield weaker protection on some tasks. Therefore, the observed difference cannot be explained solely by perturbation magnitude.

\paragraph{Dataset- and resolution-dependent behavior.}
The large cross-dataset variation in \textit{CP} reflects task heterogeneity rather than only run-to-run instability. For OrganA, OrganC, and OrganS, models trained on cooked data still generalize relatively well to raw inputs, leading to weaker protection. A plausible explanation is that the dominant anatomical features in these tasks remain discriminative under the bounded perturbation budget, although the current experiments do not establish this mechanism causally.

A related phenomenon appears in the high-resolution experiments. Under the fixed perturbation budget, several methods produce identical accuracy-based \textit{CP} values because they correctly classify the same number of unauthorized test samples. The cross-modality ablation in Appendix~\ref{app:Perturbation-budget} further shows that the CP-PP trade-off is sensitive to the selected budget on the $28\times28$, 3D, and tabular settings

\paragraph{Adaptive attacks and threat-model boundaries.}
We further evaluate input transformations, data augmentation, ordinary and robust fine-tuning, and limited-data model adaptation. These attacks partially restore unauthorized performance on some datasets but provide limited recovery on others. The results therefore indicate task-dependent resilience under the evaluated attacks rather than general robustness against all adaptive or gray-box adversaries.

Our threat model assumes that unauthorized users may access the cooked training data and labels and may train or adapt arbitrary downstream models, but do not possess the private surrogate model or complete cooking parameters. Each dataset uses an independently trained surrogate and separately generated fingerprints. However, complete leakage of the corresponding surrogate and cooking configuration could weaken protection, and the current implementation does not provide formal key rotation or revocation.

\paragraph{Image quality and clinical relevance.}


\begin{figure*}[t]
    \centering
	  \includegraphics[width=0.75\textwidth]{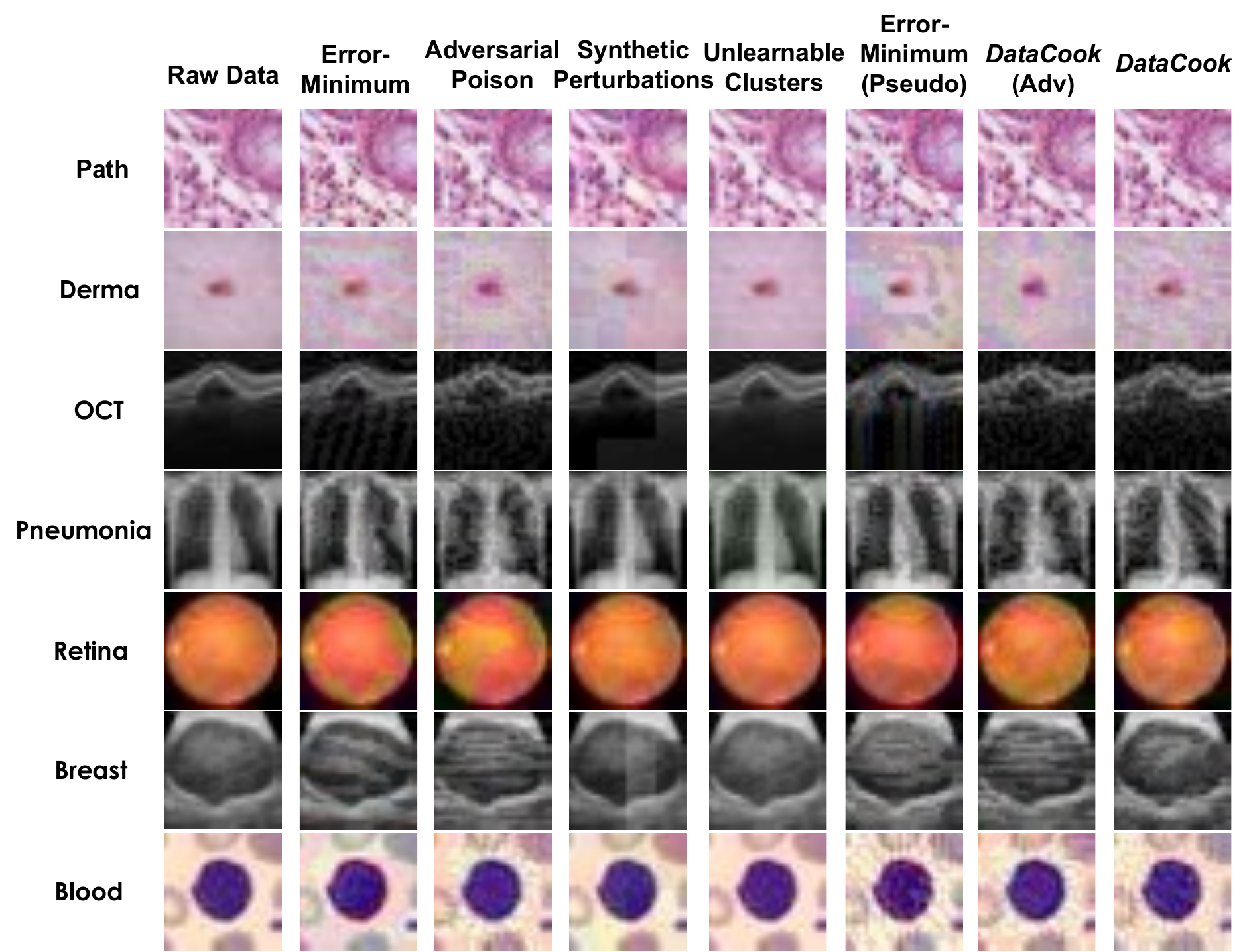}
	\caption{ \textbf{Perturbations Comparison.} This figure compares the visual effects of perturbations introduced by different data protection methods. It highlights the perceptibility and distribution patterns of the perturbations across various approaches.}
	\label{fig:Perturbation}
\end{figure*}

As shown in Fig.~\ref{fig:Perturbation}, different methods exhibit distinct perturbation patterns. Class-wise methods such as UC~\cite{UC} and SP~\cite{LSP} may introduce spatially structured artifacts, whereas sample-wise approaches generate image-specific perturbations. In the evaluated examples, \method{} generally produces more smoothly distributed and less visually apparent perturbations than several baselines. We additionally report SSIM, PSNR, LPIPS, and perturbation-norm measurements, where applicable, in Appendix~\ref{app:Image}. However, neither the qualitative comparison in Fig.~\ref{fig:Perturbation} nor quantitative similarity metrics establish diagnostic equivalence or clinical safety. Quantitative imaging features, including radiomic features, can contribute to clinically relevant prediction and treatment-response assessment
~\cite{zwanenburg2020ibsi, zhang2024defining}.Therefore, preserving
global image similarity does not necessarily establish the stability of
clinically relevant imaging features. Importantly, the current study does not include blinded radiologist assessment, lesion-level evaluation, or systematic radiomics-stability analysis. Therefore, the reported image-similarity metrics should not be interpreted as evidence of diagnostic equivalence or clinical safety. Establishing preservation of clinically relevant information will require dedicated expert evaluation in future work.

\paragraph{Computational considerations.}
Runtime, online cooking latency, protected-training cost, and GPU-memory
measurements are reported in Appendix~\ref{app:runtime}. The reported
values are hardware- and implementation-dependent and are interpreted
as descriptive system measurements.

\paragraph{Deployment limitations.}

A potential application of \method{} is multicenter radiology data sharing, in which a data-owning institution distributes cooked images while retaining control of the private cooking procedure. Multicenter
medical-imaging studies increasingly require standardized data and
learning pipelines across institutions~\cite{sui2025mri}. However, this scenario has not been validated in real DICOM transmission or PACS workflows. Scanner, protocol, reconstruction, and population differences may also affect both model generalization and protection performance across institutions~\cite{zech2018variable}. More broadly, clinical translation of deep-learning systems in radiology remains subject to modality-specific technical and workflow challenges~\cite{wang2025application}. These factors are not modeled in the
current proof-of-concept evaluation.

\method{} should not be interpreted as a standalone legal copyright enforcement or privacy mechanism. It does not establish ownership, prove infringement, identify unauthorized users, or provide a formal privacy guarantee. Federated learning~\cite{sheller2020federated}, differential privacy~\cite{abadi2016deep}, and homomorphic encryption~\cite{giladbachrach2016cryptonets} address different stages and threat models and may serve as complementary components.

\paragraph{Limitations and future work.}
The current evidence is limited by task-dependent protection, incomplete attack coverage for 3D and structured data, reliance on private cooking parameters, lack of clinical validation, radiologist-based and lesion-level clinical validation, and the computational cost of iterative sample-wise PGD for full-resolution studies. Future work should improve fingerprint revocability, robustness to model adaptation, scalability to complete clinical studies, and effectiveness on heterogeneous and multilabel tasks.

\section{Conclusion}\label{sec5}
In conclusion, we introduce \method{}, a deployment-stage framework using anti-adversarial fingerprints without ground-truth labels to make model utility depend on a private cooking procedure. Across healthcare datasets, \method{} preserves authorized utility and creates
authorization gaps in several tasks, although effectiveness remains task-dependent. These results demonstrate feasibility rather than uniform protection, privacy guarantees, legal enforcement, or clinical readiness. Future work will improve robustness, scalability, and support for heterogeneous and multilabel data.

\printcredits

\section*{RESOURCE AVAILABILITY}

\subsection*{Data Availability}

The MedMNIST datasets used in this study are publicly available through
the MedMNIST project~\cite{medmnistv2,medmnistv1} at
\url{https://medmnist.com} and
\url{https://github.com/MedMNIST/MedMNIST}.

The structured healthcare datasets used in this study, including the
Pima Indians Diabetes dataset~\cite{smith1988using}, the Hepatitis C
dataset~\cite{hcv_data_571}, and the Systemic Diseases dataset derived
from UK Biobank~\cite{sudlow2015uk}, are also publicly available from
their respective data repositories, subject to the original access
conditions and licenses.

\subsection*{Code Availability}

The code is publicly available at \url{https://github.com/HINTLab/DataCook}.

\subsection*{Additional Information}

Additional information required to reproduce or reanalyze the results
reported in this study is available from the lead contact upon
reasonable request.

\subsection*{Ethics Statement}

This study exclusively uses publicly available, de-identified datasets
and does not involve the collection of new human-participant data.
Therefore, no additional institutional ethics approval or informed
consent was required for this study.

\subsection*{Declaration of generative AI and AI-assisted technologies in the manuscript preparation process}

During the preparation of this work, the authors used ChatGPT (OpenAI) solely for language polishing and assistance with improving the organization and presentation of the manuscript. The authors reviewed and edited all AI-assisted output and take full responsibility for the content of the published article.

\section*{Funding}

This study was supported in part by ELLIS Institute Finland and the School of Electrical Engineering, Aalto University. This work was also supported in part by the Research Council of Finland through Academy Project 376209.
The work of Wei Zhao was supported in part by the National Natural
Science Foundation of China (NSFC) under Grants 62476291 and 82672638,
the Hunan Provincial Natural Science Foundation for Distinguished Young
Scholars under Grant 2025JJ20097, the Research Foundation of Education
Bureau of Hunan Province under Grant 24B0003, the Health Commission of
Hunan Province under Key Research Project Grant 20254638, and the
Medical Discipline Construction Project of Hunan Province.

\section*{Declaration of Competing Interest}

Wei Zhao is affiliated with the editorial office of \emph{Meta-Radiology}. He was not involved in the editorial review or the decision to publish this article. Apart from this disclosed editorial affiliation, the authors declare that they have no other known competing financial interests or personal relationships that could have appeared to influence the work reported in this paper.

\bibliographystyle{cas-model2-names}
\bibliography{cas-refs}

\appendix


\section{Tabular Dataset Details and Implementation}
\label{appendix:tabular-details}

\subsection{MLP Architecture}
For all tabular tasks, we use the same multilayer perceptron (MLP) architecture as the downstream predictive model. Given an input feature vector with dimension (d), the model first maps the input to a 64-dimensional hidden representation, followed by a ReLU activation and a dropout layer with a dropout rate of 0.3. The resulting representation is then projected to a 32-dimensional hidden layer, again followed by ReLU activation and dropout with the same rate. Finally, a linear output layer maps the 32-dimensional representation to the task-specific output dimension. The same architecture and hyperparameter settings are used across all tabular datasets to ensure a consistent evaluation protocol.

\subsection{Dataset Descriptions}
To evaluate the applicability of \(\method\) beyond image-based healthcare data, we further conduct experiments on three representative non-image tabular healthcare datasets. As summarized in Table~\ref{tab:tabular-summary}, these datasets cover different prediction settings, including binary classification, multi-class classification, and multi-label classification. They also vary substantially in sample size and feature composition, allowing us to examine whether \(\method\) can provide effective protection while preserving downstream utility across heterogeneous clinical tabular data.

\begin{table*}[!t]
\centering
\caption{Summary of non-image tabular datasets.}
\label{tab:tabular-summary}
\begin{tabular}{lcccp{6.2cm}}
\toprule[1pt]\midrule[0.3pt]
\textbf{Dataset} & \textbf{\#Samples} & \textbf{Task} & \textbf{\#Labels} &  \multicolumn{1}{c}{\textbf{Main Features}}  \\ \midrule[1pt]
\textbf{Pima Diabetes} & 768 & Binary & 2 & Pregnancies, Glucose, BloodPressure, BMI, Insulin, Age, etc. \\ \midrule
\textbf{Hepatitis C} & 615 & Multi-Class & 5 & Age, Sex, ALB, ALP, ALT, AST, BIL, CHE, CHOL, CREA, GGT, PROT \\ \midrule
\textbf{Systemic Diseases} & 61,605 & Multi-Label & 5 & Age, Gender, Ethnic, BMI, Cholesterol, BP, Diabetes, Hypertension, Smoking, Follow-time \\ 
\midrule[0.3pt]\bottomrule[1pt]
\end{tabular}
\end{table*}

\subsection{Training and \deleted{Attack}\added{Perturbation} Configurations}
\added{For all structured tabular datasets, the models are trained for 100 epochs using the Adam optimizer with a learning rate of $10^{-3}$ and a batch size of 32. Cross-entropy loss is used for Diabetes and HepatitisC, whereas BCEWithLogits loss is adopted for the multilabel
SystemicDiseases task.}

\added{Fingerprint generation is performed in the standardized feature space under an $L_\infty$ perturbation constraint. For Diabetes and HepatitisC, we use 300 projected-gradient steps with a step size of 0.05. For SystemicDiseases, we use 80 steps with a step size of $\epsilon/4$.  The perturbation budgets are $\epsilon\in\{0.1,0.25,0.5,1,2,5\}$, with $\epsilon=5$ used as the representative setting.}

\added{The extended experiments are conducted with eight random seeds (0--7), and the results are reported as mean and standard deviation across seeds.}

\deleted{For all tabular datasets, training is performed using the Adam optimizer with a batch size of 32. We use cross-entropy loss for binary or multi-class tasks, and binary cross-entropy with Logits loss for multi-label classification. Adversarial attacks are implemented using a PGD-style iterative method with the following settings:}
    \deleted{Maximum perturbation: $\epsilon = 5$}
    \deleted{ Step size: $\alpha = 0.05$}
    \deleted{ Number of steps: $300$}
\deleted{These settings are consistent across all datasets and evaluation protocols.}

\section{Image Training and Fingerprint Generation Configurations}
\label{app:image_config}

\added{The detailed configurations for image-based model training
and fingerprint generation are summarized in Table}~\ref{tab:image_hyperparameters}.

\begin{table*}[t]
\centering
\caption{Image Training and Fingerprint Generation Configurations.
Training and perturbation-generation settings are reported
for the three image settings.}
\label{tab:image_hyperparameters}

\small
\setlength{\tabcolsep}{8pt}
\renewcommand{\arraystretch}{1.12}

\begin{tabular}{lccc}
\toprule
\textbf{Parameter} &
\textbf{2D (28$\times$28)} &
\textbf{2D (224$\times$224)} &
\textbf{3D} \\
\midrule

\multicolumn{4}{l}{\textbf{Model Training}} \\
\addlinespace[2pt]

Surrogate training epochs
    & 100 & 100 & 100 \\
Protected-model training epochs
    & 100 & 100 & 100 \\
Optimizer
    & SGD & SGD & SGD \\
Initial learning rate
    & $0.1$ & $10^{-3}$ & $10^{-3}$ \\
Momentum
    & $0.9$ & $0.9$ & $0.9$ \\
Weight decay
    & $5\times10^{-4}$ & $5\times10^{-4}$ & $5\times10^{-4}$ \\
LR scheduler
    & CosineAnnealingLR & CosineAnnealingLR & CosineAnnealingLR \\
Minimum learning rate
    & 0 & 0 & 0 \\
Protected-model train batch
    & 128 & 32 & 32 \\
Protected-model eval batch
    & 128 & 32 & 32 \\

\midrule
\multicolumn{4}{l}{\textbf{Fingerprint Generation}} \\
\addlinespace[2pt]

Perturbation budget $\epsilon$
    & $8/255$ & $8/255$ & $32/255$ \\
Projected-gradient steps
    & 20 & 20 & 20 \\
Step size $\alpha$
    & $1.6/255$ & $1.6/255$ & $2/255$ \\
Perturbation-search batch
    & 128 & 32 & 32 \\
Perturbation-search eval batch
    & 128 & 32 & 32 \\

\midrule
\multicolumn{4}{l}{\textbf{Implementation Details}} \\
\addlinespace[2pt]

Default random seed
    & 42 & 42 & 42 \\
Loss function
    & Cross-entropy & Cross-entropy & Cross-entropy \\

\bottomrule
\end{tabular}

\vspace{2pt}
\begin{minipage}{0.96\textwidth}

\end{minipage}

\end{table*}
\section{Complete Task-aware Evaluation Metrics}
\label{app:task_aware_metrics}
Table~\ref{tab:complete_task_aware_metrics} shows that the effectiveness of \(\method\) varies substantially across modalities, datasets, and evaluation metrics. The strongest aggregate protection is observed on the 224$\times$224 datasets, where the mean CP reaches $-32.33$ percentage points for AUC and $-43.46$ percentage points for Macro-F1. However, the corresponding AUC PP is $-7.73$ percentage points, despite near-zero PP under Task Accuracy, Balanced Accuracy, and Macro-F1. This discrepancy indicates that \(\method\) may preserve authorized predictions at a fixed decision threshold while still altering the ranking or confidence structure of the predicted probabilities. A similar, although less pronounced, pattern is observed for the 2D and 3D datasets, where threshold-based PP values are generally close to zero but AUC PP decreases by $5.78$ and $5.08$ percentage points, respectively.
\begin{table*}[!t]
\caption{\deleted{\textbf{Complete Task-aware Protection, Clinical, and Calibration Metrics Across Modalities.}
Dataset-level entries report the mean $\pm$ standard deviation across random seeds in percentage points at the representative perturbation budget. Image and volumetric experiments use $\epsilon=8/255$, while tabular experiments use standardized-feature $\epsilon=0.5$. Aggregate rows report the mean $\pm$ standard deviation across dataset-level means; therefore, their standard deviations measure cross-dataset heterogeneity rather than seed uncertainty. Task Accuracy denotes accuracy for binary and multi-class tasks and subset accuracy for multi-label tasks. MC, Bin., and ML denote multi-class, binary, and multi-label tasks, respectively. AUC denotes binary AUC or macro-AUC as appropriate and is the pre-specified primary ranking metric, highlighted in \textbf{bold}. For predictive metrics, more negative CP indicates stronger unauthorized-performance degradation and PP closer to zero indicates better authorized utility preservation. ECE is lower-is-better; therefore, a positive ECE change indicates worse calibration. SynapseMNIST3D is retained for completeness but excluded from the 3D aggregate because its clean model collapsed. No result is claimed to be statistically significant after Holm correction.}\added{\textbf{Complete Task-aware Protection and Calibration Results Across Modalities.} Dataset-level values are mean ± standard deviation across seeds, while aggregate values summarize cross-dataset heterogeneity. The $28\times28$ and $224\times224$ experiments use $\epsilon=8/255$, the 3D experiments use $\epsilon=32/255$, and the tabular experiments use standardized-feature $\epsilon=5$ under the finalized protocol. AUC is the primary ranking metric. SynapseMNIST3D is excluded from the 3D aggregate because its clean model collapsed.}}
\label{tab:complete_task_aware_metrics}
\centering
\scriptsize
\setlength{\tabcolsep}{1.25pt}
\renewcommand{\arraystretch}{1.10}
\resizebox{\textwidth}{!}{%
\begin{tabular}{@{}llc*{8}{cc}@{}}
\toprule[1pt]
Dataset
& Task
& \(n\)
& \multicolumn{2}{c}{Task Acc.}
& \multicolumn{2}{c}{Bal. Acc.}
& \multicolumn{2}{c}{AUC}
& \multicolumn{2}{c}{Macro-F1}
& \multicolumn{2}{c}{Precision}
& \multicolumn{2}{c}{Sensitivity}
& \multicolumn{2}{c}{Specificity}
& \multicolumn{2}{c}{ECE \(\downarrow\)} \\
\cmidrule(lr){4-5}
\cmidrule(lr){6-7}
\cmidrule(lr){8-9}
\cmidrule(lr){10-11}
\cmidrule(lr){12-13}
\cmidrule(lr){14-15}
\cmidrule(lr){16-17}
\cmidrule(lr){18-19}
&
&
&
\(CP\downarrow\)
& \(PP\uparrow\)
& \(CP\downarrow\)
& \(PP\uparrow\)
& \(CP\downarrow\)
& \(PP\uparrow\)
& \(CP\downarrow\)
& \(PP\uparrow\)
& \(CP\downarrow\)
& \(PP\uparrow\)
& \(CP\downarrow\)
& \(PP\uparrow\)
& \(CP\downarrow\)
& \(PP\uparrow\)
& \(CP\)
& \(PP\) \\
\midrule[0.8pt]

\multicolumn{19}{l}{\textit{MedMNIST 2D}} \\

BloodMNIST
& MC
& 8
& \(-16.73\pm3.96\)
& \(0.00\pm0.00\)
& \(-25.04\pm4.58\)
& \(0.00\pm0.00\)
& \(\mathbf{-2.50\pm0.55}\)
& \(\mathbf{-1.70\pm0.44}\)
& \(-24.17\pm5.43\)
& \(0.00\pm0.00\)
& \(-10.54\pm3.01\)
& \(0.00\pm0.00\)
& \(-25.04\pm4.58\)
& \(0.00\pm0.00\)
& \(-2.50\pm0.60\)
& \(0.00\pm0.00\)
& \(6.17\pm3.76\)
& \(1.85\pm0.12\) \\

BreastMNIST
& Bin.
& 8
& \(-27.00\pm5.84\)
& \(0.00\pm0.00\)
& \(-16.69\pm5.52\)
& \(0.00\pm0.00\)
& \(\mathbf{-16.25\pm5.34}\)
& \(\mathbf{-5.17\pm3.02}\)
& \(-24.47\pm5.62\)
& \(0.00\pm0.00\)
& \(-2.85\pm4.38\)
& \(0.00\pm0.00\)
& \(-39.04\pm8.54\)
& \(0.00\pm0.00\)
& \(5.65\pm10.79\)
& \(0.00\pm0.00\)
& \(23.41\pm5.07\)
& \(0.62\pm1.09\) \\

ChestMNIST
& ML
& 3
& \(-47.05\pm2.17\)
& \(0.25\pm0.01\)
& --
& --
& \(\mathbf{1.27\pm0.39}\)
& \(\mathbf{-6.37\pm0.50}\)
& \(4.67\pm1.02\)
& \(-1.10\pm0.07\)
& \(-8.13\pm1.02\)
& \(-1.59\pm1.44\)
& \(23.01\pm1.24\)
& \(-0.75\pm0.09\)
& \(-16.62\pm2.21\)
& \(0.09\pm0.03\)
& --
& -- \\

DermaMNIST
& MC
& 8
& \(-6.71\pm1.32\)
& \(-0.01\pm0.03\)
& \(-10.07\pm5.32\)
& \(-0.08\pm0.22\)
& \(\mathbf{-5.67\pm2.20}\)
& \(\mathbf{-14.19\pm1.43}\)
& \(-16.34\pm5.11\)
& \(-0.09\pm0.23\)
& \(-16.21\pm3.09\)
& \(-0.07\pm0.20\)
& \(-10.07\pm5.32\)
& \(-0.08\pm0.22\)
& \(-2.86\pm1.15\)
& \(0.00\pm0.00\)
& \(0.90\pm3.17\)
& \(7.90\pm0.43\) \\

OrganCMNIST
& MC
& 8
& \(-2.61\pm1.51\)
& \(0.00\pm0.01\)
& \(-3.30\pm1.57\)
& \(0.00\pm0.01\)
& \(\mathbf{-0.27\pm0.27}\)
& \(\mathbf{-1.18\pm0.35}\)
& \(-3.33\pm1.67\)
& \(0.00\pm0.01\)
& \(-3.08\pm1.65\)
& \(0.00\pm0.01\)
& \(-3.30\pm1.57\)
& \(0.00\pm0.01\)
& \(-0.26\pm0.15\)
& \(0.00\pm0.00\)
& \(-1.04\pm0.59\)
& \(3.32\pm0.16\) \\

PneumoniaMNIST
& Bin.
& 8
& \(-10.00\pm5.81\)
& \(0.00\pm0.00\)
& \(-13.06\pm7.83\)
& \(0.00\pm0.00\)
& \(\mathbf{-3.09\pm2.06}\)
& \(\mathbf{-6.97\pm2.07}\)
& \(-14.06\pm8.57\)
& \(0.00\pm0.00\)
& \(-9.70\pm5.93\)
& \(0.00\pm0.00\)
& \(-0.80\pm1.08\)
& \(0.00\pm0.00\)
& \(-25.32\pm15.95\)
& \(0.00\pm0.00\)
& \(6.45\pm5.66\)
& \(1.57\pm0.38\) \\

RetinaMNIST
& MC
& 8
& \(-4.12\pm5.41\)
& \(0.09\pm0.64\)
& \(0.68\pm3.70\)
& \(0.00\pm0.81\)
& \(\mathbf{-4.72\pm4.28}\)
& \(\mathbf{-4.85\pm3.22}\)
& \(-2.31\pm3.71\)
& \(0.01\pm0.93\)
& \(-0.29\pm3.46\)
& \(0.10\pm1.21\)
& \(0.68\pm3.70\)
& \(0.00\pm0.81\)
& \(-0.54\pm1.30\)
& \(0.02\pm0.19\)
& \(2.41\pm4.61\)
& \(8.00\pm1.26\) \\

\textit{2D aggregate}
& --
& --
& \(-16.32\pm15.96\)
& \(0.05\pm0.10\)
& \(-11.25\pm9.27\)
& \(-0.01\pm0.03\)
& \(\mathbf{-4.46\pm5.73}\)
& \(\mathbf{-5.78\pm4.31}\)
& \(-11.43\pm11.34\)
& \(-0.17\pm0.41\)
& \(-7.26\pm5.52\)
& \(-0.22\pm0.60\)
& \(-7.79\pm19.87\)
& \(-0.12\pm0.28\)
& \(-6.06\pm10.85\)
& \(0.02\pm0.03\)
& \(6.38\pm8.84\)
& \(3.88\pm3.27\) \\

\addlinespace[2pt]
\multicolumn{19}{l}{\textit{MedMNIST 224$\times$224}} \\

BreastMNIST
& Bin.
& 3
& \(-15.38\pm0.00\)
& \(0.00\pm0.00\)
& \(-32.83\pm0.43\)
& \(0.00\pm0.00\)
& \(\mathbf{-37.36\pm4.40}\)
& \(\mathbf{-9.36\pm3.24}\)
& \(-42.30\pm0.16\)
& \(0.00\pm0.00\)
& \(-16.70\pm0.38\)
& \(0.00\pm0.00\)
& \(4.97\pm0.51\)
& \(0.00\pm0.00\)
& \(-70.63\pm1.37\)
& \(0.00\pm0.00\)
& \(17.63\pm0.77\)
& \(2.30\pm0.80\) \\

PneumoniaMNIST
& Bin.
& 3
& \(-36.16\pm9.76\)
& \(-0.05\pm0.09\)
& \(-39.07\pm3.27\)
& \(-0.07\pm0.12\)
& \(\mathbf{-40.72\pm10.24}\)
& \(\mathbf{-3.30\pm0.48}\)
& \(-52.84\pm1.72\)
& \(-0.06\pm0.11\)
& \(-19.55\pm14.67\)
& \(-0.07\pm0.12\)
& \(-27.44\pm48.85\)
& \(0.00\pm0.00\)
& \(-50.71\pm55.39\)
& \(-0.14\pm0.25\)
& \(34.48\pm3.96\)
& \(1.73\pm0.10\) \\

RetinaMNIST
& MC
& 3
& \(-30.92\pm6.36\)
& \(-0.33\pm0.29\)
& \(-25.29\pm2.68\)
& \(-0.27\pm0.30\)
& \(\mathbf{-18.91\pm2.38}\)
& \(\mathbf{-10.53\pm2.68}\)
& \(-35.23\pm4.41\)
& \(-0.34\pm0.26\)
& \(-31.27\pm6.54\)
& \(-0.41\pm0.25\)
& \(-25.29\pm2.68\)
& \(-0.27\pm0.30\)
& \(-7.31\pm1.37\)
& \(-0.05\pm0.07\)
& \(25.69\pm11.20\)
& \(14.15\pm1.31\) \\

\textit{224 aggregate}
& --
& --
& \(-27.49\pm10.81\)
& \(-0.13\pm0.18\)
& \(-32.40\pm6.90\)
& \(-0.11\pm0.14\)
& \(\mathbf{-32.33\pm11.74}\)
& \(\mathbf{-7.73\pm3.88}\)
& \(-43.46\pm8.86\)
& \(-0.14\pm0.18\)
& \(-22.51\pm7.72\)
& \(-0.16\pm0.22\)
& \(-15.92\pm18.12\)
& \(-0.09\pm0.16\)
& \(-42.89\pm32.38\)
& \(-0.06\pm0.07\)
& \(25.93\pm8.43\)
& \(6.06\pm7.01\) \\

\addlinespace[2pt]
\multicolumn{19}{l}{\textit{MedMNIST3D}} \\

FractureMNIST3D
& MC
& 3
& \(-27.64\pm5.67\)
& \(0.00\pm0.00\)
& \(-10.17\pm5.13\)
& \(0.00\pm0.00\)
& \(\mathbf{-7.83\pm1.60}\)
& \(\mathbf{-2.38\pm0.73}\)
& \(-30.14\pm7.64\)
& \(0.00\pm0.00\)
& \(-11.99\pm5.11\)
& \(0.00\pm0.00\)
& \(-10.17\pm5.13\)
& \(0.00\pm0.00\)
& \(-5.50\pm2.37\)
& \(0.00\pm0.00\)
& \(26.70\pm8.83\)
& \(16.82\pm2.25\) \\

NoduleMNIST3D
& Bin.
& 3
& \(-66.45\pm1.16\)
& \(0.00\pm0.00\)
& \(-30.31\pm2.18\)
& \(0.00\pm0.00\)
& \(\mathbf{-9.36\pm2.41}\)
& \(\mathbf{-10.09\pm4.29}\)
& \(-63.19\pm1.91\)
& \(0.00\pm0.00\)
& \(-48.11\pm2.71\)
& \(0.00\pm0.00\)
& \(31.25\pm4.13\)
& \(0.00\pm0.00\)
& \(-91.87\pm0.81\)
& \(0.00\pm0.00\)
& \(59.69\pm4.71\)
& \(2.34\pm0.33\) \\

OrganMNIST3D
& MC
& 3
& \(-25.52\pm5.43\)
& \(0.11\pm0.19\)
& \(-21.33\pm4.28\)
& \(0.09\pm0.16\)
& \(\mathbf{-2.65\pm0.63}\)
& \(\mathbf{-2.77\pm0.33}\)
& \(-23.74\pm3.98\)
& \(0.18\pm0.32\)
& \(-12.19\pm3.28\)
& \(0.25\pm0.43\)
& \(-21.33\pm4.28\)
& \(0.09\pm0.16\)
& \(-2.60\pm0.55\)
& \(0.01\pm0.02\)
& \(6.22\pm3.77\)
& \(2.88\pm0.56\) \\

SynapseMNIST3D\textsuperscript{\(\dagger\)}
& Bin.
& 3
& \(25.00\pm21.48\)
& \(30.49\pm25.46\)
& \(8.49\pm5.51\)
& \(3.41\pm1.74\)
& \(\mathbf{10.86\pm5.89}\)
& \(\mathbf{12.56\pm5.48}\)
& \(26.64\pm18.45\)
& \(21.64\pm11.30\)
& \(-7.01\pm10.96\)
& \(-10.85\pm14.07\)
& \(44.36\pm40.25\)
& \(62.26\pm55.95\)
& \(-27.37\pm29.34\)
& \(-55.44\pm57.17\)
& \(-18.00\pm18.39\)
& \(-17.84\pm17.65\) \\

\textit{3D aggregate}
& --
& --
& \(-39.87\pm23.04\)
& \(0.04\pm0.06\)
& \(-20.60\pm10.09\)
& \(0.03\pm0.05\)
& \(\mathbf{-6.61\pm3.52}\)
& \(\mathbf{-5.08\pm4.35}\)
& \(-39.02\pm21.17\)
& \(0.06\pm0.11\)
& \(-24.10\pm20.79\)
& \(0.08\pm0.14\)
& \(-0.08\pm27.70\)
& \(0.03\pm0.05\)
& \(-33.32\pm50.72\)
& \(0.00\pm0.01\)
& \(30.87\pm26.97\)
& \(7.35\pm8.21\) \\

\addlinespace[2pt]
\multicolumn{19}{l}{\textit{Tabular Medical Data}} \\

Diabetes
& Bin.
& 8
& \(-14.37\pm3.76\)
& \(-0.08\pm0.54\)
& \(-11.07\pm4.17\)
& \(-0.27\pm0.65\)
& \(\mathbf{-10.30\pm4.55}\)
& \(\mathbf{-0.30\pm0.42}\)
& \(-13.12\pm3.88\)
& \(-0.19\pm0.62\)
& \(-21.30\pm4.15\)
& \(0.25\pm0.97\)
& \(0.45\pm8.79\)
& \(-0.91\pm1.37\)
& \(-22.60\pm5.74\)
& \(0.38\pm0.75\)
& \(0.39\pm2.63\)
& \(-0.07\pm0.88\) \\

HepatitisC
& MC
& 8
& \(-36.28\pm6.21\)
& \(0.00\pm0.00\)
& \(-12.83\pm7.08\)
& \(0.00\pm0.00\)
& \(\mathbf{-7.96\pm3.60}\)
& \(\mathbf{-5.18\pm1.65}\)
& \(-34.12\pm5.71\)
& \(0.00\pm0.00\)
& \(-38.46\pm6.32\)
& \(0.00\pm0.00\)
& \(-12.83\pm7.08\)
& \(0.00\pm0.00\)
& \(-7.13\pm1.26\)
& \(0.00\pm0.00\)
& \(6.27\pm4.03\)
& \(0.72\pm0.35\) \\

SystemicDiseases
& ML
& 8
& \(-32.49\pm12.81\)
& \(-0.48\pm0.03\)
& --
& --
& \(\mathbf{-13.55\pm1.94}\)
& \(\mathbf{-1.01\pm0.07}\)
& \(-51.76\pm3.12\)
& \(-4.60\pm0.52\)
& \(-42.43\pm8.38\)
& \(-1.18\pm0.33\)
& \(-31.52\pm1.56\)
& \(-6.60\pm0.86\)
& \(-9.37\pm3.88\)
& \(-0.02\pm0.00\)
& \(5.26\pm1.53\)
& \(0.03\pm0.04\) \\

\textit{Tabular aggregate}
& --
& --
& \(-27.71\pm11.71\)
& \(-0.19\pm0.26\)
& \(-11.95\pm1.24\)
& \(-0.13\pm0.19\)
& \(\mathbf{-10.60\pm2.80}\)
& \(\mathbf{-2.16\pm2.64}\)
& \(-33.00\pm19.34\)
& \(-1.60\pm2.60\)
& \(-34.06\pm11.23\)
& \(-0.31\pm0.76\)
& \(-14.63\pm16.07\)
& \(-2.50\pm3.58\)
& \(-13.04\pm8.36\)
& \(0.12\pm0.23\)
& \(3.97\pm3.14\)
& \(0.23\pm0.43\) \\

\bottomrule[1pt]
\end{tabular}%
}
\end{table*}
The results also reveal substantial dataset-level heterogeneity. Within the 2D setting, BreastMNIST exhibits a relatively strong AUC CP of $-16.25$ percentage points, whereas OrganCMNIST shows only a $-0.27$-point decrease. ChestMNIST represents a particularly challenging multi-label case: its unauthorized AUC does not decrease, while its authorized AUC drops by $6.37$ percentage points. In the 3D setting, NoduleMNIST3D exhibits strong degradation in Task Accuracy and Macro-F1, but its sensitivity increases while specificity decreases substantially. This asymmetric behavior demonstrates that a large reduction in a single aggregate metric may arise from class- or threshold-specific prediction shifts and motivates reporting multiple task-aware metrics rather than accuracy alone. SynapseMNIST3D produces invalid protection trends because the underlying clean model is close to collapse and is therefore excluded from aggregate claims.

The tabular experiments demonstrate meaningful protection, with an aggregate AUC CP of $-10.60$ percentage points and an AUC PP of $-2.16$ points. Task Accuracy and specificity show CP values of $-27.71$ and $-13.04$ points, respectively, while their PP values remain close to zero, indicating substantial unauthorized degradation with relatively preserved authorized utility. Calibration results further qualify these findings: unauthorized ECE increases across the $28\times28$, $224\times224$, 3D, and tabular settings, while authorized ECE also increases, particularly for the $224\times224$ and 3D settings. Therefore, near-zero accuracy-based PP should not be interpreted as complete preservation of probabilistic reliability. Overall, \(\method\) produces authorization-dependent performance differences across imaging and structured medical data, but their magnitude and utility costs depend on the metric, perturbation protocol, dataset, and modality. Because no primary task-aware AUC comparison survives Holm correction in the unified analysis, these results are interpreted as descriptive and exploratory rather than confirmatory.

\section{Analysis of ChestMNIST}
\label{app:chestmnist_analysis}

ChestMNIST is analyzed separately because its sparse multi-label setting produces substantially different conclusions under thresholded and threshold-free metrics. We therefore report the complete metric profile, implementation diagnostics, and adaptive-attack results rather than relying on subset accuracy alone.

\paragraph{Metric-dependent protection behavior.}
Table~\ref{tab:chestmnist_focused_evaluation} reveals a substantial disagreement between thresholded exact-match accuracy and threshold-free ranking metrics. Under subset accuracy, authorized performance remains close to the clean baseline, whereas unauthorized performance decreases from 51.06\% to 4.02\%, producing a large positive authorization gap of 47.30 percentage points. Considered in isolation, this result would suggest strong protection with nearly complete authorized-utility preservation.

\begin{table*}[!t]
\small
\caption{\textbf{Focused Evaluation and Diagnostics for ChestMNIST.}
Panel A reports mean $\pm$ standard deviation over three random seeds. CP is unauthorized raw minus clean performance, PP is authorized cooked minus clean performance, and Gap is authorized cooked minus unauthorized raw, all in percentage points. More negative CP indicates stronger unauthorized degradation, PP closer to zero indicates better authorized-utility preservation, and a positive Gap indicates an authorization advantage. Macro-AUC is the pre-specified task-aware primary metric. Panel B reports pseudo-label quality, image quality, runtime, and exploratory statistical diagnostics. Runtime is obtained from one benchmark run with five internal repetitions. The statistical tests use paired macro-AUC values; with only three pairs, the exact Wilcoxon test is severely underpowered.}
\label{tab:chestmnist_focused_evaluation}
\centering
\setlength{\tabcolsep}{4.0pt}
\renewcommand{\arraystretch}{1.08}

\textit{Panel A: Multi-label predictive performance}\\[-2pt]

\resizebox{\textwidth}{!}{%
\begin{tabular}{lcccccc}
\toprule[1pt]
Metric
& Clean
& Authorized
& Unauthorized
& \(CP\downarrow\)
& \(PP\;(0\text{ best})\)
& \(Gap\uparrow\) \\
\midrule[0.8pt]

Subset ACC
& \(51.06\pm0.56\)
& \(51.31\pm0.56\)
& \(4.02\pm1.90\)
& \(-47.05\pm2.17\)
& \(0.25\pm0.01\)
& \(47.30\pm2.18\) \\

\textbf{Macro-AUC}
& \(\mathbf{65.91\pm0.11}\)
& \(\mathbf{59.54\pm0.40}\)
& \(\mathbf{67.18\pm0.48}\)
& \(\mathbf{1.27\pm0.39}\)
& \(\mathbf{-6.37\pm0.50}\)
& \(\mathbf{-7.64\pm0.77}\) \\

Micro-AUC
& \(71.32\pm0.84\)
& \(70.94\pm0.85\)
& \(74.80\pm1.52\)
& \(3.48\pm2.27\)
& \(-0.38\pm1.22\)
& \(-3.85\pm2.09\) \\

Macro-F1
& \(5.46\pm0.55\)
& \(4.35\pm0.51\)
& \(10.13\pm0.47\)
& \(4.67\pm1.02\)
& \(-1.10\pm0.07\)
& \(-5.78\pm0.98\) \\

Micro-F1
& \(10.82\pm1.39\)
& \(9.68\pm1.10\)
& \(21.62\pm1.94\)
& \(10.81\pm3.10\)
& \(-1.14\pm0.30\)
& \(-11.95\pm2.87\) \\

Macro-Precision
& \(17.89\pm0.50\)
& \(16.30\pm0.99\)
& \(9.77\pm1.06\)
& \(-8.13\pm1.02\)
& \(-1.59\pm1.44\)
& \(6.54\pm1.94\) \\

Micro-Precision
& \(34.71\pm1.58\)
& \(35.04\pm1.85\)
& \(13.97\pm1.19\)
& \(-20.74\pm0.67\)
& \(0.33\pm0.31\)
& \(21.07\pm0.81\) \\

Macro-Sensitivity
& \(3.46\pm0.44\)
& \(2.71\pm0.37\)
& \(26.48\pm1.52\)
& \(23.01\pm1.24\)
& \(-0.75\pm0.09\)
& \(-23.77\pm1.32\) \\

Micro-Sensitivity
& \(6.43\pm1.01\)
& \(5.63\pm0.77\)
& \(48.12\pm7.10\)
& \(41.69\pm7.31\)
& \(-0.80\pm0.24\)
& \(-42.49\pm7.30\) \\

Macro-Specificity
& \(99.27\pm0.16\)
& \(99.37\pm0.13\)
& \(82.66\pm2.27\)
& \(-16.62\pm2.21\)
& \(0.09\pm0.03\)
& \(16.71\pm2.23\) \\

Micro-Specificity
& \(99.32\pm0.15\)
& \(99.42\pm0.12\)
& \(83.56\pm2.08\)
& \(-15.76\pm2.01\)
& \(0.09\pm0.03\)
& \(15.85\pm2.03\) \\

\bottomrule[1pt]
\end{tabular}%
}

\vspace{5pt}
\textit{Panel B: Pseudo-label, image-quality, runtime, and statistical diagnostics}\\[-2pt]

\resizebox{0.78\textwidth}{!}{%
\begin{tabular}{llclr}
\toprule[1pt]
Family & Measure & Value & Unit & \(n\) \\
\midrule[0.8pt]

Pseudo-label
& Agreement with reference labels
& \(89.93\pm4.22\)
& \%
& 3 \\

Pseudo-label
& Mean surrogate confidence
& \(98.81\pm0.13\)
& \%
& 3 \\

\midrule

Image quality
& SSIM
& \(0.967\pm0.000\)
& score
& 3 \\

Image quality
& PSNR
& \(32.864\pm0.015\)
& dB
& 3 \\

Image quality
& LPIPS
& \(0.155\pm0.002\)
& score
& 3 \\

\midrule

Runtime
& Cooking latency
& \(2.044\)
& \(\mu\mathrm{s}/\)sample
& 1 \\

Runtime
& Raw inference latency
& \(39.465\)
& \(\mu\mathrm{s}/\)sample
& 1 \\

Runtime
& End-to-end overhead
& \(1.053\)
& ratio
& 1 \\

\midrule

Statistics
& Clean vs.\ authorized
& \(0.25\)
& Holm-adjusted \(p\)
& 3 \\

Statistics
& Clean vs.\ unauthorized
& \(1.00\)
& Holm-adjusted \(p\)
& 3 \\

\bottomrule[1pt]
\end{tabular}%
}

\end{table*}
However, the task-aware primary metric leads to the opposite conclusion. Authorized macro-AUC decreases from 65.91\% to 59.54\%, while unauthorized macro-AUC increases to 67.18\%. The resulting authorization gap is therefore negative. Micro-AUC and both macro- and micro-F1 show the same unfavorable ordering, with unauthorized performance exceeding authorized performance. Consequently, the large subset-accuracy gap does not reflect a consistent reduction in the unauthorized model's ranking or discrimination ability.

The detailed classwise metrics help explain this discrepancy. The clean and authorized models exhibit very low sensitivity but specificity above 99\%, which is consistent with a sparse prediction regime dominated by negative outputs. Such predictions can achieve relatively high subset accuracy on a sparse multi-label dataset because an instance is counted as correct only when every label is predicted exactly. In contrast, the unauthorized model has substantially higher sensitivity and lower specificity. This shift causes exact-match accuracy to collapse, while improving F1 and AUC-related behavior. Thus, subset accuracy is highly sensitive to the selected threshold and label sparsity and should not be treated as the primary evidence of protection for ChestMNIST.

\paragraph{Pseudo-label and implementation diagnostics.}
The pseudo-label diagnostics identify an additional limitation. Pseudo-label agreement is 89.93\%, substantially lower than for most other imaging datasets, despite a mean surrogate confidence of 98.81\%. This indicates that the surrogate can assign highly confident pseudo labels that do not agree with the reference multi-label targets. The result does not by itself establish that pseudo-label errors cause the unfavorable CP--PP behavior, but it identifies a plausible process-level failure that is specific to the multi-label setting.

The observed metric discrepancy is unlikely to be explained by poor image similarity or excessive runtime overhead. At the representative budget, ChestMNIST retains an SSIM of 0.967 and a PSNR of 32.864 dB, while cooking adds approximately \(2.044~\mu\mathrm{s}\) per sample and yields an end-to-end overhead of only \(1.053\times\). These measurements indicate that the principal limitation arises from the predictive behavior of the protected model rather than an unusually large computational or image-quality cost.

\paragraph{Adaptive attacks.}
Table~\ref{tab:chestmnist_adaptive_attacks} further demonstrates that attack conclusions depend on the selected metric. Most input transformations reduce macro-AUC rather than recovering it. Strong Gaussian smoothing, median filtering, and augmentation produce increasingly negative macro-AUC gains, suggesting that they remove task-relevant information rather than selectively reversing the cooking mechanism. Some stronger augmentations increase subset accuracy, but this occurs while macro-AUC decreases substantially, again illustrating that thresholded exact-match behavior can improve without recovering ranking performance.
\begin{table*}[!t]
\scriptsize
\caption{\textbf{Adaptive Attacks on ChestMNIST.}
The table jointly reports macro-AUC, the threshold-free primary ranking metric, and subset accuracy, the exact-match thresholded metric. After is the post-attack unauthorized score, Gain is After minus the unattacked unauthorized score, and CP after is After minus clean performance, all in percentage points. Positive Gain indicates that the attack increases the corresponding unauthorized score. However, because unattacked unauthorized macro-AUC already exceeds clean macro-AUC, a positive macro-AUC Gain should not automatically be interpreted as recovery from an existing protection gap.}
\label{tab:chestmnist_adaptive_attacks}
\centering
\setlength{\tabcolsep}{3.0pt}
\renewcommand{\arraystretch}{1.08}
\resizebox{\textwidth}{!}{%
\begin{tabular}{lc*{3}{c}*{3}{c}}
\toprule[1pt]
Attack
& \(n\)
& \multicolumn{3}{c}{Macro-AUC}
& \multicolumn{3}{c}{Subset ACC} \\
\cmidrule(lr){3-5}
\cmidrule(lr){6-8}
&
& After
& Gain
& CP after
& After
& Gain
& CP after \\
\midrule[0.8pt]

\multicolumn{8}{l}{\textit{Input-transformation attacks}} \\

JPEG (\(Q=50\))
& 3
& \(62.54\pm0.99\)
& \(-4.64\pm0.94\)
& \(-3.37\pm0.91\)
& \(2.16\pm0.48\)
& \(-1.86\pm1.44\)
& \(-48.91\pm0.93\) \\

JPEG (\(Q=75\))
& 3
& \(64.68\pm0.62\)
& \(-2.50\pm0.19\)
& \(-1.23\pm0.51\)
& \(2.60\pm0.58\)
& \(-1.42\pm1.34\)
& \(-48.46\pm0.86\) \\

JPEG (\(Q=95\))
& 3
& \(66.62\pm0.38\)
& \(-0.56\pm0.22\)
& \(0.71\pm0.27\)
& \(4.74\pm1.77\)
& \(0.72\pm0.21\)
& \(-46.33\pm2.01\) \\

Gaussian (\(\sigma=0.5\))
& 3
& \(65.86\pm0.36\)
& \(-1.32\pm0.40\)
& \(-0.05\pm0.37\)
& \(1.19\pm1.04\)
& \(-2.82\pm0.88\)
& \(-49.87\pm1.31\) \\

Gaussian (\(\sigma=1\))
& 3
& \(59.22\pm1.62\)
& \(-7.96\pm1.71\)
& \(-6.69\pm1.68\)
& \(0.23\pm0.03\)
& \(-3.79\pm1.93\)
& \(-50.83\pm0.54\) \\

Gaussian (\(\sigma=2\))
& 3
& \(54.14\pm2.47\)
& \(-13.04\pm2.31\)
& \(-11.77\pm2.48\)
& \(0.29\pm0.11\)
& \(-3.73\pm1.98\)
& \(-50.78\pm0.60\) \\

Median (\(K=3\))
& 3
& \(64.07\pm0.68\)
& \(-3.11\pm0.31\)
& \(-1.84\pm0.62\)
& \(2.99\pm1.63\)
& \(-1.03\pm0.28\)
& \(-48.08\pm1.89\) \\

Median (\(K=5\))
& 3
& \(59.64\pm0.29\)
& \(-7.54\pm0.64\)
& \(-6.27\pm0.29\)
& \(1.23\pm0.65\)
& \(-2.78\pm1.29\)
& \(-49.83\pm1.09\) \\

Augmentation (\(L=1\))
& 3
& \(59.00\pm0.26\)
& \(-8.18\pm0.44\)
& \(-6.91\pm0.29\)
& \(3.45\pm1.45\)
& \(-0.57\pm0.46\)
& \(-47.62\pm1.76\) \\

Augmentation (\(L=2\))
& 3
& \(54.73\pm0.36\)
& \(-12.45\pm0.18\)
& \(-11.18\pm0.25\)
& \(9.38\pm2.15\)
& \(5.37\pm0.61\)
& \(-41.68\pm2.53\) \\

Augmentation (\(L=3\))
& 3
& \(52.65\pm0.13\)
& \(-14.53\pm0.40\)
& \(-13.26\pm0.15\)
& \(14.22\pm3.16\)
& \(10.21\pm1.40\)
& \(-36.84\pm3.52\) \\

\midrule
\multicolumn{8}{l}{\textit{Full-validation fine-tuning attacks}} \\

FT-5
& 3
& \(66.69\pm1.56\)
& \(-0.49\pm2.03\)
& \(0.78\pm1.65\)
& \(43.62\pm4.92\)
& \(39.61\pm3.71\)
& \(-7.44\pm4.77\) \\

FT-10
& 3
& \(66.41\pm0.48\)
& \(-0.77\pm0.56\)
& \(0.50\pm0.51\)
& \(50.14\pm0.68\)
& \(46.12\pm1.43\)
& \(-0.93\pm1.19\) \\

RFT-5
& 3
& \(67.66\pm2.82\)
& \(0.48\pm2.36\)
& \(1.75\pm2.75\)
& \(53.15\pm0.04\)
& \(49.13\pm1.89\)
& \(2.09\pm0.59\) \\

RFT-10
& 3
& \(70.12\pm1.72\)
& \(2.94\pm1.27\)
& \(4.21\pm1.64\)
& \(53.05\pm0.22\)
& \(49.03\pm1.77\)
& \(1.98\pm0.49\) \\

\midrule
\multicolumn{8}{l}{\textit{Fixed raw-data-budget fine-tuning attacks}} \\

FT-10 (25 samples)
& 3
& \(60.40\pm4.87\)
& \(-6.78\pm4.98\)
& \(-5.51\pm4.84\)
& \(44.61\pm14.82\)
& \(40.60\pm16.70\)
& \(-6.45\pm14.54\) \\

FT-10 (50 samples)
& 3
& \(59.95\pm3.45\)
& \(-7.23\pm3.88\)
& \(-5.96\pm3.51\)
& \(26.17\pm22.81\)
& \(22.16\pm23.21\)
& \(-24.89\pm23.28\) \\

RFT-10 (25 samples)
& 3
& \(51.71\pm2.75\)
& \(-15.47\pm2.66\)
& \(-14.20\pm2.78\)
& \(1.18\pm1.33\)
& \(-2.84\pm2.63\)
& \(-49.88\pm0.77\) \\

RFT-10 (50 samples)
& 3
& \(55.04\pm2.34\)
& \(-12.14\pm1.92\)
& \(-10.87\pm2.28\)
& \(14.02\pm19.64\)
& \(10.00\pm17.80\)
& \(-37.05\pm19.97\) \\

\bottomrule[1pt]
\end{tabular}%
}
\end{table*}
Full-validation fine-tuning produces a different pattern. Ordinary and robust fine-tuning restore subset accuracy from 4.02\% to approximately 44--53\%, largely eliminating the thresholded protection gap. Robust fine-tuning also increases macro-AUC, with RFT-10 reaching 70.12\%. Nevertheless, the unattacked unauthorized macro-AUC already exceeds the clean macro-AUC. Therefore, this increase should not be described as recovery from a pre-existing AUC-based protection effect; rather, it further confirms that the intended protection ordering is absent under the task-aware primary metric.

When the attacker is restricted to 25 or 50 labeled raw samples, macro-AUC generally decreases and subset-accuracy recovery becomes highly variable. The large standard deviations, particularly for the 50-sample settings, indicate unstable adaptation across the three seeds. These results suggest that reliable thresholded recovery requires broader access to representative validation data, but they do not alter the main conclusion that ChestMNIST lacks a positive authorization gap under macro-AUC.

\paragraph{Summary.}
ChestMNIST should therefore be interpreted as a boundary case rather than a successful demonstration of metric-independent protection. \(\method\) preserves authorized subset accuracy and strongly changes unauthorized thresholded predictions, but it does not degrade unauthorized threshold-free ranking performance under macro-AUC. The results motivate task-aware evaluation for multi-label data and show that subset accuracy alone can substantially overstate protection when labels are sparse and decision thresholds are poorly aligned. Because all experiments use only three seeds and no paired test survives Holm correction, the findings are treated as descriptive diagnostics rather than statistically confirmed effects.

\section{Perturbation-budget Ablation}
\label{app:Perturbation-budget}
\begin{figure*}[!t]
    \centering
    \scriptsize
    \includegraphics[width=0.85\textwidth]{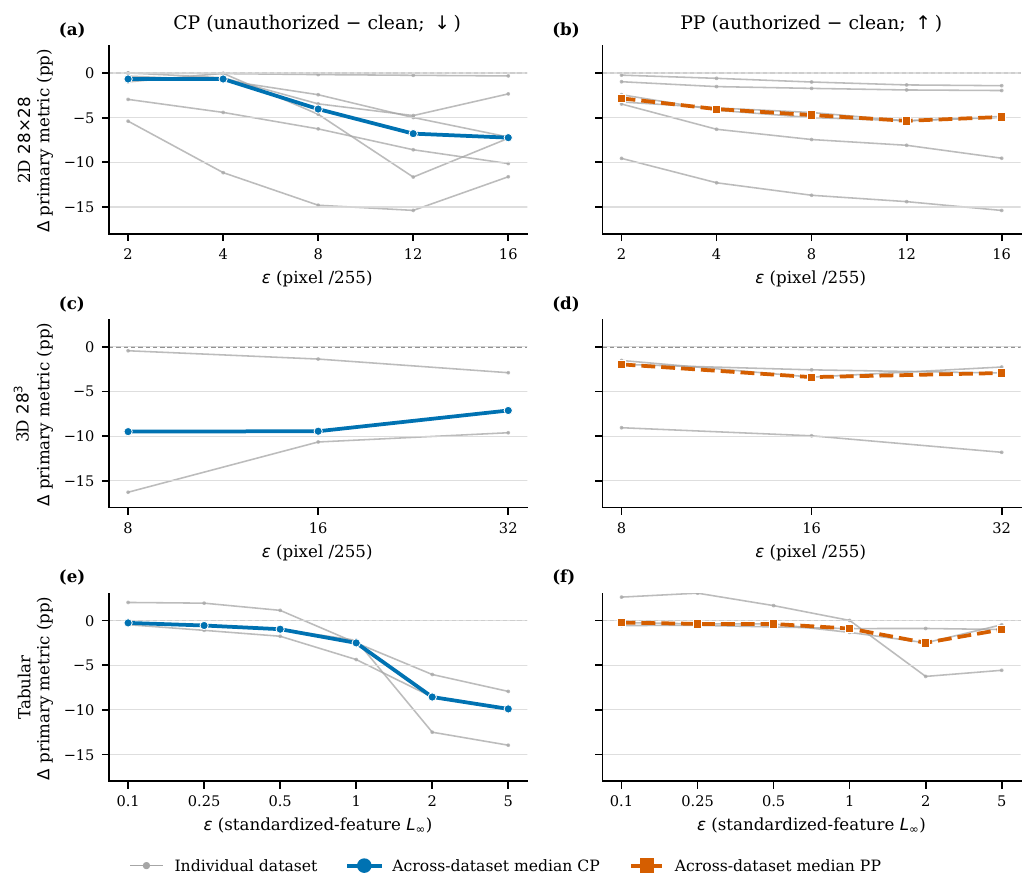}
    \caption{ \textbf{Increasing the perturbation budget generally strengthens median unauthorized degradation on 2D and tabular tasks, but also reduces authorized utility, while the 3D trend is non-monotonic.} 
    Gray curves are individual dataset/model settings and the colored curve is the across-dataset median. CP is unauthorized raw minus clean performance (lower is stronger protection), whereas PP is authorized cooked minus clean performance (closer to zero is better utility). Binary tasks use AUC and multiclass/multilabel tasks use macro-AUC. Image budgets are in pixel /255 units; tabular budgets are standardized-feature L-infinity values. SynapseMNIST3D is excluded because its clean model collapsed; ChestMNIST and 224×224 settings have only one budget and therefore cannot define an ablation trend.}
    \label{fig:perturbation_budget_overview}
\end{figure*}

Figure~\ref{fig:perturbation_budget_overview} shows that the effect of increasing the perturbation budget is strongly modality- and dataset-dependent. On the $28\times28$ 2D datasets, the across-dataset median CP becomes progressively more negative as $\epsilon$ increases from $2/255$ to $12/255$, indicating stronger degradation on unauthorized raw inputs. However, the improvement largely saturates between $12/255$ and $16/255$. At the same time, the median PP moves farther from zero, showing that stronger protection is accompanied by a gradual loss of authorized utility. The individual gray curves also vary substantially, indicating that the magnitude of this trade-off differs across datasets rather than following a uniform response.

\added{The tabular experiments exhibit a clear budget-dependent pattern. Median CP remains close to zero at small budgets but becomes substantially more negative at $\epsilon=2$ and reaches its strongest degradation at $\epsilon=5$. Median PP remains near zero up to $\epsilon=1$, decreases at $\epsilon=2$, and then partially recovers at $\epsilon=5$. Thus, larger standardized-feature perturbations strengthen unauthorized degradation, while the associated authorized-utility cost is non-monotonic and is not largest at the strongest protection budget.}

In contrast, the 3D results are non-monotonic. The median CP becomes slightly stronger from $\epsilon=8/255$ to $16/255$, but weakens again at $\epsilon=32/255$, while the median PP changes only modestly. The individual dataset curves also respond differently to increasing $\epsilon$, demonstrating that a larger feasible perturbation region does not necessarily translate into stronger protection for volumetric inputs. Overall, these results show that no single perturbation budget is uniformly optimal across datasets or modalities. Budget selection should therefore consider the joint CP--PP trade-off rather than maximizing perturbation magnitude alone. ChestMNIST and the $224\times224$ settings are not included in the trend analysis because only one perturbation budget is available for these experiments.

\section{Pseudo-label Quality}
\label{app:pseudo_quality}
Table~\ref{tab:pseudo_label_quality_representative} shows that pseudo-label generation is highly accurate and confident for most evaluated imaging settings. Excluding ChestMNIST and the collapsed SynapseMNIST3D setting, pseudo accuracy ranges from 98.24\% to 100\%, while mean surrogate confidence ranges from 97.12\% to 99.98\%. However, similarly high pseudo-label quality can correspond to markedly different CP--PP outcomes. For example, OrganCMNIST exhibits nearly perfect pseudo-label accuracy but only weak unauthorized AUC degradation, whereas the high-resolution BreastMNIST and PneumoniaMNIST settings achieve substantially stronger CP. Pseudo-label correctness is therefore not sufficient by itself to determine protection efficacy.
\begin{table*}[!t]
\scriptsize
\caption{\textbf{Pseudo-label Quality and Primary Protection Metrics at the Representative Budgets.}
\deleted{All image and volumetric experiments use $\epsilon=8/255$}\added{MedMNIST 2D and 224$\times$224 experiments use $\epsilon=8/255$, while MedMNIST3D experiments use $\epsilon=32/255$}. Pseudo accuracy measures agreement between the surrogate-generated pseudo labels and the reference labels, while confidence is the mean confidence assigned by the surrogate to its selected pseudo labels. CP-AUC and PP-AUC denote the unauthorized-raw-minus-clean and authorized-cooked-minus-clean changes in the task-aware primary metric, respectively, in \%. Binary tasks use AUC, whereas multi-class and multi-label tasks use macro-AUC. Values are reported as mean $\pm$ standard deviation across random seeds. $\ddagger$ denotes pseudo accuracy below 90\% as a diagnostic warning rather than statistical significance. $\dagger$ denotes a collapsed clean model retained for completeness but excluded from aggregate claims.}

\label{tab:pseudo_label_quality_representative}
\centering
\setlength{\tabcolsep}{4.2pt}
\renewcommand{\arraystretch}{1.08}
\resizebox{\textwidth}{!}{%
\begin{tabular}{@{}lllrcccc@{}}
\toprule[1pt]
Dataset
& Input
& Task
& \(n\)
& Pseudo ACC (\%)
& Confidence (\%)
& \(CP_{\mathrm{AUC}}\downarrow\)
& \(PP_{\mathrm{AUC}}\uparrow\) \\
\midrule[0.8pt]

\multicolumn{8}{l}{\textit{MedMNIST 2D}} \\

BloodMNIST
& \(28\times28\)
& Multi-class
& 8
& \(100.00\pm0.00\)
& \(99.90\pm0.00\)
& \(-2.50\pm0.55\)
& \(-1.70\pm0.44\) \\

BreastMNIST
& \(28\times28\)
& Binary
& 8
& \(99.95\pm0.08\)
& \(99.90\pm0.12\)
& \(-16.25\pm5.34\)
& \(-5.17\pm3.02\) \\

ChestMNIST\textsuperscript{\(\ddagger\)}
& \(28\times28\)
& Multi-label
& 3
& \(89.93\pm4.22\)
& \(98.81\pm0.13\)
& \(1.27\pm0.39\)
& \(-6.37\pm0.50\) \\

DermaMNIST
& \(28\times28\)
& Multi-class
& 8
& \(100.00\pm0.01\)
& \(99.89\pm0.01\)
& \(-5.67\pm2.20\)
& \(-14.19\pm1.43\) \\

OrganCMNIST
& \(28\times28\)
& Multi-class
& 8
& \(100.00\pm0.00\)
& \(99.90\pm0.01\)
& \(-0.27\pm0.27\)
& \(-1.18\pm0.35\) \\

PneumoniaMNIST
& \(28\times28\)
& Binary
& 8
& \(100.00\pm0.00\)
& \(99.98\pm0.00\)
& \(-3.09\pm2.06\)
& \(-6.97\pm2.07\) \\

RetinaMNIST
& \(28\times28\)
& Multi-class
& 8
& \(99.41\pm0.46\)
& \(99.23\pm0.43\)
& \(-4.72\pm4.28\)
& \(-4.85\pm3.22\) \\

\addlinespace[2pt]
\multicolumn{8}{l}{\textit{MedMNIST 224$\times$224}} \\

BreastMNIST
& \(224\times224\)
& Binary
& 3
& \(100.00\pm0.00\)
& \(99.62\pm0.06\)
& \(-37.36\pm4.40\)
& \(-9.36\pm3.24\) \\

PneumoniaMNIST
& \(224\times224\)
& Binary
& 3
& \(99.85\pm0.04\)
& \(99.79\pm0.01\)
& \(-40.72\pm10.24\)
& \(-3.30\pm0.48\) \\

RetinaMNIST
& \(224\times224\)
& Multi-class
& 3
& \(98.24\pm0.42\)
& \(97.12\pm0.32\)
& \(-18.91\pm2.38\)
& \(-10.53\pm2.68\) \\

\addlinespace[2pt]
\multicolumn{8}{l}{\textit{MedMNIST3D}} \\

FractureMNIST3D
& \(28^3\)
& Multi-class
& 3
& \(100.00\pm0.00\)
& \(99.93\pm0.01\)
& \(-7.83\pm1.60\)
& \(-2.38\pm0.73\) \\

NoduleMNIST3D
& \(28^3\)
& Binary
& 3
& \(99.97\pm0.05\)
& \(99.93\pm0.00\)
& \(-9.36\pm2.41\)
& \(-10.09\pm4.29\) \\

OrganMNIST3D
& \(28^3\)
& Multi-class
& 3
& \(99.86\pm0.06\)
& \(99.72\pm0.05\)
& \(-2.65\pm0.63\)
& \(-2.77\pm0.33\) \\

SynapseMNIST3D\textsuperscript{\(\dagger,\ddagger\)}
& \(28^3\)
& Binary
& 3
& \(42.63\pm27.15\)
& \(70.75\pm15.55\)
& \(10.86\pm5.89\)
& \(12.56\pm5.48\) \\

\bottomrule[1pt]
\end{tabular}%
}
\end{table*}
ChestMNIST illustrates that confidence and correctness should also be distinguished. Its pseudo-label accuracy decreases to 89.93\%, despite a mean surrogate confidence of 98.81\%. Its primary-metric behavior is unfavorable, with authorized macro-AUC decreasing while unauthorized macro-AUC does not decrease. SynapseMNIST3D provides a more extreme diagnostic failure, combining low and unstable pseudo-label agreement with a collapsed clean model; it is therefore excluded from aggregate and correlation analyses.
\begin{table*}[!t]
\scriptsize
\caption{\textbf{Exploratory Associations between Pseudo-label Quality and Primary CP/PP Metrics.}
Spearman correlations are computed from all available seed--budget observations within each dataset/model setting. Only associations with at least nine observations and non-zero variation in both variables are reported. CP and PP are defined using the task-aware primary AUC or macro-AUC metric. Settings with only three observations, mathematically undefined correlations, and the collapsed SynapseMNIST3D run are omitted. These associations are descriptive and do not support multiplicity-adjusted or causal inference.}
\label{tab:pseudo_label_quality_correlation}
\centering
\setlength{\tabcolsep}{5.0pt}
\renewcommand{\arraystretch}{1.08}
\resizebox{\textwidth}{!}{%
\begin{tabular}{@{}lllrrr@{}}
\toprule[1pt]
Dataset
& Input
& Pseudo-label variable
& \(n\)
& \(\rho\) with CP-AUC
& \(\rho\) with PP-AUC \\
\midrule[0.8pt]

\multicolumn{6}{l}{\textit{MedMNIST 2D}} \\

BloodMNIST
& \(28\times28\)
& Mean confidence
& 40
& \(0.026\)
& \(0.150\) \\

BreastMNIST
& \(28\times28\)
& Pseudo accuracy
& 40
& \(-0.015\)
& \(0.340\) \\

BreastMNIST
& \(28\times28\)
& Mean confidence
& 40
& \(0.084\)
& \(0.419\) \\

DermaMNIST
& \(28\times28\)
& Pseudo accuracy
& 40
& \(-0.095\)
& \(0.167\) \\

DermaMNIST
& \(28\times28\)
& Mean confidence
& 40
& \(-0.172\)
& \(-0.181\) \\

OrganCMNIST
& \(28\times28\)
& Mean confidence
& 40
& \(-0.519\)
& \(-0.340\) \\

PneumoniaMNIST
& \(28\times28\)
& Mean confidence
& 40
& \(-0.267\)
& \(-0.125\) \\

RetinaMNIST
& \(28\times28\)
& Pseudo accuracy
& 40
& \(0.064\)
& \(0.270\) \\

RetinaMNIST
& \(28\times28\)
& Mean confidence
& 40
& \(0.180\)
& \(0.371\) \\

\addlinespace[2pt]
\multicolumn{6}{l}{\textit{MedMNIST3D}} \\

FractureMNIST3D
& \(28^3\)
& Mean confidence
& 9
& \(0.422\)
& \(0.685\) \\

NoduleMNIST3D
& \(28^3\)
& Pseudo accuracy
& 9
& \(0.091\)
& \(-0.730\) \\

NoduleMNIST3D
& \(28^3\)
& Mean confidence
& 9
& \(-0.105\)
& \(-0.632\) \\

OrganMNIST3D
& \(28^3\)
& Pseudo accuracy
& 9
& \(-0.091\)
& \(0.548\) \\

OrganMNIST3D
& \(28^3\)
& Mean confidence
& 9
& \(0.000\)
& \(0.316\) \\

\bottomrule[1pt]
\end{tabular}%
}
\end{table*}
Table~\ref{tab:pseudo_label_quality_correlation} reports only mathematically defined associations supported by at least nine seed--budget observations. The resulting correlations remain inconsistent across datasets. For example, pseudo accuracy is negatively associated with PP-AUC on NoduleMNIST3D but positively associated with PP-AUC on OrganMNIST3D. Confidence-based associations similarly vary in direction and magnitude. Several datasets are absent because their pseudo-label accuracy is effectively constant, making correlation analysis uninformative rather than indicating missing data.

Overall, pseudo-label accuracy and confidence are best interpreted as process-level diagnostics. They help verify whether the cooking objective is based on stable surrogate predictions and identify problematic settings, but the current evidence does not support a universal or causal relationship between pseudo-label quality and downstream CP or PP.

\section{Adaptive Attacks}
\label{app:Adaptive}
Table~\ref{tab:adaptive_attack_detailed} provides the complete results for the evaluated input-transformation attacks and limited raw-data fine-tuning attacks. On most $28\times28$ datasets, input transformations do not restore unauthorized performance. Mild JPEG compression generally produces changes close to zero, whereas stronger Gaussian smoothing, median filtering, and augmentation frequently reduce the primary metric further. These transformations therefore tend to remove task-relevant information together with any fingerprint-related structure, rather than selectively eliminating the protection mechanism.
\begin{table*}[!t]
\scriptsize
\caption{\deleted{\textbf{Detailed Adaptive Attack Evaluation.}
Panel A reports changes in the task-aware primary metric under input-transformation attacks. JPEG columns denote compression quality, Gaussian columns denote standard deviation $\sigma$, median-filter columns denote kernel size, and augmentation columns denote augmentation strength. Panel B evaluates ordinary and robust 10-epoch fine-tuning when the attacker has access to 25 or 50 labeled raw samples. FT and RFT denote ordinary and robust fine-tuning, respectively, and their numerical suffixes indicate the number of labeled raw samples. All entries report the attack-induced change in unauthorized primary performance, in percentage points, as mean $\pm$ standard deviation across the available seed cohort. Positive values indicate attacker recovery, negative values indicate further performance degradation, and a dash denotes an unmeasured configuration rather than a zero-valued result.}\added{\textbf{Detailed Adaptive Attack Evaluation.} Panel A reports input-transformation attacks, and Panel B reports limited-data fine-tuning attacks. Values indicate changes in unauthorized task-aware performance (percentage points), reported as mean ± standard deviation across available seeds. Positive values indicate attacker recovery.
}}
\label{tab:adaptive_attack_detailed}
\centering
\setlength{\tabcolsep}{2.5pt}
\renewcommand{\arraystretch}{1.08}

\textit{Panel A: Input-transformation attacks}\\[0pt]

\resizebox{\textwidth}{!}{%
\begin{tabular}{lrrccccccccccc}
\toprule[1pt]
Dataset
& \(\epsilon\)
& \(n\)
& JPEG 50
& JPEG 75
& JPEG 95
& Gauss. 0.5
& Gauss. 1
& Gauss. 2
& Median 3
& Median 5
& Aug. 1
& Aug. 2
& Aug. 3 \\
\midrule[0.8pt]

Blood-28
& 8 & 8
& \(-1.8\pm1.1\)
& \(0.0\pm0.1\)
& \(0.0\pm0.0\)
& \(-2.2\pm1.4\)
& \(-12.9\pm7.4\)
& \(-19.6\pm7.2\)
& \(-3.6\pm2.7\)
& \(-8.1\pm5.2\)
& \(-28.9\pm2.5\)
& \(-42.7\pm1.4\)
& \(-45.2\pm0.6\) \\

Breast-28
& 8 & 8
& \(0.9\pm0.8\)
& \(0.6\pm0.8\)
& \(-0.0\pm0.2\)
& \(-1.7\pm2.0\)
& \(-11.7\pm7.9\)
& \(-18.7\pm8.3\)
& \(-8.0\pm5.4\)
& \(-17.4\pm6.9\)
& \(-9.7\pm8.8\)
& \(-14.6\pm6.7\)
& \(-19.1\pm6.6\) \\

Chest-28
& 8 & 3
& \(-4.6\pm0.9\)
& \(-2.5\pm0.2\)
& \(-0.6\pm0.2\)
& \(-1.3\pm0.4\)
& \(-8.0\pm1.7\)
& \(-13.0\pm2.3\)
& \(-3.1\pm0.3\)
& \(-7.5\pm0.6\)
& \(-8.2\pm0.4\)
& \(-12.5\pm0.2\)
& \(-14.5\pm0.4\) \\

Derma-28
& 8 & 8
& \(-2.0\pm0.7\)
& \(0.0\pm0.2\)
& \(0.0\pm0.1\)
& \(-0.3\pm0.2\)
& \(-1.8\pm0.8\)
& \(-3.6\pm1.2\)
& \(-1.0\pm0.5\)
& \(-3.1\pm1.1\)
& \(-14.1\pm1.9\)
& \(-24.0\pm2.3\)
& \(-30.2\pm3.6\) \\

OrganC-28
& 8 & 8
& \(-0.4\pm0.3\)
& \(-0.2\pm0.1\)
& \(-0.0\pm0.0\)
& \(-0.4\pm0.5\)
& \(-16.3\pm4.2\)
& \(-26.3\pm4.8\)
& \(-3.6\pm2.4\)
& \(-10.7\pm3.9\)
& \(-14.9\pm1.8\)
& \(-28.6\pm0.9\)
& \(-37.9\pm0.4\) \\

Pneumonia-28
& 8 & 8
& \(-3.0\pm1.0\)
& \(-0.1\pm0.0\)
& \(0.0\pm0.1\)
& \(0.1\pm1.0\)
& \(-6.7\pm8.4\)
& \(-17.7\pm18.0\)
& \(-0.5\pm1.4\)
& \(-5.4\pm4.4\)
& \(-11.0\pm4.2\)
& \(-24.4\pm5.8\)
& \(-30.9\pm5.3\) \\

Retina-28
& 8 & 8
& \(0.3\pm1.7\)
& \(-0.7\pm1.0\)
& \(0.3\pm0.5\)
& \(-0.2\pm0.3\)
& \(-0.9\pm1.4\)
& \(-1.6\pm2.1\)
& \(-0.6\pm0.5\)
& \(-1.2\pm1.2\)
& \(-5.5\pm3.1\)
& \(-7.3\pm4.6\)
& \(-9.6\pm4.9\) \\

Retina-28 (\(\epsilon=12\))
& 12 & 8
& \(0.6\pm2.2\)
& \(0.3\pm0.7\)
& \(0.1\pm0.6\)
& \(-0.2\pm0.3\)
& \(-0.8\pm1.5\)
& \(-1.6\pm2.5\)
& \(-0.3\pm0.8\)
& \(-1.2\pm2.0\)
& \(-2.7\pm2.6\)
& \(-3.7\pm3.4\)
& \(-3.8\pm6.2\) \\

\midrule

Breast-224
& 8 & 3
& \(-4.0\pm1.0\)
& \(-1.4\pm0.2\)
& \(-0.5\pm0.5\)
& \(1.0\pm1.4\)
& \(14.4\pm7.7\)
& \(18.1\pm9.2\)
& \(4.5\pm4.4\)
& \(14.6\pm4.0\)
& \(2.8\pm4.6\)
& \(0.9\pm3.9\)
& \(1.5\pm6.0\) \\

Pneumonia-224
& 8 & 3
& \(-15.6\pm11.0\)
& \(-14.4\pm10.9\)
& \(-8.1\pm7.0\)
& \(-12.8\pm6.4\)
& \(-15.0\pm7.7\)
& \(-14.5\pm7.2\)
& \(-13.7\pm8.9\)
& \(-15.7\pm9.0\)
& \(-5.8\pm6.8\)
& \(-0.3\pm8.1\)
& \(-2.5\pm8.8\) \\

Retina-224
& 8 & 3
& \(-3.8\pm1.6\)
& \(-1.3\pm0.7\)
& \(1.1\pm0.7\)
& \(-1.7\pm0.8\)
& \(-6.9\pm3.9\)
& \(-11.1\pm4.7\)
& \(-3.8\pm1.5\)
& \(-9.6\pm4.4\)
& \(-5.2\pm2.3\)
& \(-12.0\pm2.1\)
& \(-10.9\pm2.1\) \\

\bottomrule[1pt]
\end{tabular}%
}

\vspace{1pt}
\textit{Panel B: Fixed raw-data-budget fine-tuning attacks}\\[0pt]

\resizebox{0.74\textwidth}{!}{%
\begin{tabular}{lrrcccc}
\toprule[1pt]
Dataset
& \(\epsilon\)
& \(n\)
& FT-25
& FT-50
& RFT-25
& RFT-50 \\
\midrule[0.8pt]

Blood-28
& 8 & 8
& \(-2.8\pm2.9\)
& \(-0.9\pm0.4\)
& \(-26.6\pm8.5\)
& \(-21.8\pm6.6\) \\

Breast-28
& 8 & 8
& \(1.7\pm2.0\)
& \(3.2\pm3.8\)
& \(2.7\pm1.9\)
& \(2.7\pm2.5\) \\

Chest-28
& 8 & 3
& \(-6.8\pm5.0\)
& \(-7.2\pm3.9\)
& \(-15.5\pm2.7\)
& \(-12.1\pm1.9\) \\

Derma-28
& 8 & 8
& \(-3.5\pm3.1\)
& \(-3.4\pm2.3\)
& \(-6.3\pm2.0\)
& \(-4.6\pm2.7\) \\

OrganC-28
& 8 & 8
& \(-2.5\pm1.8\)
& \(-3.2\pm3.4\)
& \(-13.0\pm5.4\)
& \(-12.5\pm3.7\) \\

Pneumonia-28
& 8 & 8
& \(-1.2\pm2.6\)
& \(-0.7\pm2.6\)
& \(-5.6\pm6.2\)
& \(-5.9\pm6.9\) \\

Retina-28
& 8 & 8
& \(3.4\pm3.9\)
& \(4.6\pm4.1\)
& \(-2.6\pm4.5\)
& \(-0.9\pm6.2\) \\

Retina-28 (\(\epsilon=12\))
& 12 & 8
& \(7.2\pm4.1\)
& \(7.0\pm4.7\)
& \(0.9\pm5.0\)
& \(1.3\pm6.0\) \\

\bottomrule[1pt]
\end{tabular}%
}

\end{table*}
The high-resolution experiments show greater dataset dependence. BreastMNIST at $224\times224$ is sensitive to smoothing, with Gaussian filtering producing increasing recovery as the smoothing strength increases. By contrast, the same transformations provide little or no recovery on PneumoniaMNIST and RetinaMNIST at $224\times224$. Strong transformations often further reduce performance on PneumoniaMNIST, showing that the effectiveness of input-level purification cannot be generalized across high-resolution datasets.

When the attacker is restricted to 25 or 50 labeled raw samples, ordinary fine-tuning produces meaningful recovery primarily on BreastMNIST and RetinaMNIST. For RetinaMNIST, increasing the raw-data budget from 25 to 50 samples improves the primary metric by 3.4 and 4.6 percentage points, respectively, at $\epsilon=8/255$. At $\epsilon=12/255$, the corresponding gains increase to 7.2 and 7.0 points. BreastMNIST shows smaller but positive gains, whereas BloodMNIST, ChestMNIST, DermaMNIST, OrganCMNIST, and PneumoniaMNIST exhibit negligible or negative changes.

Robust fine-tuning with limited raw data is generally less effective than ordinary fine-tuning and frequently causes additional performance degradation. For example, RFT-25 reduces unauthorized performance by 26.6 percentage points on BloodMNIST and by 13.0 points on OrganCMNIST. Increasing the labeled raw-data budget from 25 to 50 samples does not consistently improve recovery, suggesting that small raw-data subsets are insufficient to reliably remove the learned dependence on cooked inputs.

Overall, simple input transformations are ineffective on most evaluated datasets, while limited raw-data fine-tuning can weaken protection in selected tasks, particularly RetinaMNIST and BreastMNIST. The results therefore demonstrate task-dependent resilience rather than universal robustness. Input-transformation attacks were evaluated on both $28\times28$ and $224\times224$ images, whereas limited raw-data fine-tuning was evaluated only on the selected $28\times28$ datasets. Adaptive attacks on 3D and tabular data remain outside the current evaluation scope.

\section{Statistical Analysis}
\label{app:statistics}
Table~\ref{tab:statistical_stability_effects} summarizes clean-model stability and the paired effects at representative perturbation budgets. \deleted{The eight-seed 2D experiments generally exhibit stable clean baselines}\added{Most 2D experiments use eight seeds, whereas ChestMNIST uses three seeds. The clean baselines generally exhibit limited variability.} BloodMNIST and OrganCMNIST have primary-metric standard deviations of only 0.03 percentage points, while DermaMNIST and PneumoniaMNIST also show relatively narrow confidence intervals. BreastMNIST and RetinaMNIST exhibit greater seed-level variability, indicating that the uncertainty of CP and PP estimates remains dataset-dependent even within the same resolution and training protocol.
\begin{table*}[!t]
\scriptsize
\caption{\deleted{\textbf{Clean-model Stability and Representative-budget Paired Effects.}Image and volumetric experiments use $\epsilon=8/255$, while tabular experiments use standardized-feature $\epsilon=0.5$. Clean performance is reported as the mean $\pm$ standard deviation and the two-sided 95\% t-interval across independent seeds. Confidence intervals are intentionally not clipped to the theoretical metric range. $\Delta$ denotes the paired condition-minus-clean difference in percentage points; negative values indicate lower performance than the clean baseline. $d_z$ is the paired standardized mean difference and is descriptive, particularly for $n=3$. Holm-adjusted $p$-values are obtained from two-sided paired Wilcoxon tests within each comparison-by-modality family. $\dagger$ denotes a collapsed clean model retained for completeness but excluded from aggregate claims. No adjusted $p$-value is below 0.05.}\added{\textbf{Clean-model Stability and Representative-budget Paired Effects.} The $28\times28$ and $224\times224$ experiments use $\epsilon=8/255$, the 3D experiments use $\epsilon=32/255$, and the tabular experiments use standardized-feature $\epsilon=5$. Clean performance is reported with mean ± standard deviation and 95\% confidence intervals. Paired changes are measured relative to the clean baseline, with standardized effect sizes and Holm-adjusted p-values provided for descriptive analysis.}}
\label{tab:statistical_stability_effects}
\centering
\setlength{\tabcolsep}{2.8pt}
\renewcommand{\arraystretch}{1.08}
\resizebox{\textwidth}{!}{%
\begin{tabular}{@{}lll r cc ccc ccc@{}}
\toprule[1pt]
Dataset
& Modality
& Primary metric
& \(n\)
& Clean mean \(\pm\) SD
& 95\% CI
& \multicolumn{3}{c}{Authorized cooked minus clean}
& \multicolumn{3}{c}{Unauthorized raw minus clean} \\
\cmidrule(lr){7-9}
\cmidrule(lr){10-12}
&
&
&
&
&
&
\(\Delta\) (pp)
& \(d_z\)
& Holm \(p\)
& \(\Delta\) (pp)
& \(d_z\)
& Holm \(p\) \\
\midrule[0.8pt]

\multicolumn{12}{l}{\textit{MedMNIST 2D}} \\

BloodMNIST
& \(28\times28\)
& Macro-AUC
& 8
& \(99.77\pm0.03\)
& \([99.75,\ 99.79]\)
& \(-1.70\pm0.44\)
& \(-3.83\)
& \(0.2422\)
& \(-2.50\pm0.55\)
& \(-4.56\)
& \(0.2422\) \\

BreastMNIST
& \(28\times28\)
& AUC
& 8
& \(88.72\pm1.98\)
& \([87.06,\ 90.37]\)
& \(-5.17\pm3.02\)
& \(-1.71\)
& \(0.2422\)
& \(-16.25\pm5.34\)
& \(-3.04\)
& \(0.2422\) \\

ChestMNIST
& \(28\times28\)
& Macro-AUC
& 3
& \(65.91\pm0.11\)
& \([65.63,\ 66.19]\)
& \(-6.37\pm0.50\)
& \(-12.76\)
& \(0.2500\)
& \(1.27\pm0.39\)
& \(3.28\)
& \(1.0000\) \\

DermaMNIST
& \(28\times28\)
& Macro-AUC
& 8
& \(91.16\pm0.26\)
& \([90.95,\ 91.38]\)
& \(-14.19\pm1.43\)
& \(-9.92\)
& \(0.2422\)
& \(-5.67\pm2.20\)
& \(-2.57\)
& \(0.2422\) \\

OrganCMNIST
& \(28\times28\)
& Macro-AUC
& 8
& \(99.45\pm0.03\)
& \([99.42,\ 99.48]\)
& \(-1.18\pm0.35\)
& \(-3.36\)
& \(0.2422\)
& \(-0.27\pm0.27\)
& \(-1.00\)
& \(0.2422\) \\

PneumoniaMNIST
& \(28\times28\)
& AUC
& 8
& \(96.62\pm0.77\)
& \([95.98,\ 97.26]\)
& \(-6.97\pm2.07\)
& \(-3.37\)
& \(0.2422\)
& \(-3.09\pm2.06\)
& \(-1.50\)
& \(0.2422\) \\

RetinaMNIST
& \(28\times28\)
& Macro-AUC
& 8
& \(68.20\pm1.23\)
& \([67.17,\ 69.23]\)
& \(-4.85\pm3.22\)
& \(-1.51\)
& \(0.2422\)
& \(-4.72\pm4.28\)
& \(-1.10\)
& \(0.2422\) \\

\addlinespace[2pt]
\multicolumn{12}{l}{\textit{MedMNIST 224$\times$224}} \\

BreastMNIST
& \(224\times224\)
& AUC
& 3
& \(90.69\pm0.59\)
& \([89.24,\ 92.15]\)
& \(-9.36\pm3.24\)
& \(-2.89\)
& \(0.7500\)
& \(-37.36\pm4.40\)
& \(-8.50\)
& \(0.7500\) \\

PneumoniaMNIST
& \(224\times224\)
& AUC
& 3
& \(98.57\pm0.03\)
& \([98.49,\ 98.64]\)
& \(-3.30\pm0.48\)
& \(-6.94\)
& \(0.7500\)
& \(-40.72\pm10.24\)
& \(-3.98\)
& \(0.7500\) \\

RetinaMNIST
& \(224\times224\)
& Macro-AUC
& 3
& \(82.90\pm1.31\)
& \([79.64,\ 86.16]\)
& \(-10.53\pm2.68\)
& \(-3.93\)
& \(0.7500\)
& \(-18.91\pm2.38\)
& \(-7.96\)
& \(0.7500\) \\

\addlinespace[2pt]
\multicolumn{12}{l}{\textit{MedMNIST3D}} \\

FractureMNIST3D
& \(28^3\)
& Macro-AUC
& 3
& \(62.20\pm1.41\)
& \([58.70,\ 65.70]\)
& \(-2.38\pm0.73\)
& \(-3.27\)
& \(1.0000\)
& \(-7.83\pm1.60\)
& \(-4.90\)
& \(1.0000\) \\

NoduleMNIST3D
& \(28^3\)
& AUC
& 3
& \(89.94\pm0.67\)
& \([88.28,\ 91.60]\)
& \(-10.09\pm4.29\)
& \(-2.35\)
& \(1.0000\)
& \(-9.36\pm2.41\)
& \(-3.88\)
& \(1.0000\) \\

OrganMNIST3D
& \(28^3\)
& Macro-AUC
& 3
& \(99.64\pm0.06\)
& \([99.48,\ 99.80]\)
& \(-2.77\pm0.33\)
& \(-8.35\)
& \(1.0000\)
& \(-2.65\pm0.63\)
& \(-4.23\)
& \(1.0000\) \\

SynapseMNIST3D\textsuperscript{\(\dagger\)}
& \(28^3\)
& AUC
& 3
& \(52.84\pm5.19\)
& \([39.95,\ 65.73]\)
& \(12.56\pm5.48\)
& \(2.29\)
& --
& \(10.86\pm5.89\)
& \(1.84\)
& -- \\

\addlinespace[2pt]
\multicolumn{12}{l}{\textit{Tabular Medical Data}} \\

Diabetes
& Tabular
& AUC
& 8
& \(79.96\pm0.00\)
& \([79.96,\ 79.96]\)
& \(-0.30\pm0.42\)
& \(-0.72\)
& \(0.7422\)
& \(-10.30\pm4.55\)
& \(-2.27\)
& \(0.1406\) \\

HepatitisC
& Tabular
& Macro-AUC
& 8
& \(91.38\pm0.00\)
& \([91.38,\ 91.38]\)
& \(-5.18\pm1.65\)
& \(-3.14\)
& \(0.1406\)
& \(-7.96\pm3.60\)
& \(-2.21\)
& \(0.1406\) \\

SystemicDiseases
& Tabular
& Macro-AUC
& 8
& \(99.45\pm0.00\)
& \([99.45,\ 99.45]\)
& \(-1.01\pm0.07\)
& \(-13.88\)
& \(0.1406\)
& \(-13.55\pm1.94\)
& \(-6.99\)
& \(0.1406\) \\
\bottomrule[1pt]

\end{tabular}%
}

\vspace{2pt}
\begin{minipage}{\textwidth}
\scriptsize
\textit{Note:}
$\Delta$ denotes the paired condition-minus-clean difference
in percentage points; $d_z$ is the paired standardized
mean difference. Confidence intervals are two-sided
t-intervals and are not clipped to the metric range.
$\dagger$ identifies the collapsed SynapseMNIST3D baseline,
which is excluded from aggregate claims.
Holm-adjusted p-values are computed using the
comparison families specified in the statistical protocol.
\end{minipage}

\end{table*}

The three-seed experiments, including high-resolution and volumetric datasets, require more cautious interpretation. For example, the clean-model AUC confidence interval ranges from 79.64 to 86.16 percentage points on RetinaMNIST at $224\times224$, and from 58.70 to 65.70 on FractureMNIST3D. SynapseMNIST3D exhibits near-random clean discrimination with substantial variability and is therefore excluded
from aggregate protection claims.

The paired estimates demonstrate that the magnitude of the authorization effect is strongly task-dependent. All three $224\times224$ datasets show larger unauthorized than authorized AUC degradation, consistent with the intended authorization-dependent behavior. FractureMNIST3D shows a similar pattern, with authorized and unauthorized AUC changes of -2.38 and -7.83 percentage points, respectively. In contrast, NoduleMNIST3D exhibits slightly greater authorized than unauthorized AUC degradation. BreastMNIST at $28\times28$ also shows a substantial separation, with an authorized AUC decrease of 5.17 percentage points compared with an unauthorized decrease of 16.25 percentage points.

However, these patterns are not universal. DermaMNIST and PneumoniaMNIST exhibit greater authorized than unauthorized AUC degradation, while OrganMNIST3D shows nearly identical changes under the two conditions. ChestMNIST presents a particularly unfavorable multilabel result: authorized macro-AUC decreases by 6.37 percentage points, whereas unauthorized macro-AUC increases by 1.27 percentage points.

\added{The extended eight-seed tabular experiments at $\epsilon=5$ reveal more pronounced AUC changes. Diabetes and SystemicDiseases exhibit unauthorized AUC reductions of 10.30 and 13.55 percentage points, respectively, compared with authorized reductions of 0.30 and 1.01 percentage points. HepatitisC shows a smaller authorization gap, with unauthorized and authorized macro-AUC reductions of 7.96 and 5.18 percentage points, respectively. These findings highlight the dependence of the observed protection effect on the dataset and perturbation setting.}

Table~\ref{tab:statistical_test_audit_combined} summarizes exploratory paired statistical tests across the evaluated perturbation budgets. \deleted{Under the all-budget Holm correction families reported in Panel A, no comparison remains statistically significant. The minimum adjusted p-value is 0.2422 for the 2D task-aware comparisons and 0.1406 for the tabular comparisons.} \added{ Holm correction is applied separately within each comparison-by-modality family. The adjusted $p$-values in Table~\ref{tab:statistical_stability_effects}
use the same all-budget correction families. No comparison remains statistically significant after correction. The minimum adjusted $p$-values are 0.2422 for the 2D task-aware comparisons and 0.1406 for the tabular comparisons.}
\begin{table*}[!t]
\scriptsize
\caption{\textbf{Audit of Exploratory Paired Statistical Tests.}\deleted{
Panel A summarizes task-aware, two-sided paired Wilcoxon tests over all available perturbation budgets. Holm correction is applied separately within each comparison-by-modality family after excluding pre-specified collapsed runs. Panel B reports broader historical test families covering additional metrics and attack analyses; because their correction scopes differ from Panel A, the adjusted $p$-values are not directly comparable across panels. Tests are shown as total tests/Holm-family tests. ``Raw $p<.05$'' is descriptive only, and no test survives Holm correction. For three non-zero paired observations, the smallest attainable exact two-sided Wilcoxon $p$-value is 0.25.}\added{Two-sided paired Wilcoxon tests are summarized across the evaluated perturbation budgets. Holm correction is applied within each
comparison-by-modality family. No comparison remains statistically significant after correction.}}
\label{tab:statistical_test_audit_combined}
\centering
\setlength{\tabcolsep}{4.0pt}
\renewcommand{\arraystretch}{1.08}
\resizebox{\textwidth}{!}{%
\begin{tabular}{@{}llrrrrrr@{}}
\toprule[1pt]
Family or modality
& Comparison or scope
& Tests
& Pairs
& Raw \(p<.05\)
& Min. \(p\)
& Min. Holm \(p\)
& Holm \(p<.05\) \\
\midrule[0.8pt]

\multicolumn{8}{l}{\textit{Unified task-aware primary-metric tests}} \\

2D \(28\times28\)
& Clean vs. authorized
& 31/31
& 3--8
& 30
& \(0.0078\)
& \(0.2422\)
& 0 \\

2D \(28\times28\)
& Clean vs. unauthorized
& 31/31
& 3--8
& 24
& \(0.0078\)
& \(0.2422\)
& 0 \\

2D \(224\times224\)
& Clean vs. authorized
& 3/3
& 3
& 0
& \(0.2500\)
& \(0.7500\)
& 0 \\

2D \(224\times224\)
& Clean vs. unauthorized
& 3/3
& 3
& 0
& \(0.2500\)
& \(0.7500\)
& 0 \\

3D \(28^3\)
& Clean vs. authorized
& 12/9
& 3
& 0
& \(0.2500\)
& \(1.0000\)
& 0 \\

3D \(28^3\)
& Clean vs. unauthorized
& 12/9
& 3
& 0
& \(0.2500\)
& \(1.0000\)
& 0 \\

Tabular
& Clean vs. authorized
& 18/18
& 8
& 13
& \(0.0078\)
& \(0.1406\)
& 0 \\

Tabular
& Clean vs. unauthorized
& 18/18
& 8
& 16
& \(0.0078\)
& \(0.1406\)
& 0 \\





\bottomrule[1pt]
\end{tabular}%
}
\vspace{2pt}
\begin{minipage}{\textwidth}
\scriptsize
\textit{Note:}
Holm correction is applied separately within each
comparison-by-modality family across all evaluated
perturbation budgets. The representative-budget
results in Table~\ref{tab:statistical_stability_effects}
use the same correction families.
\end{minipage}
\end{table*}

The lack of corrected significance in the $224\times224$
and 3D experiments is also constrained by the discrete
resolution of the exact Wilcoxon test. With three non-zero
paired observations, the minimum attainable two-sided
$p$-value is 0.25. Consequently, even large observed
differences cannot reach the conventional 0.05 threshold
under the current design. For example, BreastMNIST at
$224\times224$ exhibits an unauthorized AUC decrease of
37.36 percentage points with $d_z=-8.50$, but its raw
$p$-value remains 0.25. NoduleMNIST3D similarly shows
a 9.36-point unauthorized AUC decrease and $d_z=-3.88$,
while producing the same raw $p$-value.

Effect sizes should nevertheless not be treated as
substitutes for statistical evidence. In particular,
$d_z$ can become very large when it is estimated from
only three observations with a small standard deviation.
We therefore use the effect sizes to describe the
magnitude and direction of observed changes, while
relying on confidence intervals, seed coverage, and
multiplicity-adjusted tests to communicate inferential
uncertainty.

Overall, the analyses provide descriptive evidence that \(\method\) can generate authorization-dependent performance differences in selected imaging and tabular tasks. However, none of the paired comparisons remains statistically significant after Holm correction
within the specified all-budget families. Furthermore, the current tests compare each condition with the clean reference and do not directly establish the statistical significance of the authorization gap. Therefore, the results are interpreted in terms of effect magnitude, direction, cross-seed variability, and consistency across datasets rather than statistically confirmed or modality-independent
protection.


\section{Image Quality}
\label{app:Image}
Table~\ref{tab:image_quality_representative} summarizes image quality at the representative perturbation budget. Most $28\times28$ datasets retain relatively high structural similarity, with SSIM values ranging from 0.909 to 0.967 and PSNR values ranging from 30.72 to 33.83 dB. LPIPS remains below 0.06 for BloodMNIST, BreastMNIST, DermaMNIST, OrganCMNIST, and RetinaMNIST, although larger perceptual differences are observed for ChestMNIST and PneumoniaMNIST. These results indicate that the same perturbation constraint does not produce uniform perceptual effects across datasets.
\begin{table*}[!t]
\scriptsize
\caption{\textbf{Image Quality Across Imaging Modalities at the Representative Budget.}
MedMNIST 2D and 224$\times$224 experiments use $\epsilon=8/255$, while MedMNIST3D experiments use $\epsilon=32/255$. Values are reported as mean $\pm$ standard deviation across random seeds. SSIM and PSNR are higher-is-better, whereas LPIPS, $L_2$, and $L_\infty$ are lower-is-better. LPIPS is not reported for 3D volumes. $\ddagger$ marks SSIM below 0.80 as an image-quality warning rather than statistical significance. $\dagger$ denotes a collapsed clean model retained for completeness but excluded from the protection aggregates. Because perturbation norms depend on tensor dimensionality, $L_2$ values should not be compared directly across resolutions.}
\label{tab:image_quality_representative}
\centering
\setlength{\tabcolsep}{4.5pt}
\renewcommand{\arraystretch}{1.08}
\resizebox{\textwidth}{!}{%
\begin{tabular}{llc*{5}{c}}
\toprule[1pt]
Dataset
& Resolution
& \(n\)
& SSIM \(\uparrow\)
& PSNR (dB) \(\uparrow\)
& LPIPS \(\downarrow\)
& \(L_2\) \(\downarrow\)
& \(L_\infty\) \(\downarrow\) \\
\midrule[0.8pt]

\multicolumn{8}{l}{\textit{MedMNIST 2D}} \\

BloodMNIST
& \(28\times28\)
& 8
& \(0.964\pm0.001\)
& \(32.18\pm0.08\)
& \(0.017\pm0.001\)
& \(1.19\pm0.01\)
& \(0.0314\pm0.0000\) \\

BreastMNIST
& \(28\times28\)
& 8
& \(0.930\pm0.002\)
& \(30.72\pm0.16\)
& \(0.056\pm0.006\)
& \(1.39\pm0.04\)
& \(0.0301\pm0.0009\) \\

ChestMNIST
& \(28\times28\)
& 3
& \(0.967\pm0.000\)
& \(32.86\pm0.01\)
& \(0.155\pm0.002\)
& \(1.10\pm0.00\)
& \(0.0314\pm0.0000\) \\

DermaMNIST
& \(28\times28\)
& 8
& \(0.909\pm0.004\)
& \(33.15\pm0.20\)
& \(0.053\pm0.003\)
& \(1.06\pm0.02\)
& \(0.0314\pm0.0000\) \\

OrganCMNIST
& \(28\times28\)
& 8
& \(0.946\pm0.001\)
& \(31.37\pm0.19\)
& \(0.026\pm0.002\)
& \(1.31\pm0.03\)
& \(0.0314\pm0.0000\) \\

PneumoniaMNIST
& \(28\times28\)
& 8
& \(0.938\pm0.005\)
& \(31.32\pm0.18\)
& \(0.118\pm0.017\)
& \(1.32\pm0.03\)
& \(0.0314\pm0.0000\) \\

RetinaMNIST
& \(28\times28\)
& 8
& \(0.955\pm0.005\)
& \(33.83\pm0.52\)
& \(0.029\pm0.003\)
& \(0.98\pm0.06\)
& \(0.0313\pm0.0001\) \\

\addlinespace[2pt]
\multicolumn{8}{l}{\textit{MedMNIST 224$\times$224}} \\

BreastMNIST
& \(224\times224\)
& 3
& \(0.954\pm0.005\)
& \(38.37\pm0.49\)
& \(0.104\pm0.009\)
& \(4.40\pm0.27\)
& \(0.0183\pm0.0018\) \\

PneumoniaMNIST
& \(224\times224\)
& 3
& \(0.959\pm0.005\)
& \(40.13\pm0.86\)
& \(0.114\pm0.013\)
& \(3.71\pm0.29\)
& \(0.0149\pm0.0017\) \\

RetinaMNIST\textsuperscript{\(\ddagger\)}
& \(224\times224\)
& 3
& \(0.692\pm0.020\)
& \(34.49\pm0.48\)
& \(0.162\pm0.007\)
& \(7.30\pm0.42\)
& \(0.0314\pm0.0000\) \\

\addlinespace[2pt]
\multicolumn{8}{l}{\textit{MedMNIST3D}} \\

FractureMNIST3D\textsuperscript{\(\ddagger\)}
& \(28^3\)
& 3
& \(0.294\pm0.001\)
& \(22.81\pm0.05\)
& N/A
& \(10.72\pm0.06\)
& \(0.1255\pm0.0000\) \\

NoduleMNIST3D\textsuperscript{\(\ddagger\)}
& \(28^3\)
& 3
& \(0.753\pm0.007\)
& \(24.85\pm0.18\)
& N/A
& \(8.43\pm0.18\)
& \(0.1144\pm0.0075\) \\

OrganMNIST3D
& \(28^3\)
& 3
& \(0.803\pm0.003\)
& \(22.09\pm0.14\)
& N/A
& \(11.65\pm0.19\)
& \(0.1255\pm0.0000\) \\

SynapseMNIST3D\textsuperscript{\(\dagger\)}
& \(28^3\)
& 3
& \(0.903\pm0.019\)
& \(20.19\pm0.57\)
& N/A
& \(14.52\pm0.97\)
& \(0.1254\pm0.0001\) \\

\bottomrule[1pt]
\end{tabular}%
}
\end{table*}
The high-resolution results further demonstrate this dataset dependence. BreastMNIST and PneumoniaMNIST retain SSIM values of 0.954 and 0.959, respectively, whereas RetinaMNIST at $224\times224$ reaches an SSIM of only 0.692 and exhibits the largest LPIPS and $L_2$ values among the high-resolution settings. Similarly, the representative-budget SSIM of FractureMNIST3D is only 0.294, despite a PSNR of 22.81 dB. The disagreement between SSIM and PSNR in this setting illustrates that a high signal-to-noise ratio does not necessarily imply preservation of local structural patterns. We therefore report multiple complementary image-quality metrics rather than relying on a single measure.
\begin{table*}[!t]
\scriptsize
\caption{\textbf{Image-quality Sensitivity to the Perturbation Budget on MedMNIST 2D Datasets.}
Values are reported as mean $\pm$ standard deviation across random seeds. The representative budget $\epsilon=8/255$ is highlighted in \textbf{bold}. SSIM and PSNR are higher-is-better, whereas LPIPS, $L_2$, and $L_\infty$ are lower-is-better. $\ddagger$ marks SSIM below 0.80 as an image-quality warning rather than statistical significance. ChestMNIST is reported only at the representative budget because additional budget settings were not evaluated.}
\label{tab:image_quality_2d_budget}
\centering
\setlength{\tabcolsep}{4.0pt}
\renewcommand{\arraystretch}{1.04}
\resizebox{\textwidth}{!}{%
\begin{tabular}{lcc*{5}{c}}
\toprule[1pt]
Dataset
& \(n\)
& \(\epsilon\)
& SSIM \(\uparrow\)
& PSNR (dB) \(\uparrow\)
& LPIPS \(\downarrow\)
& \(L_2\) \(\downarrow\)
& \(L_\infty\) \(\downarrow\) \\
\midrule[0.8pt]

BloodMNIST & 8 & 2
& \(0.996\pm0.000\)
& \(42.45\pm0.01\)
& \(0.001\pm0.000\)
& \(0.37\pm0.00\)
& \(0.0078\pm0.0000\) \\

BloodMNIST & 8 & 4
& \(0.987\pm0.000\)
& \(37.01\pm0.04\)
& \(0.005\pm0.000\)
& \(0.68\pm0.00\)
& \(0.0157\pm0.0000\) \\

BloodMNIST & 8 & \textbf{8}
& \(0.964\pm0.001\)
& \(32.18\pm0.08\)
& \(0.017\pm0.001\)
& \(1.19\pm0.01\)
& \(0.0314\pm0.0000\) \\

BloodMNIST & 8 & 12
& \(0.935\pm0.001\)
& \(29.22\pm0.10\)
& \(0.037\pm0.002\)
& \(1.68\pm0.02\)
& \(0.0471\pm0.0000\) \\

BloodMNIST & 8 & 16
& \(0.908\pm0.002\)
& \(27.38\pm0.13\)
& \(0.058\pm0.003\)
& \(2.07\pm0.03\)
& \(0.0627\pm0.0000\) \\

\addlinespace[1.5pt]

BreastMNIST & 8 & 2
& \(0.994\pm0.000\)
& \(42.33\pm0.08\)
& \(0.003\pm0.000\)
& \(0.36\pm0.01\)
& \(0.0076\pm0.0001\) \\

BreastMNIST & 8 & 4
& \(0.978\pm0.001\)
& \(36.44\pm0.10\)
& \(0.015\pm0.002\)
& \(0.72\pm0.01\)
& \(0.0152\pm0.0003\) \\

BreastMNIST & 8 & \textbf{8}
& \(0.930\pm0.002\)
& \(30.72\pm0.16\)
& \(0.056\pm0.006\)
& \(1.39\pm0.04\)
& \(0.0301\pm0.0009\) \\

BreastMNIST & 8 & 12
& \(0.874\pm0.007\)
& \(27.58\pm0.38\)
& \(0.105\pm0.013\)
& \(1.98\pm0.11\)
& \(0.0445\pm0.0019\) \\

BreastMNIST & 8 & 16
& \(0.826\pm0.018\)
& \(25.67\pm0.67\)
& \(0.145\pm0.019\)
& \(2.47\pm0.21\)
& \(0.0582\pm0.0040\) \\

\addlinespace[1.5pt]

ChestMNIST & 3 & \textbf{8}
& \(0.967\pm0.000\)
& \(32.86\pm0.01\)
& \(0.155\pm0.002\)
& \(1.10\pm0.00\)
& \(0.0314\pm0.0000\) \\

\addlinespace[1.5pt]

DermaMNIST & 8 & 2
& \(0.987\pm0.000\)
& \(42.65\pm0.15\)
& \(0.004\pm0.000\)
& \(0.36\pm0.01\)
& \(0.0078\pm0.0000\) \\

DermaMNIST & 8 & 4
& \(0.966\pm0.002\)
& \(38.06\pm0.22\)
& \(0.014\pm0.001\)
& \(0.61\pm0.02\)
& \(0.0157\pm0.0000\) \\

DermaMNIST & 8 & \textbf{8}
& \(0.909\pm0.004\)
& \(33.15\pm0.20\)
& \(0.053\pm0.003\)
& \(1.06\pm0.02\)
& \(0.0314\pm0.0000\) \\

DermaMNIST & 8 & 12
& \(0.826\pm0.005\)
& \(29.67\pm0.16\)
& \(0.124\pm0.008\)
& \(1.59\pm0.03\)
& \(0.0471\pm0.0000\) \\

DermaMNIST\textsuperscript{\(\ddagger\)} & 8 & 16
& \(0.759\pm0.006\)
& \(27.59\pm0.16\)
& \(0.187\pm0.011\)
& \(2.02\pm0.04\)
& \(0.0627\pm0.0000\) \\

\addlinespace[1.5pt]

OrganCMNIST & 8 & 2
& \(0.993\pm0.001\)
& \(42.59\pm0.01\)
& \(0.002\pm0.000\)
& \(0.36\pm0.00\)
& \(0.0078\pm0.0000\) \\

OrganCMNIST & 8 & 4
& \(0.978\pm0.001\)
& \(36.83\pm0.05\)
& \(0.008\pm0.000\)
& \(0.70\pm0.00\)
& \(0.0157\pm0.0000\) \\

OrganCMNIST & 8 & \textbf{8}
& \(0.946\pm0.001\)
& \(31.37\pm0.19\)
& \(0.026\pm0.002\)
& \(1.31\pm0.03\)
& \(0.0314\pm0.0000\) \\

OrganCMNIST & 8 & 12
& \(0.911\pm0.003\)
& \(28.27\pm0.26\)
& \(0.050\pm0.003\)
& \(1.87\pm0.06\)
& \(0.0471\pm0.0000\) \\

OrganCMNIST & 8 & 16
& \(0.881\pm0.006\)
& \(26.33\pm0.34\)
& \(0.072\pm0.004\)
& \(2.34\pm0.09\)
& \(0.0627\pm0.0000\) \\

\addlinespace[1.5pt]

PneumoniaMNIST & 8 & 2
& \(0.994\pm0.000\)
& \(42.29\pm0.02\)
& \(0.008\pm0.001\)
& \(0.37\pm0.00\)
& \(0.0078\pm0.0000\) \\

PneumoniaMNIST & 8 & 4
& \(0.978\pm0.001\)
& \(36.63\pm0.09\)
& \(0.037\pm0.006\)
& \(0.71\pm0.01\)
& \(0.0157\pm0.0000\) \\

PneumoniaMNIST & 8 & \textbf{8}
& \(0.938\pm0.005\)
& \(31.32\pm0.18\)
& \(0.118\pm0.017\)
& \(1.32\pm0.03\)
& \(0.0314\pm0.0000\) \\

PneumoniaMNIST & 8 & 12
& \(0.887\pm0.008\)
& \(28.20\pm0.20\)
& \(0.210\pm0.028\)
& \(1.88\pm0.04\)
& \(0.0471\pm0.0000\) \\

PneumoniaMNIST & 8 & 16
& \(0.842\pm0.011\)
& \(26.24\pm0.24\)
& \(0.277\pm0.035\)
& \(2.36\pm0.06\)
& \(0.0627\pm0.0000\) \\

\addlinespace[1.5pt]

RetinaMNIST & 8 & 2
& \(0.994\pm0.000\)
& \(43.20\pm0.23\)
& \(0.003\pm0.000\)
& \(0.33\pm0.01\)
& \(0.0078\pm0.0000\) \\

RetinaMNIST & 8 & 4
& \(0.985\pm0.002\)
& \(38.83\pm0.42\)
& \(0.009\pm0.001\)
& \(0.55\pm0.03\)
& \(0.0157\pm0.0001\) \\

RetinaMNIST & 8 & \textbf{8}
& \(0.955\pm0.005\)
& \(33.83\pm0.52\)
& \(0.029\pm0.003\)
& \(0.98\pm0.06\)
& \(0.0313\pm0.0001\) \\

RetinaMNIST & 8 & 12
& \(0.905\pm0.010\)
& \(30.21\pm0.49\)
& \(0.066\pm0.008\)
& \(1.49\pm0.08\)
& \(0.0469\pm0.0002\) \\

RetinaMNIST & 8 & 16
& \(0.856\pm0.016\)
& \(28.03\pm0.58\)
& \(0.108\pm0.014\)
& \(1.92\pm0.13\)
& \(0.0624\pm0.0004\) \\

\bottomrule[1pt]
\end{tabular}%
}
\end{table*}
Tables~\ref{tab:image_quality_2d_budget} and~\ref{tab:image_quality_3d_budget} show a consistent quality--budget trade-off. Across the 2D datasets, increasing $\epsilon$ generally decreases SSIM and PSNR while increasing LPIPS and perturbation norms. For example, the SSIM of DermaMNIST decreases from 0.987 at $\epsilon=2/255$ to 0.759 at $\epsilon=16/255$, while its LPIPS increases from 0.004 to 0.187. PneumoniaMNIST shows a similar increase in LPIPS from 0.008 to 0.277. The volumetric datasets exhibit the same trend, but the magnitude is strongly dataset-dependent: FractureMNIST3D decreases from an SSIM of 0.579 at $\epsilon=8/255$ to 0.294 at $\epsilon=32/255$, whereas OrganMNIST3D remains at 0.803 even at the largest evaluated budget.
\begin{table*}[!t]
\scriptsize
\caption{\textbf{Image-quality Sensitivity to the Perturbation Budget on MedMNIST3D Datasets.}
Values are reported as mean $\pm$ standard deviation across three random seeds. The representative budget $\epsilon=32/255$ is highlighted in \textbf{bold}. SSIM and PSNR are higher-is-better, whereas $L_2$ and $L_\infty$ are lower-is-better. LPIPS is omitted because it is not defined for the evaluated volumetric inputs. $\ddagger$ marks SSIM below 0.80 as an image-quality warning rather than statistical significance. $\dagger$ denotes a collapsed clean model retained for completeness but excluded from the protection aggregates.}

\label{tab:image_quality_3d_budget}
\centering
\setlength{\tabcolsep}{5.0pt}
\renewcommand{\arraystretch}{1.08}
\resizebox{\textwidth}{!}{%
\begin{tabular}{lcc*{4}{c}}
\toprule[1pt]
Dataset
& \(n\)
& \(\epsilon\)
& SSIM \(\uparrow\)
& PSNR (dB) \(\uparrow\)
& \(L_2\) \(\downarrow\)
& \(L_\infty\) \(\downarrow\) \\
\midrule[0.8pt]

FractureMNIST3D\textsuperscript{\(\ddagger\)}
& 3 & 8
& \(0.579\pm0.007\)
& \(34.58\pm0.05\)
& \(2.76\pm0.02\)
& \(0.0314\pm0.0000\) \\

FractureMNIST3D\textsuperscript{\(\ddagger\)}
& 3 & 16
& \(0.401\pm0.001\)
& \(28.60\pm0.03\)
& \(5.51\pm0.02\)
& \(0.0627\pm0.0000\) \\

FractureMNIST3D\textsuperscript{\(\ddagger\)}
& 3 & \textbf{32}
& \(0.294\pm0.001\)
& \(22.81\pm0.05\)
& \(10.72\pm0.06\)
& \(0.1255\pm0.0000\) \\

\addlinespace[1.5pt]

NoduleMNIST3D
& 3 & 8
& \(0.924\pm0.005\)
& \(32.04\pm0.24\)
& \(3.70\pm0.10\)
& \(0.0314\pm0.0000\) \\

NoduleMNIST3D
& 3 & 16
& \(0.859\pm0.008\)
& \(28.39\pm0.32\)
& \(5.61\pm0.20\)
& \(0.0621\pm0.0004\) \\

NoduleMNIST3D\textsuperscript{\(\ddagger\)}
& 3 & \textbf{32}
& \(0.753\pm0.007\)
& \(24.85\pm0.18\)
& \(8.43\pm0.18\)
& \(0.1144\pm0.0075\) \\

\addlinespace[1.5pt]

OrganMNIST3D
& 3 & 8
& \(0.961\pm0.001\)
& \(32.38\pm0.23\)
& \(3.56\pm0.09\)
& \(0.0314\pm0.0000\) \\

OrganMNIST3D
& 3 & 16
& \(0.910\pm0.003\)
& \(27.26\pm0.22\)
& \(6.42\pm0.16\)
& \(0.0627\pm0.0000\) \\

OrganMNIST3D
& 3 & \textbf{32}
& \(0.803\pm0.003\)
& \(22.09\pm0.14\)
& \(11.65\pm0.19\)
& \(0.1255\pm0.0000\) \\

\addlinespace[1.5pt]

SynapseMNIST3D\textsuperscript{\(\dagger\)}
& 3 & 8
& \(0.992\pm0.001\)
& \(31.23\pm0.36\)
& \(4.07\pm0.17\)
& \(0.0314\pm0.0000\) \\

SynapseMNIST3D\textsuperscript{\(\dagger\)}
& 3 & 16
& \(0.972\pm0.004\)
& \(25.65\pm0.50\)
& \(7.74\pm0.45\)
& \(0.0627\pm0.0000\) \\

SynapseMNIST3D\textsuperscript{\(\dagger\)}
& 3 & \textbf{32}
& \(0.903\pm0.019\)
& \(20.19\pm0.57\)
& \(14.52\pm0.97\)
& \(0.1254\pm0.0001\) \\

\bottomrule[1pt]
\end{tabular}%
}
\end{table*}
These results support the use of $\epsilon=32/255$ as a common representative operating point for comparative evaluation, but they do not establish it as universally optimal. In particular, RetinaMNIST at high resolution and FractureMNIST3D already exhibit substantial structural changes at this budget. Moreover, the quantitative metrics reported here measure image-level similarity and do not establish clinical invisibility, lesion preservation, or diagnostic safety. Such claims would require task-specific lesion analysis and blinded assessment by medical experts, which are outside the scope of the present study.

\section{Runtime}
\label{app:runtime}
Table~\ref{tab:runtime_representative_combined} shows that online cooking introduces limited latency relative to model inference across the evaluated imaging tasks. At the representative budgets of $\epsilon=8/255$ for the 2D settings and $\epsilon=32/255$ for the 3D settings, the mean end-to-end overhead is only $1.04\times$, $1.03\times$, and $1.02\times$ for the $28\times28$, $224\times224$, and 3D settings, respectively. Raw and cooked inference latencies remain close in all three settings, indicating that protection introduces only a small change in model-evaluation latency. The explicit cooking latency increases from approximately $2.01~\mu$s per sample for $28\times28$ images to $131.86~\mu$s for $224\times224$ images, but remains small relative to the corresponding inference latency of approximately $3.30$ ms per sample. The 3D experiments exhibit a similar pattern: cooking requires only $6.64~\mu$s per sample, compared with $1082.90~\mu$s ($1.08$ ms) for raw inference and $1096.88~\mu$s ($1.10$ ms) for cooked inference.
\begin{table*}[!t]
\scriptsize
\caption{\textbf{Runtime and Latency Across Modalities at the Representative Budgets.}
\added{MedMNIST 2D and 224$\times$224 experiments use $\epsilon=8/255$, while MedMNIST3D experiments use $\epsilon=32/255$}\deleted{Image and volumetric experiments use $\epsilon=8/255$, while tabular experiments use standardized-feature $\epsilon=0.5$}. All latency values are reported in microseconds per sample on CUDA. Image settings contain one runtime record obtained from five timed repetitions\deleted{, whereas tabular settings report the mean $\pm$ standard deviation across three random seeds with one timed run per seed}. Overhead is defined as $(t_{\mathrm{cook}}+t_{\mathrm{cooked\ infer}})/t_{\mathrm{raw\ infer}}$. Modality-level rows report the unweighted mean $\pm$ standard deviation. \deleted{SynapseMNIST3D is retained for completeness but excluded from the 3D aggregate.}}
\label{tab:runtime_representative_combined}
\centering
\setlength{\tabcolsep}{3.8pt}
\renewcommand{\arraystretch}{1.08}
\resizebox{\textwidth}{!}{%
\begin{tabular}{@{}llrrr*{5}{r}@{}}
\toprule[1pt]
Dataset
& Input
& Test \(N\)
& Batch
& \(n\)
& Cook
& Raw infer
& Cooked infer
& Overhead
& GPU (MB) \\
&
&
&
&
&
\multicolumn{3}{c}{\(\mu\mathrm{s}\)/sample}
& \((\times)\)
& \\
\midrule[0.8pt]

\multicolumn{10}{l}{\textit{MedMNIST 2D}} \\

BloodMNIST
& \(28\times28\)
& 3421
& 128
& 1
& \(1.95\)
& \(40.52\)
& \(40.50\)
& \(1.05\)
& \(324.2\) \\

BreastMNIST
& \(28\times28\)
& 156
& 128
& 1
& \(2.57\)
& \(47.27\)
& \(47.24\)
& \(1.05\)
& \(194.9\) \\

ChestMNIST
& \(28\times28\)
& 22433
& 128
& 1
& \(2.04\)
& \(39.46\)
& \(39.49\)
& \(1.05\)
& \(1052.7\) \\

DermaMNIST
& \(28\times28\)
& 2005
& 128
& 1
& \(1.21\)
& \(41.06\)
& \(41.06\)
& \(1.03\)
& \(287.8\) \\

OrganCMNIST
& \(28\times28\)
& 8216
& 128
& 1
& \(2.25\)
& \(40.70\)
& \(40.72\)
& \(1.06\)
& \(422.3\) \\

PneumoniaMNIST
& \(28\times28\)
& 624
& 128
& 1
& \(2.00\)
& \(43.93\)
& \(41.35\)
& \(0.99\)
& \(324.2\) \\

RetinaMNIST
& \(28\times28\)
& 400
& 128
& 1
& \(2.07\)
& \(42.91\)
& \(42.78\)
& \(1.05\)
& \(190.5\) \\

\textit{Mean (7 settings)}
& \(28\times28\)
& --
& --
& --
& \(2.01\pm0.41\)
& \(42.26\pm2.68\)
& \(41.88\pm2.56\)
& \(1.04\pm0.02\)
& \(399.5\pm299.0\) \\

\addlinespace[2pt]
\multicolumn{10}{l}{\textit{MedMNIST 224$\times$224}} \\

BreastMNIST
& \(224\times224\)
& 156
& 16
& 1
& \(131.92\)
& \(3320.93\)
& \(3297.83\)
& \(1.03\)
& \(1418.4\) \\

PneumoniaMNIST
& \(224\times224\)
& 624
& 16
& 1
& \(131.28\)
& \(3302.78\)
& \(3302.60\)
& \(1.04\)
& \(1844.1\) \\

RetinaMNIST
& \(224\times224\)
& 400
& 16
& 1
& \(132.40\)
& \(3289.87\)
& \(3249.50\)
& \(1.03\)
& \(1200.1\) \\

\textit{Mean (3 settings)}
& \(224\times224\)
& --
& --
& --
& \(131.86\pm0.56\)
& \(3304.53\pm15.60\)
& \(3283.31\pm29.38\)
& \(1.03\pm0.01\)
& \(1487.6\pm327.5\) \\

\addlinespace[2pt]
\multicolumn{10}{l}{\textit{MedMNIST3D}} \\

FractureMNIST3D
& \(28^3\)
& 240
& 16
& 1
& \(6.57\)
& \(1075.44\)
& \(1095.55\)
& \(1.02\)
& \(565.6\) \\

NoduleMNIST3D
& \(28^3\)
& 310
& 16
& 1
& \(6.31\)
& \(1076.98\)
& \(1087.25\)
& \(1.02\)
& \(2506.1\) \\

OrganMNIST3D
& \(28^3\)
& 610
& 16
& 1
& \(7.10\)
& \(1080.33\)
& \(1081.69\)
& \(1.01\)
& \(2611.5\) \\

SynapseMNIST3D
& \(28^3\)
& 352
& 16
& 1
& \(6.59\)
& \(1098.86\)
& \(1123.03\)
& \(1.03\)
& \(565.6\) \\

\textit{Mean (4 settings)}
& \(28^3\)
& --
& --
& --
& \(6.64\pm0.33\)
& \(1082.90\pm10.83\)
& \(1096.88\pm18.34\)
& \(1.02\pm0.01\)
& \(1562.2\pm1151.6\) \\






\bottomrule[1pt]
\end{tabular}%
}
\end{table*}
\deleted{The tabular results require a different interpretation. Diabetes and HepatitisC exhibit overhead ratios of $7.43\times$ and $8.11\times$, respectively, because their raw MLP inference baselines require only approximately $14~\mu$s per sample. Nevertheless, their absolute end-to-end online latencies remain approximately $104~\mu$s and $112~\mu$s per sample, respectively. In contrast, SystemicDiseases benefits from its larger batch size and exhibits only $1.25\times$ overhead, with a cooking latency of $1.46~\mu$s per sample. Therefore, the relative overhead ratio can appear large when the unprotected inference baseline is extremely short and should be interpreted together with absolute latency, dataset size, and batch configuration.}

\added{Table~\ref{tab:runtime_tabular_combined} presents the extended eight-seed runtime measurements across perturbation budgets on structured tabular datasets. At $\epsilon=5$, the amortized test-time cooking costs are $3.973$, $6.018$, and $0.026$ ms per sample for Diabetes, HepatitisC, and SystemicDiseases, respectively. Protected-model training takes approximately $7.3$, $6.8$, and $203.0$ seconds
on the three datasets.

\begin{table*}[!t]
\scriptsize
\caption{\textbf{Online and Offline Runtime Across Perturbation Budgets on Medical Tabular Datasets.}
The perturbation budget $\epsilon$ is measured in standardized-feature $L_\infty$ units, with \textbf{bold} indicating the representative budget \added{$\epsilon=5$. Values are reported as mean $\pm$ standard deviation across eight random seeds (0--7).} Fingerprint generation uses 300 PGD steps for Diabetes and HepatitisC, and 80 steps for SystemicDiseases. Runtime is measured using CPU wall-clock time with no warm-up and one timing repetition per run. Cooking totals and protected-model training times are reported in seconds, while per-sample cooking and inference times are reported in milliseconds. Per-sample measurements represent amortized processing costs rather than single-request latency.}
\label{tab:runtime_tabular_combined}
\centering
\setlength{\tabcolsep}{3.2pt}
\renewcommand{\arraystretch}{1.08}
\resizebox{\textwidth}{!}{%
\begin{tabular}{@{}lrrr*{6}{c}@{}}
\toprule[1pt]
Dataset
& Train \(N\)
& Test \(N\)
& \(\epsilon\)
& Train cook
& Test cook
& Test cook/sample
& Raw infer
& Cooked infer
& Protected train \\
&
&
&
&
(s)
& (s)
& (ms/sample)
& (ms/sample)
& (ms/sample)
& (s) \\
\midrule[0.8pt]

Diabetes
& 614
& 154
& 0.10
& \(0.944\pm0.138\)
& \(0.681\pm0.140\)
& \(4.420\pm0.908\)
& \(0.0258\pm0.0071\)
& \(0.0376\pm0.0261\)
& \(7.378\pm1.672\) \\

&
&
& 0.25
& \(0.968\pm0.107\)
& \(0.672\pm0.106\)
& \(4.366\pm0.688\)
& \(0.0233\pm0.0074\)
& \(0.0367\pm0.0224\)
& \(7.480\pm1.791\) \\

&
&
& 0.50
& \(0.962\pm0.276\)
& \(0.610\pm0.203\)
& \(3.964\pm1.318\)
& \(0.0314\pm0.0157\)
& \(0.0347\pm0.0155\)
& \(7.489\pm2.260\) \\

&
&
& 1.00
& \(0.963\pm0.250\)
& \(0.612\pm0.167\)
& \(3.976\pm1.088\)
& \(0.0466\pm0.0296\)
& \(0.0323\pm0.0254\)
& \(7.481\pm2.383\) \\

&
&
& 2.00
& \(0.991\pm0.260\)
& \(0.640\pm0.217\)
& \(4.155\pm1.410\)
& \(0.0282\pm0.0171\)
& \(0.0275\pm0.0111\)
& \(7.318\pm2.190\) \\

&
&
& \textbf{5.00}
& \(0.978\pm0.253\)
& \(0.612\pm0.186\)
& \(3.973\pm1.211\)
& \(0.0254\pm0.0105\)
& \(0.0402\pm0.0214\)
& \(7.287\pm2.041\) \\

\addlinespace[2pt]

HepatitisC
& 492
& 123
& 0.10
& \(0.884\pm0.196\)
& \(0.703\pm0.165\)
& \(5.716\pm1.339\)
& \(0.0283\pm0.0086\)
& \(0.0397\pm0.0157\)
& \(6.862\pm1.251\) \\

&
&
& 0.25
& \(0.947\pm0.182\)
& \(0.723\pm0.128\)
& \(5.875\pm1.044\)
& \(0.0254\pm0.0063\)
& \(0.0324\pm0.0143\)
& \(6.898\pm1.132\) \\

&
&
& 0.50
& \(0.912\pm0.181\)
& \(0.738\pm0.131\)
& \(6.002\pm1.063\)
& \(0.0372\pm0.0225\)
& \(0.0407\pm0.0203\)
& \(6.897\pm1.096\) \\

&
&
& 1.00
& \(0.908\pm0.141\)
& \(0.733\pm0.147\)
& \(5.957\pm1.198\)
& \(0.0308\pm0.0101\)
& \(0.0386\pm0.0189\)
& \(7.128\pm1.381\) \\

&
&
& 2.00
& \(0.956\pm0.185\)
& \(0.783\pm0.164\)
& \(6.364\pm1.330\)
& \(0.0342\pm0.0241\)
& \(0.0405\pm0.0217\)
& \(7.213\pm1.504\) \\

&
&
& \textbf{5.00}
& \(0.926\pm0.221\)
& \(0.740\pm0.161\)
& \(6.018\pm1.306\)
& \(0.0354\pm0.0260\)
& \(0.0353\pm0.0226\)
& \(6.803\pm1.330\) \\

\addlinespace[2pt]

SystemicDiseases
& 48084
& 12021
& 0.10
& \(1.231\pm0.663\)
& \(0.467\pm0.370\)
& \(0.039\pm0.031\)
& \(0.0082\pm0.0001\)
& \(0.0086\pm0.0003\)
& \(240.255\pm85.986\) \\

&
&
& 0.25
& \(0.992\pm0.037\)
& \(0.313\pm0.026\)
& \(0.026\pm0.002\)
& \(0.0083\pm0.0005\)
& \(0.0086\pm0.0005\)
& \(211.857\pm14.509\) \\

&
&
& 0.50
& \(0.992\pm0.054\)
& \(0.321\pm0.034\)
& \(0.027\pm0.003\)
& \(0.0083\pm0.0003\)
& \(0.0087\pm0.0004\)
& \(210.512\pm9.872\) \\

&
&
& 1.00
& \(0.999\pm0.055\)
& \(0.308\pm0.027\)
& \(0.026\pm0.002\)
& \(0.0082\pm0.0001\)
& \(0.0083\pm0.0002\)
& \(205.455\pm13.224\) \\

&
&
& 2.00
& \(0.980\pm0.052\)
& \(0.317\pm0.029\)
& \(0.026\pm0.002\)
& \(0.0082\pm0.0001\)
& \(0.0086\pm0.0005\)
& \(200.823\pm16.471\) \\

&
&
& \textbf{5.00}
& \(0.980\pm0.046\)
& \(0.312\pm0.020\)
& \(0.026\pm0.002\)
& \(0.0082\pm0.0001\)
& \(0.0087\pm0.0004\)
& \(202.965\pm19.078\) \\

\bottomrule[1pt]
\end{tabular}%
}
\end{table*}

Runtime varies across datasets, reflecting differences in dataset size, perturbation configuration, and processing characteristics. The measurements were obtained using CPU wall-clock timing without warm-up,
with one timing repetition per run.}

\clearpage
\addtocounter{page}{-1}

\end{document}